\documentclass[10pt]{article} % For LaTeX2e
\usepackage[preprint]{tmlr}
\usepackage{amsmath,amsfonts,bm}

\def\eqref#1{equation~\ref{#1}}
\def\1{\bm{1}}

\DeclareMathAlphabet{\mathsfit}{\encodingdefault}{\sfdefault}{m}{sl}
\SetMathAlphabet{\mathsfit}{bold}{\encodingdefault}{\sfdefault}{bx}{n}

\usepackage{hyperref}
\usepackage{url}

\usepackage{graphicx}
\usepackage{booktabs}
\usepackage{amsmath}
\usepackage{float}
\usepackage{xcolor}
\usepackage{subcaption}
\usepackage{multirow}
\usepackage{rotating}
\usepackage{algorithm}
\usepackage{algpseudocode} % algorithmic environments/commands
\usepackage{tikz}
\usetikzlibrary{positioning,fit}
\usepackage{makecell}
\usepackage{amssymb}
\usepackage{colortbl}

\definecolor{greenfirst}{RGB}{225, 245, 220}   % lightest green for first column
\definecolor{greenalt}{RGB}{210, 240, 190}   % very light green for alternating rows
\definecolor{greencol}{RGB}{180, 230, 140}   % medium green for first column
\definecolor{greenbest}{RGB}{130, 200, 80}   % dark green for best values

\newcommand{\bn}[2]{\shortstack{#1 \\[1.5pt] #2}}

\title{Hierarchical Prompt Learning for Hyperbolic \\Vision-Language Models}

\author{\name Andro Erdelez \email aer@ece.au.dk \\
      \addr Aarhus University
      \AND
      \name Pascal Mettes \email P.S.M.Mettes@uva.nl \\
      \addr University of Amsterdam
      \AND
      \name Behzad Bozorgtabar \email behzad@ece.au.dk \\
      \addr Aarhus University}

\def\month{MM}  % Insert correct month for camera-ready version
\def\year{YYYY} % Insert correct year for camera-ready version
\def\openreview{\url{https://openreview.net/forum?id=XXXX}} % Insert correct link to OpenReview for camera-ready version

\begin{document}

\maketitle

\begin{abstract}
Hyperbolic vision-language models (VLMs) represent image and text features in a geometry naturally suited to hierarchy, but their adaptation to downstream tasks has largely relied on fixed prompts. Existing prompt learning methods, meanwhile, treat class labels as a flat set and do not exploit available taxonomic structure. We address this gap with a hierarchical prompt learning plug-in for frozen hyperbolic VLMs. Given a fixed offline parent-class hierarchy, it augments a class prompt learner with a separate parent prompt learner, parent-level supervision, hyperbolic entailment regularization, and parent-feedback logit fusion. We instantiate the method with CoOp, CoCoOp and MaPLe, yielding HyPLO, CoHyPLO and MaHyPLO. Across the standard 11-dataset benchmark, all variants improve base-to-new generalization and cross-dataset transfer, and remain comparable to their prompt learning baselines under domain shift. Six hierarchical metrics and embedding analyses show that the method produces more taxonomically consistent predictions and induces a hierarchy-consistent organization of parent, class, and image embeddings in hyperbolic space. Its gains are largest when novel classes must be placed within a fixed taxonomy, and smallest for fine-grained confusions among sibling classes or shifts affecting only the image distribution.
\end{abstract}

\pagenumbering{arabic}

\section{Introduction}

Hyperbolic geometry has emerged in recent years as a compelling alternative inductive bias for representation learning. Its exponential volume growth with distance can represent hierarchical relations compactly, often with lower distortion than Euclidean embeddings of comparable dimension \citep{nickel2017poincare, nickel2018learning, sala2018representation}. This connection has motivated work on hyperbolic graph learning, language modeling, computer vision, and foundation models \citep{mettes2024hyperbolic, he2025hyperbolic}.

Vision-language models (VLMs) are a particularly natural setting for this perspective. The concepts they align (\textit{words, objects, scenes}) are themselves organized into rich semantic hierarchies: a \textit{golden retriever} is a \textit{dog} that is an \textit{animal}; an image of a \textit{kitchen} is also an image of an \textit{indoor space}. A recent line of work has begun to bring hyperbolic geometry to vision-language alignment, lifting image and text representations from CLIP-style encoders \citep{radford2021learning} into the Poincar\'e or Lorentz model and reporting encouraging results in hierarchical retrieval, compositional reasoning, and uncertainty-aware classification \citep{desai2023hyperbolic, mandica2024hyperbolic, pal2024compositional, poppi2025hyperbolic}. The picture that emerges is one in which hyperbolic representations offer a valuable complement to the Euclidean default in vision-language learning, particularly where hierarchy is a part of the signal.

In parallel, prompt learning has become a widely used approach for adapting large VLMs to downstream tasks without updating the backbone parameters. Methods such as CoOp and CoCoOp \citep{zhou2022learning, zhou2022conditional} replace hand-written templates \citep{radford2021learning} with continuous prompt vectors optimized through a frozen text encoder. Subsequent methods have extended this paradigm by, for example, adapting prompts to individual images or coupling textual prompts with visual prompting \citep{khattak2023maple, khattak2023self, roy2024consistency}. Although recent work has explored hyperbolic prompting for VLMs \citep{peng2026hpl}, the integration of hyperbolic geometry with general-purpose hierarchical prompt learning for frozen vision-language adaptation remains underexplored.

A second observation concerns the data side. Many prompt learning benchmarks come with, or naturally admit, semantic hierarchies, such as WordNet hypernyms for ImageNet \citep{miller1995wordnet, deng2009imagenet} and scene-type groupings for SUN397 \citep{xiao2010sun}. Standard prompt learners nevertheless predict over a flat vocabulary of leaf classes, leaving this supervisory signal unused. Prior work has incorporated hierarchy through multimodal coupling or dedicated graph-based components \citep{khattak2023maple, zheng2025hierarchical, xia2025hgclip}. In contrast, we encode a fixed parent-class hierarchy directly through parallel class and parent prompt learners, hyperbolic entailment regularization, and hierarchy-aware inference.

In this work, we propose a prompt-learner-agnostic hyperbolic hierarchical plug-in for frozen hyperbolic VLMs. Given a fixed parent-class hierarchy, it augments an existing prompt learner with: (i) a separate prompt learner for parent labels, (ii) class and parent classification in hyperbolic space, (iii) entailment-based regularization of parent-class-image relations, and (iv) parent-feedback logit fusion at inference. Applied to CoOp, CoCoOp, and MaPLe, the framework yields \textbf{HyPLO}, \textbf{CoHyPLO}, and \textbf{MaHyPLO} respectively. For ImageNet, the fixed hierarchy is derived from WordNet. For datasets without a suitable native taxonomy, we construct a fixed four-level (root-grandparent-parent-class) taxonomy offline by grouping fine-grained classes into coarse semantic parent categories through a structured and validated LLM-assisted procedure. The hierarchy remains fixed throughout training and inference, and training uses only direct parent-class relations.

We evaluate the proposed methods on the standard 11-dataset CoOp benchmark, including base-to-new generalization, cross-dataset transfer, and domain generalization. Across these settings, hierarchical prompt learning improves or matches the underlying prompt learners while yielding predictions that better respect the supplied taxonomy, as measured by six hierarchical metrics. Our analyses further identify settings in which the benefits are limited, including fine-grained confusions among sibling classes and the variable-depth WordNet structure of ImageNet. 

\paragraph{Contributions.} We summarize our contributions as follows:
\begin{itemize}
  \item We introduce a prompt-learner-agnostic hyperbolic hierarchical plug-in for frozen hyperbolic VLMs. It augments an existing prompt learner with parent prompts, hyperbolic class and parent supervision, entailment-based geometric regularization, and hierarchy-aware inference.

  \item We instantiate the framework for CoOp, CoCoOp, and MaPLe, yielding \textbf{HyPLO}, \textbf{CoHyPLO}, and \textbf{MaHyPLO}. We evaluate these variants on the 11-dataset CoOp benchmark, including base-to-new generalization, cross-dataset transfer, and domain generalization.

  \item We provide an offline hierarchy construction protocol that combines native taxonomies where available with LLM-assisted generation, structural constraints, and validation otherwise. The resulting fixed hierarchy supplies parent targets, parent-class entailment relations, and inference-time parent indices.

  \item Beyond classification accuracy, we evaluate six hierarchical metrics and analyze the learned hyperbolic geometry. The analyses suggest that the learned embeddings exhibit the radial ordering induced by the supplied hierarchy, placing parent classes closer to the origin than their child classes and associated images.
\end{itemize}

\section{Related work}

\subsection{Prompt Learning}

\paragraph{Prompt Learning for Vision-Language Models.}

The success of CLIP-style VLMs \citep{radford2021learning} established prompting as a flexible interface for downstream adaptation, complementing fully supervised fine-tuning with task-specific textual templates. Prompt learning extends this idea by replacing hand-written templates with continuous, optimizable tokens. CoOp \citep{zhou2022learning} introduces learnable context vectors that are concatenated with class names and passed through the frozen text encoder. These vectors are optimized end-to-end with a cross-entropy objective. CoCoOp \citep{zhou2022conditional} makes the context vectors instance-conditional through a lightweight meta-network, mitigating overfitting to base classes. Subsequent work has refined this paradigm along complementary axes. MaPLe \citep{khattak2023maple} introduces joint vision-language prompting with a learnable coupling between the two branches. PromptSRC \citep{khattak2023self} improves generalization through self-regularization, prompt self-ensembling, and textual diversity. CoPrompt \citep{roy2024consistency} adds a consistency constraint between learned prompts and the frozen CLIP backbone, supplemented by output adapters. More recent efforts have explored interpretable \citep{du2024ipointerpretablepromptoptimization} and structured \citep{zheng2025hierarchical} variants of prompt optimization.

Most prompt learning methods are developed with Euclidean CLIP representations and predict over a flat vocabulary of class labels. Our method is complementary to these approaches: it preserves the underlying class prompt learner while adding a separate parent prompt learner and hierarchical supervision in a shared hyperbolic representation space. Because the framework does not depend on a particular prompt parameterization, it can be instantiated with global, image-conditional, or multimodal prompt learners.

\paragraph{Hierarchy in Prompt Learning.}

A small but growing line of work considers structured prompting beyond a flat class vocabulary. MaPLe \citep{khattak2023maple} couples textual and visual prompts, while HiCroPL \citep{zheng2025hierarchical} introduces bidirectional knowledge flow across the text and image branches through layer-specific proxies and direction-switched mappers. These methods impose structure on the interaction between modalities or network layers, rather than modeling explicit taxonomic relations among class labels.

HGCLIP \citep{xia2025hgclip} is more closely related to taxonomic prompt learning. It adapts CLIP to hierarchical image classification by combining hierarchical prompts with a class hierarchy graph that propagates structural information into text and image features. Our approach differs in how it represents and uses this structure. Rather than introducing a graph encoder, we use a fixed direct parent-class mapping to train parallel class and parent prompt learners in a shared hyperbolic space. Parent-class relations are regularized with entailment cones and incorporated at inference through parent-feedback logit fusion.

\subsection{Hyperbolic Deep Learning}

\paragraph{Hyperbolic Representation Learning for Vision.}

Hyperbolic representation learning provides a natural geometric framework for data with hierarchical structure \citep{nickel2017poincare, nickel2018learning, sala2018representation}. Hyperbolic neural networks and entailment cones \citep{ganea2018cone, ganea2018hyperbolic} enable hierarchical relations to be represented through manifold-aware operations and directional "is-a" constraints. These ideas have been incorporated into visual recognition models, including hyperbolic vision transformers \citep{ermolov2022hyperbolic, fein2024hvt}, prototype-based classifiers \citep{khrulkov2020hyperbolic}, and more recently hyperbolic approaches to continual test-time and open-vocabulary semantic segmentation \citep{gole2026hyperbolic, truong2026semantic}.

A related line of work uses known taxonomies to structure hyperbolic class representations. \cite{long2020searching} and \cite{dhall2020hierarchical} encode action and image hierarchies with entailment cones, while later work considers hierarchical or boundary-based prototypes for classification and segmentation \citep{liu2020hyperbolic, ghadimi2021hyperbolic, ghadimi2022hyperbolic}. Complementarily, \cite{wang2025learning} study the learning of multi-level visual hierarchies in hyperbolic space for image retrieval. 

Together, these methods motivate hyperbolic geometry when semantic class structure is available. Our setting differs in that the learnable objects are prompt representations of labels rather than conventional class prototypes, and the hierarchy is modeled through entailment relations between learned parent and class embeddings. We retain the frozen hyperbolic vision-language backbone and use learned parent representations for both training-time supervision and inference-time logit fusion.

\paragraph{Hyperbolic Vision-Language Models.}

Hyperbolic geometry has recently been incorporated into joint vision-language representation learning. MERU \citep{desai2023hyperbolic} maps image and text representations to the Lorentz model and combines hyperbolic contrastive learning with an entailment objective. HyCoCLIP \citep{pal2024compositional} extends this formulation with compositional supervision over image regions and textual concepts. Other work has explored hyperbolic vision-language modeling for large-scale multimodal generation, safety-aware learning, fully hyperbolic encoders, and hierarchy-aware retrieval \citep{mandica2024hyperbolic, poppi2025hyperbolic, he2025hypercore, srivastava2025hypervlm}. PHyCLIP \citep{yoshikawa2026phyclip} further models hierarchy and compositionality in a product of hyperbolic factors.

These methods primarily introduce hyperbolic geometry through the representation space, encoder architecture, or pretraining objective. In contrast, we study hierarchical prompt learning for downstream adaptation with frozen hyperbolic VLM encoders. Our framework adds a parent prompt learner alongside the class prompt learner and uses the resulting parent representations for parent-level supervision, entailment-cone regularization, and inference-time logit fusion. We instantiate this framework with both MERU and HyCoCLIP.

HPL-CLIP \citep{peng2026hpl} more directly studies hyperbolic prompt learning, but for CLIP-based zero-shot anomaly detection through multi-level visual feature alignment. \cite{chen2024hyperbolic} similarly investigate hyperbolic prompting for frozen language models by projecting prefix prompts and token representations into the Poincar\'e disk. Unlike these settings, our method uses a supplied semantic parent-class hierarchy for hierarchical adaptation of frozen hyperbolic VLMs.

\section{Methodology}

\subsection{Preliminaries}
\label{sec:preliminaries}

We use the Lorentz (hyperboloid) model of hyperbolic space, which provides stable closed-form expressions for hyperbolic distance and mappings from tangent spaces to the manifold \citep{nickel2018learning}. We denote the curvature by $-\kappa$, where $\kappa>0$.

\subsubsection{The Lorentz Model}
\label{sec:lorentz-model}

The $n$-dimensional Lorentz model is the upper sheet of a hyperboloid in $\mathbb{R}^{n+1}$:
\begin{equation}
\mathbb{L}^n =
\left\{
\mathbf{p}=(p_0,\tilde{\mathbf{p}})\in\mathbb{R}^{n+1} :
\langle \mathbf{p},\mathbf{p}\rangle_{\mathbb{L}}=-\tfrac{1}{\kappa},
\; p_0>0
\right\},
\label{eq:lorentz-model}
\end{equation}
where $\tilde{\mathbf{p}}\in\mathbb{R}^n$ denotes the spatial coordinates and
\begin{equation}
\langle \mathbf{p},\mathbf{q}\rangle_{\mathbb{L}}
= -p_0q_0+\langle\tilde{\mathbf{p}},\tilde{\mathbf{q}}\rangle_{\mathbb{E}}
\label{eq:lorentzian-inner-product}
\end{equation}
is the Lorentzian inner product. The origin is
$\mathbf{0}=(\sqrt{1/\kappa},0,\ldots,0)^\top$. Hyperbolic space provides a geometric inductive bias in which general concepts can be represented nearer the origin and specific concepts farther away.

The geodesic distance between $\mathbf{p},\mathbf{q}\in\mathbb{L}^n$ is
\begin{equation}
d_{\mathbb{L}}(\mathbf{p},\mathbf{q})
=
\sqrt{\tfrac{1}{\kappa}}\,
\cosh^{-1}\!\left(
-\kappa\langle\mathbf{p},\mathbf{q}\rangle_{\mathbb{L}}
\right).
\label{eq:lorentzian-distance}
\end{equation}
We use this distance to construct hyperbolic classification logits.

\subsubsection{Exponential Map}
\label{sec:exp-map}

Euclidean encoder outputs are lifted to $\mathbb{L}^n$ through the exponential map. For $\mathbf{v}\in T_{\mathbf{p}}\mathbb{L}^n$,
\begin{equation}
\exp_{\mathbf{p}}^\kappa(\mathbf{v})
=
\cosh\!\bigl(\sqrt{\kappa}\|\mathbf{v}\|_{\mathbb{L}}\bigr)\mathbf{p}
+
\frac{\sinh\!\bigl(\sqrt{\kappa}\|\mathbf{v}\|_{\mathbb{L}}\bigr)}
{\sqrt{\kappa}\|\mathbf{v}\|_{\mathbb{L}}}\mathbf{v},
\label{eq:exp-map}
\end{equation}
where $\|\mathbf{v}\|_{\mathbb{L}}
=
\sqrt{\langle\mathbf{v},\mathbf{v}\rangle_{\mathbb{L}}}$.
In practice, we apply $\exp_{\mathbf{0}}^\kappa$ to image, class, and parent features, treating the Euclidean encoder outputs as tangent vectors at the origin \citep{khrulkov2020hyperbolic}.

\subsubsection{Entailment Cones}
\label{sec:entailment-cones}

Entailment cones encode a partial-order inductive bias in hyperbolic space: a candidate descendant embedding $\mathbf{p}$ is geometrically consistent with an ancestor embedding $\mathbf{q}$ when it lies inside the cone rooted at $\mathbf{q}$ \citep{ganea2018cone}. The cone half-aperture is
\begin{equation}
\omega(\mathbf{q})
=
\sin^{-1}\!\left(
\frac{2K}{\sqrt{\kappa}\|\tilde{\mathbf{q}}\|}
\right),
\label{eq:half-aperture}
\end{equation}
where $K=0.1$ follows \citet{ganea2018cone}. This expression is defined for $\|\tilde{\mathbf{q}}\|\geq 2K/\sqrt{\kappa}$. Under this parameterization, embeddings nearer the origin have wider cones, whereas embeddings farther from the origin have narrower cones.

For candidate child $\mathbf{p}$ and parent $\mathbf{q}$, the exterior angle is
\begin{equation}
\phi(\mathbf{p},\mathbf{q})
=
\cos^{-1}\!\left(
\frac{
p_0+q_0\kappa\langle\mathbf{p},\mathbf{q}\rangle_{\mathbb{L}}
}{
\|\tilde{\mathbf{q}}\|
\sqrt{
\left(\kappa\langle\mathbf{p},\mathbf{q}\rangle_{\mathbb{L}}\right)^2-1
}
}
\right).
\label{eq:exterior-angle}
\end{equation}
The relation $\phi(\mathbf{p},\mathbf{q})\leq\omega(\mathbf{q})$ indicates that $\mathbf{p}$ lies inside $\mathbf{q}$'s cone and is therefore geometrically consistent with being its descendant. Section~\ref{sec:losses} uses this relation to regularize parent-class, class-image, and parent-image relations.

\subsection{Method}
\label{sec:method}

We introduce a hyperbolic hierarchical prompt learning plug-in that augments an existing prompt learner with parent-level supervision and hierarchical geometry while preserving class-level prediction. Given a fixed parent-class hierarchy, the plug-in adds a parallel parent prompt learner and a hierarchical training and inference procedure that combines hyperbolic class and parent classification, entailment-based geometric regularization, and parent-feedback logit fusion. The hierarchy is constructed offline and remains fixed throughout training and inference; its construction procedure is described in Section~\ref{sec:hierarchy-construction}.

The framework is agnostic to the parameterization of the underlying prompt learner. Applied to CoOp~\citep{zhou2022learning}, it yields \textbf{HyPLO}; applied to CoCoOp~\citep{zhou2022conditional}, it yields \textbf{CoHyPLO}; and applied to MaPLe~\citep{khattak2023maple}, it yields \textbf{MaHyPLO}. All three instantiations share the same fixed hierarchy, hyperbolic representation space, hierarchical loss framework, and inference rule. They differ in their prompt parameterization and in the entailment terms activated by default.

We first describe the prompt learning architecture, then the training objective, hierarchy-aware inference and fixed hierarchy.

% -----------------------------------------------------------------------------
\subsubsection{Architecture}
\label{sec:architecture}

We assume a hyperbolic CLIP-style backbone with frozen image and text encoders, $E_I$ and $E_T$. Given an image $x$, the image encoder produces an image feature $\mathbf{f}_{\mathrm{img}} = E_I(x)$. A prompt learner constructs text prompts by concatenating learnable context tokens with a label-name embedding and passes the resulting prompt through $E_T$ to obtain a label text feature. In CoOp, the context tokens are global and shared across images. In CoCoOp, they are adapted to each image by an offset predicted from $\mathbf{f}_{\mathrm{img}}$ by a lightweight meta-network. In MaPLe, text-side prompts are coupled with visual prompts that adapt the image encoder.

\begin{figure}[t]
    \centering
    \includegraphics[width=0.9\textwidth]{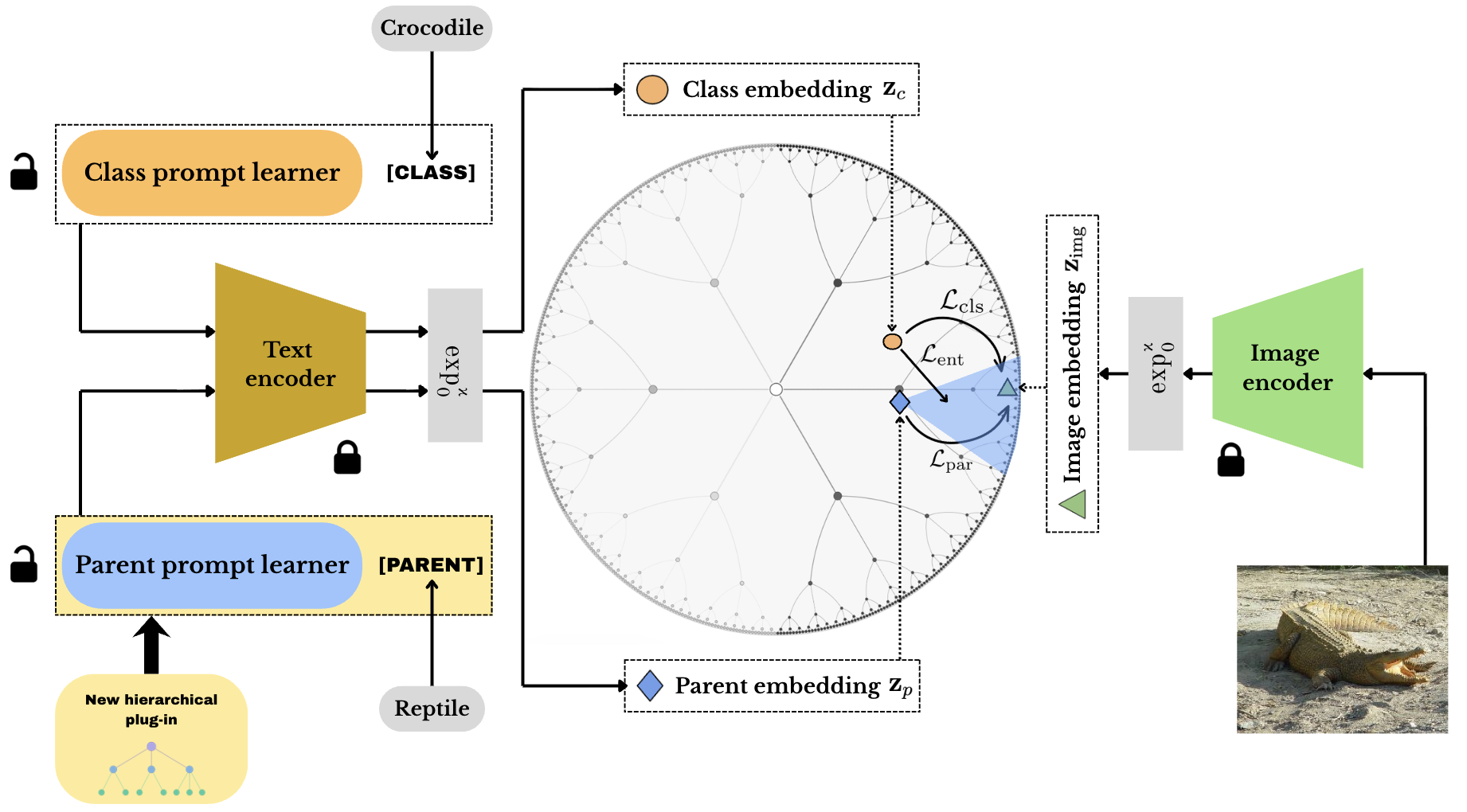}
    \caption{Overview of the hyperbolic hierarchical prompt learning plug-in. A learnable \textcolor{blue}{parent prompt learner} is added alongside the \textcolor{orange}{class prompt learner}. The fixed dataset hierarchy provides the parent vocabulary $\mathcal{P}$ and class-to-parent mapping $\pi$. Frozen text and image encoders produce features that are lifted to the Lorentz manifold through $\exp_{\mathbf{0}}^{\kappa}$, yielding class, parent, and image embeddings $\mathbf{z}_{c}$, $\mathbf{z}_{p}$, and $\mathbf{z}_{\mathrm{img}}$. Locks denote frozen encoder parameters; prompt learning modules are optimized. The framework yields HyPLO, CoHyPLO, and MaHyPLO when instantiated with CoOp, CoCoOp, and MaPLe respectively. In MaHyPLO, visual prompts from the class learner adapt the image encoder output, whereas parent predictions use parent text embeddings and the resulting shared image embedding. The Poincar\'e disk is shown only for visualization of hierarchical relations.}
    \label{fig:arch}
\end{figure}

Our method adds a separate parent prompt learner over the parent vocabulary $\mathcal{P}$ induced by the fixed class-to-parent mapping $\pi$ (Figure~\ref{fig:arch}). The original learner produces one class embedding for each class $c \in \mathcal{C}$, while the parent learner produces one parent embedding for each parent $p \in \mathcal{P}$. The two learners use separate prompt parameters and are coupled through the hierarchical training objective. Both class and parent embeddings are compared with the same image embedding.

For HyPLO, the parent learner consists of separate global text context tokens. For CoHyPLO, it additionally has a separate meta-network that produces image-conditional parent context tokens. For MaHyPLO, the parent learner is instantiated using MaPLe's prompt parameterization; however, only its text prompts are used to compute parent embeddings. The image embedding is produced by the class learner's visual prompt pathway, and MaHyPLO does not perform a separate parent-conditioned image encoder forward pass.

\paragraph{Hyperbolic projection.}
Following Section~\ref{sec:exp-map}, we lift the image, class, and parent features to the Lorentz manifold through the exponential map at the origin. This yields the hyperbolic image embedding $\mathbf{z}_{\mathrm{img}}$, class embeddings $\{\mathbf{z}_{c}\}_{c \in \mathcal{C}}$, and parent embeddings $\{\mathbf{z}_{p}\}_{p \in \mathcal{P}}$. We use the curvature $\kappa$ inherited from the HyCoCLIP backbone~\citep{pal2024compositional}, and perform all subsequent classification and geometric operations in $\mathbb{L}^{n}$. We use HyCoCLIP’s numerically stabilized Lorentz operations.

% -----------------------------------------------------------------------------
\subsubsection{Hierarchical Training Objective}
\label{sec:losses}

\begin{figure}[t]
    \centering
    \includegraphics[width=0.9\textwidth]{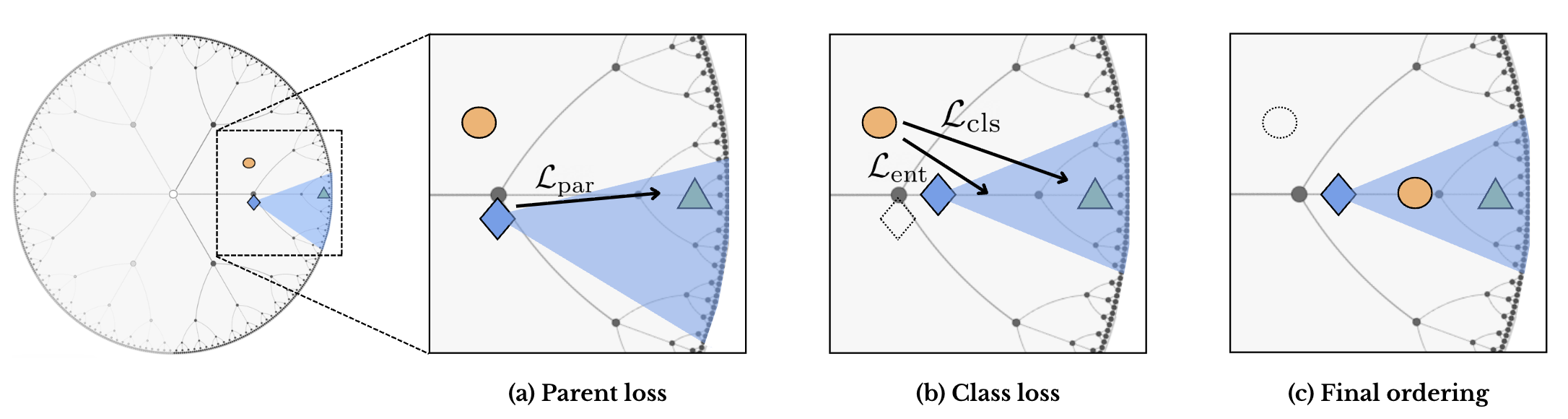}
    \caption{Illustration of the restricted hierarchical objective used by HyPLO and CoHyPLO. The \textcolor{blue}{parent embedding} $\mathbf{z}_{p}$ and \textcolor{orange}{class embedding} $\mathbf{z}_{c}$ are trainable through their respective prompt learners, whereas the \textcolor{green}{image embedding} $\mathbf{z}_{\mathrm{img}}$ is generated by the frozen backbone. Parent and class cross-entropy increase relative compatibility with the ground-truth image, while parent-to-class entailment encourages the class embedding to lie inside the parent's entailment cone. MaHyPLO additionally applies class-to-image and parent-to-image entailment regularization, as defined below.}
    \label{fig:embeddings_visualized}
\end{figure}

For a labeled example $(x,y)$, the objective combines class-level classification, parent-level classification, and entailment regularization along the ground-truth hierarchy path $\pi(y) \rightarrow y \rightarrow x$. The class and parent classification losses compare every image in a mini-batch with the complete class and parent vocabularies respectively. The entailment terms are evaluated on the ground-truth parent-class-image relations for each mini-batch example and then averaged over the mini-batch. The full objective is depicted in Figure~\ref{fig:embeddings_visualized}.

\paragraph{Class cross-entropy.}
For an image with ground-truth class $y$, we score every class $c \in \mathcal{C}$ by its scaled negative Lorentzian distance from the image embedding:
$
s_{\mathrm{cls}}(c)
=
-\gamma \,
d_{\mathbb{L}}
\left(
\mathbf{z}_{\mathrm{img}},
\mathbf{z}_{c}
\right),
$
where $\gamma = \exp(\ell)$ is the frozen exponentiated logit scale.
\begin{equation}
\mathcal{L}_{\mathrm{cls}}
= -\log \frac{\exp\!\big(s_{\mathrm{cls}}(y)\big)}
              {\sum_{c \in \mathcal{C}} \exp\!\big(s_{\mathrm{cls}}(c)\big)}.
\label{eq:cls-loss}
\end{equation}
This loss increases the relative similarity of the image to its ground-truth class while decreasing its similarity to incorrect class labels.

\paragraph{Parent cross-entropy.}
Each class $c$ has a unique direct parent $\pi(c) \in \mathcal{P}$ specified by the fixed hierarchy. We score each parent $p \in \mathcal{P}$ using
$
s_{\mathrm{par}}(p)
=
-\gamma \,
d_{\mathbb{L}}
\left(
\mathbf{z}_{\mathrm{img}},
\mathbf{z}_{p}
\right),
$
and supervise the parent learner with
\begin{equation}
\mathcal{L}_{\mathrm{par}}
= -\log \frac{\exp\!\big(s_{\mathrm{par}}(\pi(y))\big)}
              {\sum_{p \in \mathcal{P}} \exp\!\big(s_{\mathrm{par}}(p)\big)}.
\label{eq:par-loss}
\end{equation}
This objective aligns image embeddings with their ground-truth parent embeddings relative to competing parents. It also trains the parent logits used by the inference rule in Section~\ref{sec:inference}.

\paragraph{Entailment regularizer.}
Classification alone does not explicitly constrain the parent-class geometry. We therefore regularize ground-truth ancestor-descendant relations using the entailment cones from Section~\ref{sec:entailment-cones}. For a descendant embedding $\mathbf{z}_{\mathrm{desc}}$ and ancestor embedding $\mathbf{z}_{\mathrm{anc}}$, we define
\begin{equation}
\mathcal{L}_{\mathrm{ent}}^{\mathrm{anc\rightarrow desc}}
=
\max\!\left(
0,\,
\phi\!\left(
\mathbf{z}_{\mathrm{desc}},
\mathbf{z}_{\mathrm{anc}}
\right)
-
\eta\,\omega\!\left(
\mathbf{z}_{\mathrm{anc}}
\right)
\right).
\label{eq:entailment-residual}
\end{equation}
The residual is zero when the descendant lies within the $\eta$-scaled cone of the ancestor. Thus, $\eta$ controls the strictness of the constraint: smaller values impose a narrower admissible cone.

For each example $(x,y)$, we instantiate this loss for the class-image, parent-class, and parent-image relations by setting the descendant and ancestor embeddings to $(\mathbf{z}_{\mathrm{img}}, \mathbf{z}_{y})$, $(\mathbf{z}_{y}, \mathbf{z}_{\pi(y)})$, and $(\mathbf{z}_{\mathrm{img}}, \mathbf{z}_{\pi(y)})$ respectively. We use fixed aperture factors $\eta_{\mathrm{cls\rightarrow img}}=0.3$, $\eta_{\mathrm{par\rightarrow cls}}=0.6$, and $\eta_{\mathrm{par\rightarrow img}}=0.6$. The factors reflect the intended specificity ordering: the same wider scaled parent cone is used for both parent-class and parent-image relations, while the class-image relation uses a stricter scaled cone. We follow HyCoCLIP's relative aperture ratio of $2{:}1$ between the broader and narrower constraints.

We regularize observed hierarchy-consistent relations only. Non-descendant pairs are distinguished indirectly through the class-level and parent-level cross-entropy losses rather than through an explicit cone exclusion loss.

\paragraph{Composite objective.}
The training objective is
\begin{equation}
\mathcal{L}
=
\mathcal{L}_{\mathrm{cls}}
+
w_{\mathrm{par}}\mathcal{L}_{\mathrm{par}}
+
\underbrace{
w_{\mathrm{cls\rightarrow img}}\mathcal{L}_{\mathrm{ent}}^{\mathrm{cls\rightarrow img}}
+
w_{\mathrm{par\rightarrow cls}}\mathcal{L}_{\mathrm{ent}}^{\mathrm{par\rightarrow cls}}
+
w_{\mathrm{par\rightarrow img}}\mathcal{L}_{\mathrm{ent}}^{\mathrm{par\rightarrow img}}
}_{\mathcal{L}_{\mathrm{ent}}}.
\label{eq:total-loss}
\end{equation}

HyPLO and CoHyPLO activate only parent-to-class entailment, with entailment weights $(0, 0.2, 0)$ for class-to-image, parent-to-class, and parent-to-image regularization, respectively. In these variants, image embeddings are produced by a frozen visual encoder. We therefore omit image-related entailment terms to avoid forcing the trainable class and parent prompts to conform to potentially imperfect fixed image representations. MaHyPLO, in contrast, uses all three terms, with weights $(0.1, 0.2, 0.1)$, because its visual prompt pathway can jointly adapt the image embedding and text-side prompt representations. We use $w_{\mathrm{par}}=0.5$ by default and average all loss terms over the mini-batch.

HyPLO and CoHyPLO optimize separate class and parent text-prompt learners while keeping the image embeddings fixed. Their class and parent classification losses update the corresponding prompt learner, and parent-to-class entailment updates both learners through their text embeddings. In MaHyPLO, the class learner additionally contains visual prompts that produce the shared image embedding. Consequently, the class classification loss and all image-related terms update the class learner's visual prompt pathway. The parent classification loss and parent-to-image entailment also update this pathway through their dependence on the shared image embedding, while updating the parent text prompts. Although the MaHyPLO parent learner uses MaPLe's prompt parameterization, its visual prompts are not used to encode images.

% -----------------------------------------------------------------------------
\subsubsection{Hierarchical Inference}
\label{sec:inference}

At inference time, predictions remain class-level, while the parent learner provides an auxiliary parent-level signal. Let $\boldsymbol{\ell}_{\mathrm{cls}}$ and $\boldsymbol{\ell}_{\mathrm{par}}$ denote the scaled negative hyperbolic distance logits over classes and parents, respectively. We broadcast each parent logit to all of its child classes and compute
\begin{equation}
s_{\mathrm{final}}(c)
= \big[\boldsymbol{\ell}_{\mathrm{cls}}\big]_c
+ \lambda_{\mathrm{par}}\,
  \big[\boldsymbol{\ell}_{\mathrm{par}}\big]_{\pi(c)},
\;
\hat{y} = \arg\max_c\, s_{\mathrm{final}}(c).
\label{eq:final-score}
\end{equation}
The weight $\lambda_{\mathrm{par}}$ controls the influence of this parent-level signal relative to the class-level logits. Sibling classes receive the same parent contribution, so fusion does not directly distinguish siblings. Instead, it can resolve confusions between classes assigned to different parents when the class-level evidence is ambiguous. Setting $\lambda_{\mathrm{par}} = 0$ recovers class-only inference.

\begin{algorithm}[H]
\caption{Hierarchy construction for a class set $\mathcal{C}$.}
\label{alg:hierarchy}
\begin{algorithmic}[1]
\Require class names $\mathcal{C}$; optional native taxonomy
         $\mathcal{T}$; structural constraints $\mathcal{S}$
\Ensure validated hierarchy $\mathcal{H} = (\mathcal{P}, \pi)$
\If{$\mathcal{T}$ is available}
    \State $\mathcal{H} \gets \mathcal{T}$
    \State resolve any multi-parent leaf to its most specific parent
\Else
    \State prompt an LLM for $\mathcal{H}$ given $\mathcal{C}$ and
           $\mathcal{S}$
\EndIf
\Repeat
    \State $V \gets$ constraints in $\mathcal{S}$ violated by $\mathcal{H}$
    \ForAll{violation $v \in V$}
        \If{$v$ is a singleton parent}
            \State merge it into a related parent
        \ElsIf{$v$ is an oversized parent}
            \State split it along natural visual sub-groupings
        \ElsIf{$v$ is a label collision}
            \State rename the offending node
        \ElsIf{$v$ is a verbose parent name}
            \State shorten it to a CLIP-compatible noun phrase
        \EndIf
    \EndFor
\Until{$V = \varnothing$}
\State \Return $\mathcal{H}$
\end{algorithmic}
\end{algorithm}

% -----------------------------------------------------------------------------
\subsection{Hierarchy Construction}
\label{sec:hierarchy-construction}

The preceding architecture, objective, and inference rule require a fixed parent vocabulary $\mathcal{P}$ and a unique direct-parent mapping $\pi : \mathcal{C} \rightarrow \mathcal{P}$. We construct these structures offline before training and keep them fixed throughout optimization and inference. The hierarchy supplies parent targets for parent-level classification, parent-class relations for entailment regularization, and parent indices for logit fusion.

We require each hierarchy to provide a unique direct parent for every class, to avoid singleton parent groups where possible, and to use distinct, concise parent labels that can be inserted naturally into CLIP-style prompts. These conditions are intended to ensure that parent supervision is informative and that the class-to-parent lookup is unambiguous. Although a source taxonomy may contain multiple levels, the proposed method uses only the direct parent assignment $\pi(c)$ for each class $c$.

For datasets with an appropriate native taxonomy, we derive direct parent assignments from that taxonomy and post-process them to satisfy the above criteria. For datasets without a suitable taxonomy, we use an LLM-assisted construction procedure followed by validation and refinement. The LLM, Claude Opus 4.7, is provided with the class vocabulary and structural criteria, after which the proposed hierarchy is checked and corrected as necessary. Algorithm~\ref{alg:hierarchy} summarizes this offline procedure.

The validation criteria are directly checkable from the resulting hierarchy. The final parent vocabularies and class-to-parent mappings are reported in Appendix~\ref{app:hierarchies}. The hierarchy construction procedure is not a learned component of the model; it is an offline annotation protocol that supplies the fixed supervision structure used by the proposed framework.

\section{Experiments}
\label{sec:experiments}

\paragraph{Setup.}
We evaluate on the standard 11-dataset benchmark introduced by CoOp \citep{zhou2022learning}: ImageNet \citep{deng2009imagenet}, Caltech101 \citep{fei2004learning}, OxfordPets \citep{parkhi2012cats}, StanfordCars \citep{krause20133d}, Flowers102 \citep{nilsback2008automated}, Food101 \citep{bossard2014food}, FGVCAircraft \citep{maji2013fine}, SUN397 \citep{xiao2010sun}, DTD \citep{cimpoi2014describing}, EuroSAT \citep{helber2019eurosat}, and UCF101 \citep{soomro2012ucf101}. We consider three evaluation settings following CoCoOp \citep{zhou2022conditional}: \emph{base-to-new generalization}, \emph{cross-dataset transfer}, and \emph{domain generalization}. In base-to-new generalization, prompts are trained on the base class split of each dataset and evaluated on both base and novel classes; we report base class accuracy, novel class accuracy, and their harmonic mean (H). For cross-dataset transfer, prompts are trained on ImageNet and evaluated without further adaptation on the remaining ten datasets. For domain generalization, the same ImageNet-trained prompt learners are evaluated on ImageNetV2, ImageNet-Sketch, ImageNet-A, and ImageNet-R. All results are averaged over three random seeds and their standard deviations are shown in Appendix~\ref{app:stds}.

Unless stated otherwise, we use frozen HyCoCLIP \citep{pal2024compositional} with a ViT-B/16 backbone in the 16-shot setting. HyPLO, CoHyPLO, and MaHyPLO extend CoOp \citep{zhou2022learning}, CoCoOp \citep{zhou2022conditional}, and MaPLe \citep{khattak2023maple}, respectively, using the corresponding training protocols. The offline parent-class hierarchies remain fixed throughout training and inference; ImageNet uses its native WordNet hierarchy. We use a parent-fusion weight of $\lambda_{\mathrm{par}} = 0.25$, selected in the ablation in Section~\ref{sec:infer_fusion}.

We compare with zero-shot HyCoCLIP, its hierarchical template variant HyCoCLIP$^{\dagger}$ (explained in Appendix~\ref{app:hier_hyco}), and the prompt learners extended by our method. In addition to classification accuracy, we report six hierarchical metrics: tree-induced error (TIE), lowest common ancestor distance (LCA), Jaccard similarity, hierarchical F1, sibling accuracy, and cousin accuracy (explained in Appendix~\ref{app:hier_extended}). We report these metrics for base-to-new generalization, first averaging results over three random seeds within each dataset and then macro-averaging across all 11 datasets.

\begin{table}
\centering
\scriptsize
\setlength{\tabcolsep}{4pt}
\renewcommand{\arraystretch}{1.05}

% ---------- Row 1 ----------
\begin{subtable}{0.33\textwidth}
\centering
\caption{ImageNet}
\begin{tabular}{lccc}
\toprule
\textbf{Method} & \textbf{Base} & \textbf{New} & \textbf{H} \\
\midrule
HyCoCLIP           & 46.40 & 52.10 & 49.09 \\
\midrule
% HyCoCLIP$^\dagger$ & 46.2 & \textbf{52.1} & 48.98 \\
+ CoOp               & 48.90 & 49.17 & 49.03 \\
\rowcolor{gray!10}\textbf{+ HyPLO}   & \textbf{49.07} & \textbf{50.47} & \textbf{49.76} \\
\midrule
+ CoCoOp             & \textbf{48.90} & 50.47 & 49.67 \\
\rowcolor{gray!10}\textbf{+ CoHyPLO} & 48.47 & \textbf{52.03} & \textbf{50.19} \\
\midrule
+ MaPLe  & 50.27 & 50.97 & 50.62 \\
\rowcolor{gray!10}\textbf{+ MaHyPLO} & \textbf{50.50} & \underline{\textbf{52.33}} & \textbf{51.40} \\
\midrule
+ PromptSRC               & \underline{52.77} & 51.30 & \underline{52.02} \\
\midrule
+ CoPrompt               & 50.17 & 50.90 & 50.53 \\
\bottomrule
\end{tabular}
\end{subtable}%
\begin{subtable}{0.33\textwidth}
\centering
\caption{Caltech101}
\begin{tabular}{lccc}
\toprule
\textbf{Method} & \textbf{Base} & \textbf{New} & \textbf{H} \\
\midrule
HyCoCLIP           & 92.90 & \underline{91.00} & 91.94 \\
% HyCoCLIP$^\dagger$ & 93.2 & 89.7 & 91.42 \\
\midrule
+ CoOp               & \textbf{95.53} & 83.07 & 88.87 \\
\rowcolor{gray!10}\textbf{+ HyPLO}   & 95.37 & \textbf{84.20} & \textbf{89.44} \\
\midrule
+ CoCoOp             & 95.10 & 88.13 & 91.43 \\
\rowcolor{gray!10}\textbf{+ CoHyPLO} & \textbf{95.23} & \textbf{89.20} & \textbf{92.12} \\
\midrule
+ MaPLe   & 95.57 & \textbf{90.40} & 92.91 \\
\rowcolor{gray!10}\textbf{+ MaHyPLO} & \textbf{95.83} & 90.30 & \underline{\textbf{92.98}} \\
\midrule
+ PromptSRC               & \underline{96.13} & 85.30 & 90.39 \\
\midrule
+ CoPrompt               & 96.03 & 88.47 & 92.10 \\
\bottomrule
\end{tabular}
\end{subtable}%
\begin{subtable}{0.33\textwidth}
\centering
\caption{OxfordPets}
\begin{tabular}{lccc}
\toprule
\textbf{Method} & \textbf{Base} & \textbf{New} & \textbf{H} \\
\midrule
HyCoCLIP           & 57.50 & \underline{74.30} & 64.83 \\
% HyCoCLIP$^\dagger$ & 60.4 & \textbf{74.9} & 66.87 \\
\midrule
+ CoOp               & 76.60 & \textbf{61.83} & 68.43 \\
\rowcolor{gray!10}\textbf{+ HyPLO}   & \textbf{78.63} & \textbf{61.83} & \textbf{69.23} \\
\midrule
+ CoCoOp             & 75.8 & 66.77 & 70.95 \\
\rowcolor{gray!10}\textbf{+ CoHyPLO} & \textbf{76.40} & \textbf{68.47} & \textbf{72.22} \\
\midrule
+ MaPLe   & 74.60 & 70.97 & 72.74 \\
\rowcolor{gray!10}\textbf{+ MaHyPLO} & \textbf{77.73} & \textbf{73.13} & \textbf{75.36} \\
\midrule
+ PromptSRC               & \underline{79.90} & 71.33 & \underline{75.37} \\
\midrule
+ CoPrompt               & 76.93 & 69.07 & 72.79 \\
\bottomrule
\end{tabular}
\end{subtable}

\vspace{0.8em}

% ---------- Row 2 ----------
\begin{subtable}{0.33\textwidth}
\centering
\caption{StanfordCars}
\begin{tabular}{lccc}
\toprule
\textbf{Method} & \textbf{Base} & \textbf{New} & \textbf{H} \\
\midrule
HyCoCLIP           & 14.40 & 19.50 & 16.57 \\
% HyCoCLIP$^\dagger$ & 15.3 & 20.3 & 17.45 \\
\midrule
+ CoOp               & \textbf{27.47} & 15.60 & 19.90 \\
\rowcolor{gray!10}\textbf{+ HyPLO}   & 26.20 & \textbf{16.90} & \textbf{20.55} \\
\midrule
+ CoCoOp             & \textbf{20.57} & 20.10 & 20.33 \\
\rowcolor{gray!10}\textbf{+ CoHyPLO} & 19.83 & \underline{\textbf{20.97}} & \textbf{20.38} \\
\midrule
+ MaPLe   & 22.77 & 19.73 & 21.14 \\
\rowcolor{gray!10}\textbf{+ MaHyPLO} & \textbf{23.70} & \textbf{20.53} & \textbf{22.00} \\
\midrule
+ PromptSRC               & \underline{31.07} & 20.93 & \underline{25.01} \\
\midrule
+ CoPrompt               & 25.37 & \underline{20.97} & 22.96 \\
\bottomrule
\end{tabular}
\end{subtable}%
\begin{subtable}{0.33\textwidth}
\centering
\caption{Flowers102}
\begin{tabular}{lccc}
\toprule
\textbf{Method} & \textbf{Base} & \textbf{New} & \textbf{H} \\
\midrule
HyCoCLIP           & 36.80 & \underline{38.10} & 37.44 \\
% HyCoCLIP$^\dagger$ & 36.1 & 37.2 & 36.64 \\
\midrule
+ CoOp               & \textbf{78.80} & \textbf{22.00} & \textbf{34.40} \\
\rowcolor{gray!10}\textbf{+ HyPLO}   & 78.73 & 19.67 & 31.48 \\
\midrule
+ CoCoOp             & \textbf{59.63} & 26.77 & 36.96 \\
\rowcolor{gray!10}\textbf{+ CoHyPLO} & 59.03 & \textbf{29.70} & \textbf{39.52} \\
\midrule
+ MaPLe   & 63.97 & 32.20 & 42.84 \\
\rowcolor{gray!10}\textbf{+ MaHyPLO} & \textbf{65.80} & \textbf{34.80} & \underline{\textbf{45.52}} \\
\midrule
+ PromptSRC               & \underline{85.77} & 29.27 & 43.65 \\
\midrule
+ CoPrompt               & 70.27 & 27.37 & 39.40 \\
\bottomrule
\end{tabular}
\end{subtable}%
\begin{subtable}{0.33\textwidth}
\centering
\caption{Food101}
\begin{tabular}{lccc}
\toprule
\textbf{Method} & \textbf{Base} & \textbf{New} & \textbf{H} \\
\midrule
HyCoCLIP           & 66.90 & \underline{71.20} & 68.98 \\
% HyCoCLIP$^\dagger$ & 66.6 & 69.8 & 68.16 \\
\midrule
+ CoOp               & 72.20 & 60.47 & 65.82 \\
\rowcolor{gray!10}\textbf{+ HyPLO}   & \textbf{73.37} & \textbf{61.90} & \textbf{67.15} \\
\midrule
+ CoCoOp             & 72.30 & 66.43 & 69.24 \\
\rowcolor{gray!10}\textbf{+ CoHyPLO} & \textbf{72.60} & \textbf{69.50} & \textbf{71.02} \\
\midrule
+ MaPLe   & 73.20 & 70.27 & 71.71 \\
\rowcolor{gray!10}\textbf{+ MaHyPLO} & \textbf{73.57} & \textbf{71.00} & \underline{\textbf{72.26}} \\
\midrule
+ PromptSRC               & \underline{74.57} & 69.67 & 72.04 \\
\midrule
+ CoPrompt               & 73.20 & 69.77 & 71.44 \\
\bottomrule
\end{tabular}
\end{subtable}

\vspace{0.8em}

% ---------- Row 3 ----------
\begin{subtable}{0.33\textwidth}
\centering
\caption{FGVCAircraft}
\begin{tabular}{lccc}
\toprule
\textbf{Method} & \textbf{Base} & \textbf{New} & \textbf{H} \\
\midrule
HyCoCLIP           & 2.80 & 6.60 & 3.94 \\
% HyCoCLIP$^\dagger$ & 3.9 & 7.4 & 5.11 \\
\midrule
+ CoOp               & 14.20 & 5.57 & 8.00 \\
\rowcolor{gray!10}\textbf{+ HyPLO}   & \underline{\textbf{15.30}} & \textbf{6.30} & \textbf{8.93} \\
\midrule
+ CoCoOp             & \textbf{10.33} & 5.90 & 7.51 \\
\rowcolor{gray!10}\textbf{+ CoHyPLO} & \textbf{10.33} & \textbf{7.17} & \textbf{8.46} \\
\midrule
+ MaPLe   & 8.07 & 6.13 & 6.97 \\
\rowcolor{gray!10}\textbf{+ MaHyPLO} & \textbf{11.63} & \underline{\textbf{7.93}} & \underline{\textbf{9.43}} \\
\midrule
+ PromptSRC               & 12.73 & 5.30 & 7.48 \\
\midrule
+ CoPrompt               & 10.40 & 4.10 & 5.88 \\
\bottomrule
\end{tabular}
\end{subtable}%
\begin{subtable}{0.33\textwidth}
\centering
\caption{SUN397}
\begin{tabular}{lccc}
\toprule
\textbf{Method} & \textbf{Base} & \textbf{New} & \textbf{H} \\
\midrule
HyCoCLIP           & 64.10 & \underline{69.70} & 66.78 \\
% HyCoCLIP$^\dagger$ & 63.0 & 69.4 & 66.05 \\
\midrule
+ CoOp               & 71.43 & 61.47 & 66.08 \\
\rowcolor{gray!10}\textbf{+ HyPLO}   & \textbf{71.63} & \textbf{63.47} & \textbf{67.30} \\
\midrule
+ CoCoOp             & 68.90 & 66.33 & 67.59 \\
\rowcolor{gray!10}\textbf{+ CoHyPLO} & \textbf{69.57} & \textbf{68.00} & \textbf{68.78} \\
\midrule
+ MaPLe   & 70.30 & 67.07 & 68.65 \\
\rowcolor{gray!10}\textbf{+ MaHyPLO} & \textbf{71.00} & \textbf{68.50} & \textbf{69.73} \\
\midrule
+ PromptSRC               & \underline{73.13} & 68.73 & \underline{70.86} \\
\midrule
+ CoPrompt               & 70.37 & 66.90 & 68.59 \\
\bottomrule
\end{tabular}
\end{subtable}%
\begin{subtable}{0.33\textwidth}
\centering
\caption{DTD}
\begin{tabular}{lccc}
\toprule
\textbf{Method} & \textbf{Base} & \textbf{New} & \textbf{H} \\
\midrule
HyCoCLIP           & 32.90 & 37.40 & 35.01 \\
% HyCoCLIP$^\dagger$ & 31.1 & \textbf{41.5} & 35.56 \\
\midrule
+ CoOp               & 68.10 & \textbf{27.93} & \textbf{39.61} \\
\rowcolor{gray!10}\textbf{+ HyPLO}   & \textbf{70.3} & 27.57 & \textbf{39.61} \\
\midrule
+ CoCoOp             & 60.00 & 33.87 & 43.30 \\
\rowcolor{gray!10}\textbf{+ CoHyPLO} & \textbf{62.70} & \textbf{35.40} & \textbf{45.25} \\
\midrule
+ MaPLe   & 64.33 & \textbf{38.23} & 47.96 \\
\rowcolor{gray!10}\textbf{+ MaHyPLO} & \textbf{67.73} & 37.87 & \textbf{48.58} \\
\midrule
+ PromptSRC               & \underline{71.63} & 36.63 & 48.47 \\
\midrule
+ CoPrompt               & 68.83 & \underline{39.30} & \underline{50.03} \\
\bottomrule
\end{tabular}
\end{subtable}

\vspace{0.8em}

% ---------- Row 4 ----------
\begin{subtable}{0.33\textwidth}
\centering
\caption{EuroSAT}
\begin{tabular}{lccc}
\toprule
\textbf{Method} & \textbf{Base} & \textbf{New} & \textbf{H} \\
\midrule
HyCoCLIP           & 45.90 & 55.60 & 50.29 \\
% HyCoCLIP$^\dagger$ & 37.1 & \textbf{62.3} & 46.51 \\
\midrule
+ CoOp               & 91.07 & 46.60 & 61.65 \\
\rowcolor{gray!10}\textbf{+ HyPLO}   & \underline{\textbf{91.70}} & \textbf{51.83} & \textbf{66.23} \\
\midrule
+ CoCoOp             & 86.53 & 49.13 & 62.67 \\
\rowcolor{gray!10}\textbf{+ CoHyPLO} & \textbf{87.33} & \textbf{52.60} & \textbf{65.66} \\
\midrule
+ MaPLe   & 86.57 & 51.70 & 64.74 \\
\rowcolor{gray!10}\textbf{+ MaHyPLO} & \textbf{88.73} & \underline{\textbf{59.90}} & \underline{\textbf{71.52}} \\
\midrule
+ PromptSRC               & 87.23 & 43.87 & 58.38 \\
\midrule
+ CoPrompt               & 86.10 & 47.20 & 60.97 \\
\bottomrule
\end{tabular}
\end{subtable}%
\begin{subtable}{0.33\textwidth}
\centering
\caption{UCF101}
\begin{tabular}{lccc}
\toprule
\textbf{Method} & \textbf{Base} & \textbf{New} & \textbf{H} \\
\midrule
HyCoCLIP           & 55.00 & \underline{58.90} & 56.88 \\
% HyCoCLIP$^\dagger$ & 48.6 & 57.3 & 52.59 \\
\midrule
+ CoOp               & 72.20 & 46.63 & 56.66 \\
\rowcolor{gray!10}\textbf{+ HyPLO}   & \textbf{72.33} & \textbf{47.50} & \textbf{57.34} \\
\midrule
+ CoCoOp             & 65.17 & 52.37 & 58.07 \\
\rowcolor{gray!10}\textbf{+ CoHyPLO} & \textbf{65.57} & \textbf{54.27} & \textbf{59.39} \\
\midrule
+ MaPLe   & 67.03 & 53.20 & 59.32 \\
\rowcolor{gray!10}\textbf{+ MaHyPLO} & \textbf{69.00} & \textbf{57.87} & \textbf{62.95} \\
\midrule
+ PromptSRC               & \underline{73.67} & 55.90 & \underline{63.57} \\
\midrule
+ CoPrompt               & 69.77 & 52.23 & 59.74 \\
\bottomrule
\end{tabular}
\end{subtable}%
\begin{subtable}{0.33\textwidth}
\centering
\caption{Average}
\begin{tabular}{lccc}
\toprule
\textbf{Method} & \textbf{Base} & \textbf{New} & \textbf{H} \\
\midrule
HyCoCLIP           & 46.87 & \underline{52.22} & 49.40 \\
% HyCoCLIP$^\dagger$ & 45.6 & \textbf{52.9} & 48.98 \\
\midrule
+ CoOp               & 65.14 & 43.67 & 52.29 \\
\rowcolor{gray!10}\textbf{+ HyPLO}   & \textbf{65.69} & \textbf{44.69} & \textbf{53.19} \\
\midrule
+ CoCoOp             & 60.29 & 47.84 & 53.35 \\
\rowcolor{gray!10}\textbf{+ CoHyPLO} & \textbf{60.64} & \textbf{49.76} & \textbf{54.66} \\
\midrule
+ MaPLe   & 61.52 & 50.08 & 55.21 \\
\rowcolor{gray!10}\textbf{+ MaHyPLO} & \textbf{63.20} & \textbf{52.20} & \underline{\textbf{57.18}} \\
\midrule
+ PromptSRC               & \underline{67.15} & 48.93 & 56.61 \\
\midrule
+ CoPrompt               & 63.40 & 50.35 & 56.13 \\
\bottomrule
\end{tabular}
\end{subtable}

\caption{
Base-to-new generalization accuracy (\%) across 11 datasets. Each subtable reports base class accuracy (\textbf{Base}), novel class accuracy (\textbf{New}), and their harmonic mean (\textbf{H}). Results for each dataset are averaged over three random seeds. The \emph{Average} subtable reports a macro-average across datasets; \emph{Average} \textbf{H} is a harmonic mean of \emph{Average} \textbf{Base} and \emph{Average} \textbf{New}. HyPLO, CoHyPLO, and MaHyPLO are the hierarchical extensions of CoOp, CoCoOp, and MaPLe, respectively. \textbf{Bold} indicates the better result between each baseline and its hierarchical extension. \underline{Underlining} indicates the best result in each column across all methods.
}
\label{tab:b2n_all}
\end{table}

\subsection{Base-to-New Generalization}
\label{sec:base2new}

Table~\ref{tab:b2n_all} reports base-to-new generalization across the 11 datasets. The hierarchy-aware variants generally improve the prompt learners they extend. HyPLO improves the harmonic mean over CoOp on 9 of 11 datasets, while CoHyPLO and MaHyPLO improve over CoCoOp and MaPLe on all 11 datasets respectively. On the 11-dataset macro-average, HyPLO improves the harmonic mean from 52.29 to 53.19, and CoHyPLO improves it from 53.35 to 54.66. MaHyPLO achieves the strongest average result, improving MaPLe from 55.21 to 57.18.

The results preserve the characteristic trade-off between static and image-conditional prompting. HyPLO tends to favor base class accuracy, whereas CoHyPLO more consistently improves the balance between base and novel classes. This difference is reflected in their macro-averaged results: HyPLO improves base accuracy more strongly than novel class accuracy, while CoHyPLO yields a larger relative improvement on novel classes. MaHyPLO provides the strongest aggregate performance, improving both base and novel accuracy over MaPLe.

The hierarchy is most useful when it helps distinguish candidate classes from different parent groups. Its limitations are most visible for fine-grained within-parent confusions, where the parent signal is shared by competing classes and therefore cannot resolve their distinction. Thus, parent feedback complements class-level prediction most directly when the main ambiguity crosses parent boundaries. We examine this behavior further through hierarchical metrics and qualitative examples below.

\paragraph{Failure case.}
HyPLO underperforms CoOp on Flowers102 because its novel class accuracy decreases despite comparable base class accuracy. CoHyPLO and MaHyPLO, in contrast, improve over CoCoOp and MaPLe on Flowers102, suggesting that including images in prompt learning can transfer hierarchical information to unseen classes more effectively than static parent prompts in this setting. Overall, CoHyPLO and MaHyPLO provide the more reliable improvement when novel class generalization is the primary objective.

\begin{table}[t]
\centering
\footnotesize
\setlength{\tabcolsep}{6pt}
\renewcommand{\arraystretch}{1.15}
\resizebox{\textwidth}{!}{%
\begin{tabular}{l c @{\hspace{12pt}} cccccccccc c}
\toprule
& \textbf{Source} & \multicolumn{11}{c}{\textbf{Target}} \\
\cmidrule(lr){2-2} \cmidrule(lr){3-13}
\textbf{Method}
& \rotatebox{90}{ImageNet}
& \rotatebox{90}{Caltech101}
& \rotatebox{90}{OxfordPets}
& \rotatebox{90}{StanfordCars}
& \rotatebox{90}{Flowers102}
& \rotatebox{90}{Food101}
& \rotatebox{90}{Aircraft}
& \rotatebox{90}{SUN397}
& \rotatebox{90}{DTD}
& \rotatebox{90}{EuroSAT}
& \rotatebox{90}{UCF101}
& \rotatebox{90}{\textbf{Average}} \\
\midrule
CoOp                & \textit{47.23} & 86.17 & 54.80 & 10.27 & 29.10 & 56.40 & \textbf{3.70} & 51.33 & 27.10 & 33.43 & 45.03 & 39.73 \\
\rowcolor{gray!10}\textbf{HyPLO}   & \textbf{\textit{47.53}} & \textbf{86.80} & \textbf{55.27} & \textbf{11.83} & \textbf{30.10} & \textbf{58.20} & 3.33 & \textbf{52.83} & \textbf{27.17} & \textbf{34.90} & \textbf{46.63} & \textbf{40.71} \\
\midrule
CoCoOp              & \textit{46.90} & 85.90 & 53.70 & 11.33 & 28.77 & 56.70 & \textbf{3.27} & 52.30 & \textbf{27.23} & \textbf{35.67} & 45.90 & 40.08 \\
\rowcolor{gray!10}\textbf{CoHyPLO} & \textbf{\textit{47.60}} & \textbf{86.90} & \textbf{55.97} & \underline{\textbf{11.93}} & \textbf{30.17} & \textbf{58.57} & 3.23 & \textbf{54.30} & 26.87 & 34.97 & \textbf{47.10} & \textbf{41.00} \\
\midrule
MaPLe              & \textit{46.57} & 86.43 & 55.10 & 10.40 & 26.77 & 54.55 & 3.53 & 51.73 & 27.40 & \underline{\textbf{35.83}} & 45.50 & 39.72 \\
\rowcolor{gray!10}\textbf{MaHyPLO} & \textbf{\textit{47.37}} & \textbf{88.03} & \textbf{55.90} & \textbf{11.27} & \underline{\textbf{30.70}} & \underline{\textbf{59.00}} & \textbf{3.93} & \underline{\textbf{54.93}} & \underline{\textbf{27.97}} & 35.77 & \textbf{47.20} & \underline{\textbf{41.47}} \\
\midrule
PromptSRC              & \underline{\textit{48.63}} & 87.40 & 56.80 & 10.57 & 29.00 & 55.67 & 3.70 & 52.70 & 26.73 & 34.20 & 46.43 & 40.34 \\
\midrule
CoPrompt              & \underline{\textit{48.63}} & \underline{88.80} & \underline{57.20} & 10.83 & 28.27 & 57.37 & \underline{4.53} & 53.43 & 25.93 & 22.43 & \underline{48.57} & 39.74 \\
\bottomrule
\end{tabular}%
}
\caption{
Cross-dataset transfer accuracy (\%). All methods are trained on ImageNet (source) and evaluated on 10 target datasets without further adaptation. Results are averaged over three random seeds, and \textbf{Average} denotes the mean across the 10 target datasets. HyPLO, CoHyPLO, and MaHyPLO are the hierarchical extensions of CoOp, CoCoOp, and MaPLe respectively. \textbf{Bold} indicates the better result among each baseline and its hierarchical extension. \underline{Underlining} indicates the best result in each column across all methods.
}
\label{tab:xdataset}
\end{table}

\subsection{Cross-Dataset Transfer}

We train prompts on ImageNet and evaluate them on the remaining ten datasets without further adaptation (Table~\ref{tab:xdataset}). This setting assesses whether prompt representations learned on the ImageNet label space and visual domain transfer to previously unseen label spaces and domains.

Across the ten target datasets, the hierarchy-aware variants improve their corresponding prompt-learning baselines by 0.92-1.75\% in average accuracy. HyPLO improves over CoOp on 9 of 10 targets, CoHyPLO improves over CoCoOp on 7 of 10 targets, and MaHyPLO improves over MaPLe on 9 of 10 targets. MaHyPLO achieves the strongest target average, improving over MaPLe by 1.75\%. Thus, while the aggregate gains are moderate, they are generally distributed across target datasets rather than driven by a single transfer setting.

These results indicate that parent-level supervision learned on the source domain can provide a transferable semantic prior. At the same time, transfer occurs across datasets with distinct, independently constructed hierarchies: the WordNet-derived ImageNet hierarchy differs from the hierarchy used for each target dataset. Consequently, source-learned parent relations need not correspond directly to the semantic organization of target labels, which may limit the extent of hierarchical transfer.

\begin{figure}[t]
    \centering
    % ----------------------------- Row 1 -----------------------------
    \begin{subfigure}[b]{0.48\textwidth}
        \centering
        \includegraphics[trim={0 0 0 25pt}, clip, width=\linewidth]{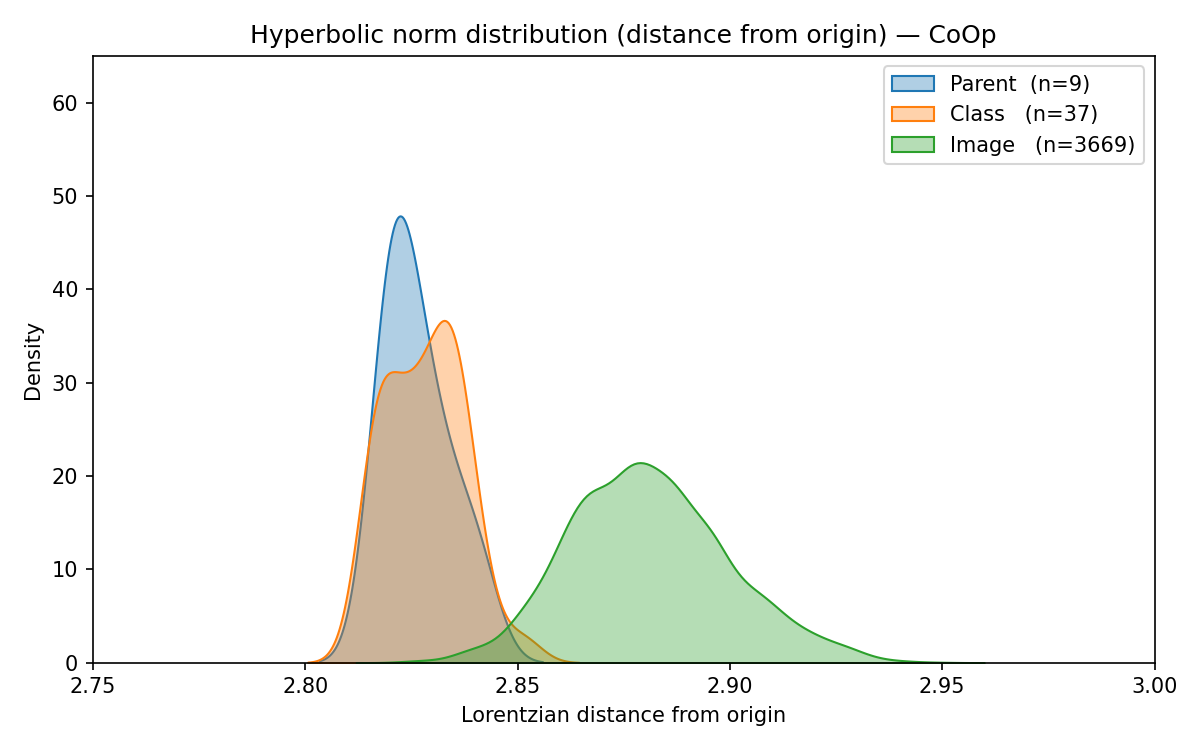}
        \caption{CoOp}
        \label{fig:sub-a}
    \end{subfigure}
    \hfill
    \begin{subfigure}[b]{0.48\textwidth}
        \centering
        \includegraphics[trim={0 0 0 25pt}, clip, width=\linewidth]{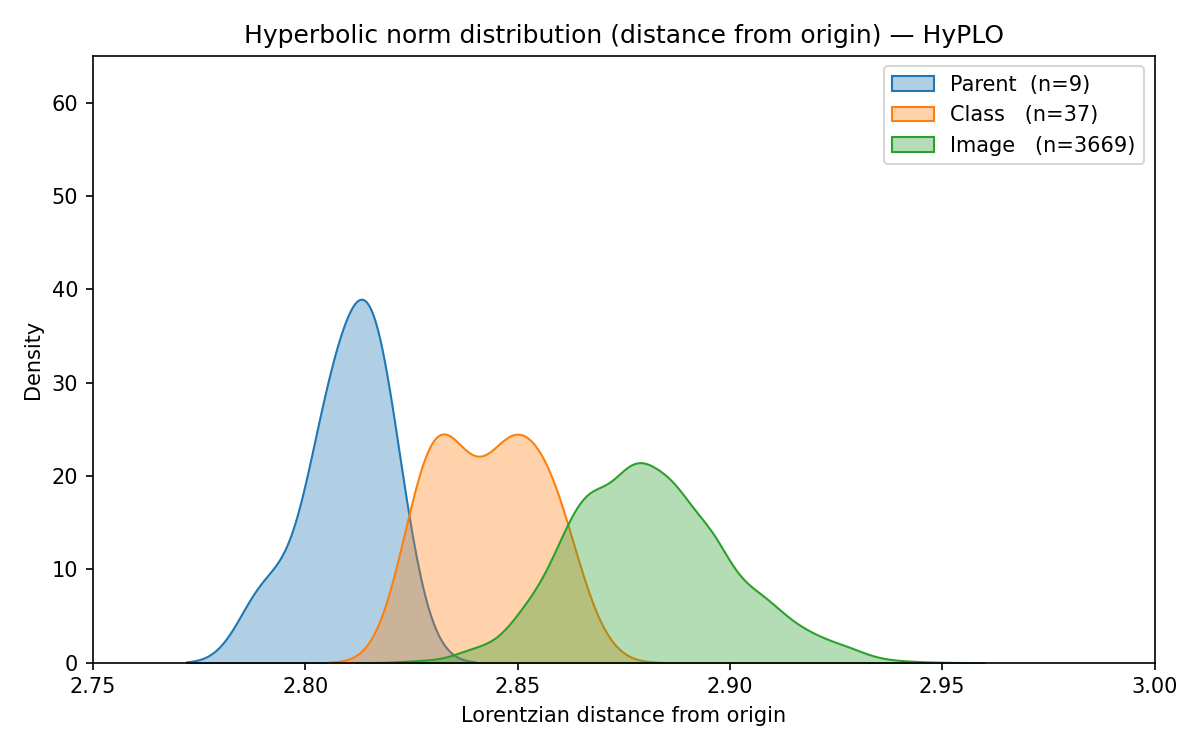}
        \caption{HyPLO}
        \label{fig:sub-b}
    \end{subfigure}

    \vskip\baselineskip   % vertical gap between rows

    % ----------------------------- Row 2 -----------------------------
    \begin{subfigure}[b]{0.48\textwidth}
        \centering
        \includegraphics[trim={0 0 0 25pt}, clip, width=\linewidth]{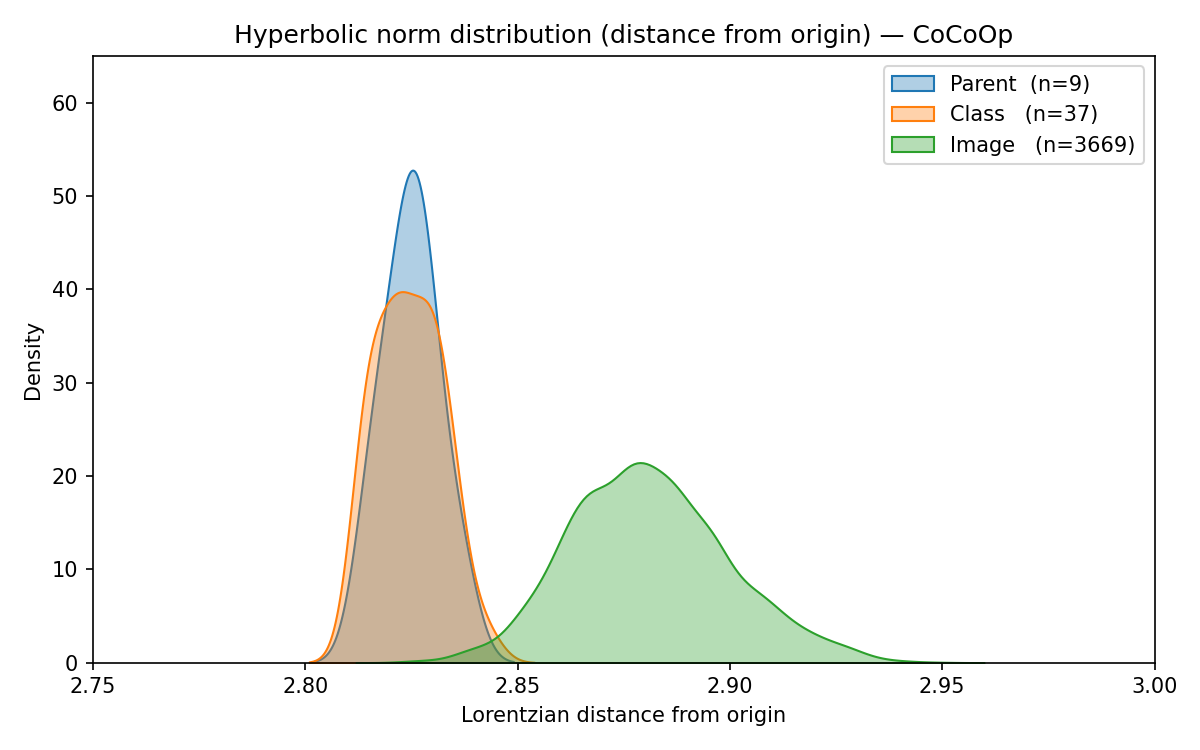}
        \caption{CoCoOp}
        \label{fig:sub-c}
    \end{subfigure}
    \hfill
    \begin{subfigure}[b]{0.48\textwidth}
        \centering
        \includegraphics[trim={0 0 0 25pt}, clip, width=\linewidth]{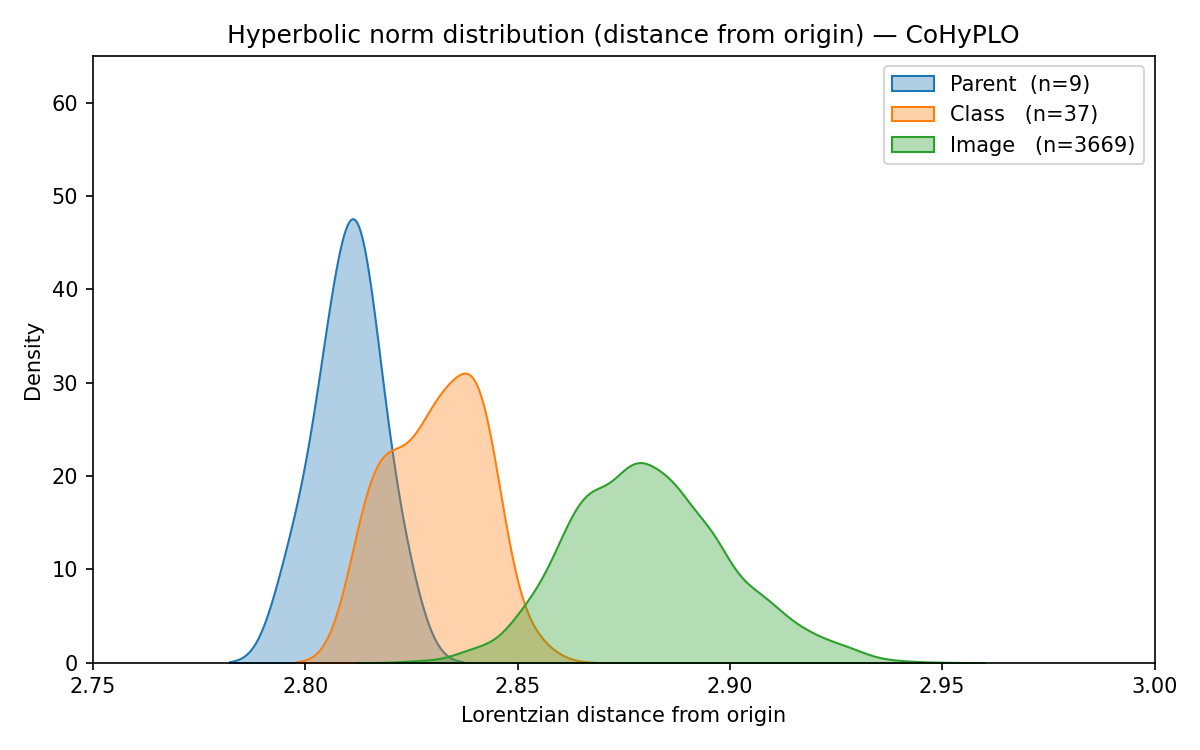}
        \caption{CoHyPLO}
        \label{fig:sub-d}
    \end{subfigure}

    \vskip\baselineskip

    % ----------------------------- Row 3 -----------------------------
    \begin{subfigure}[b]{0.48\textwidth}
        \centering
        \includegraphics[trim={0 0 0 25pt}, clip, width=\linewidth]{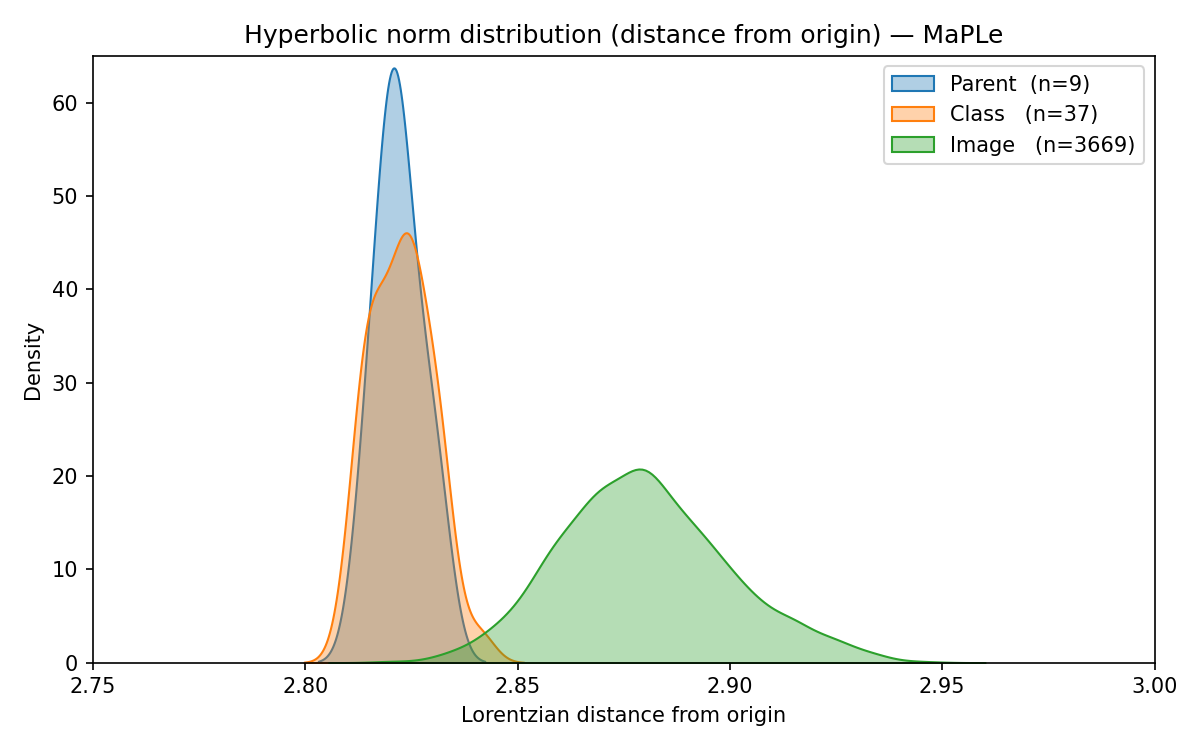}
        \caption{MaPLe}
        \label{fig:sub-e}
    \end{subfigure}
    \hfill
    \begin{subfigure}[b]{0.48\textwidth}
        \centering
        \includegraphics[trim={0 0 0 25pt}, clip, width=\linewidth]{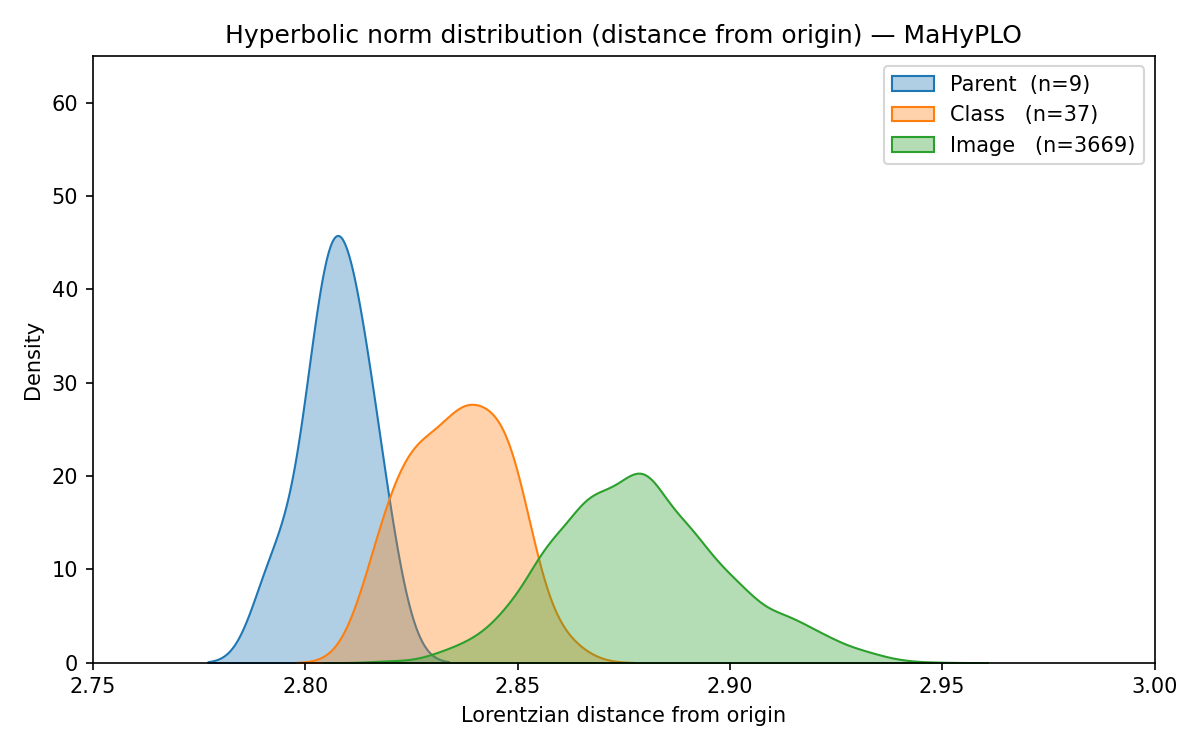}
        \caption{MaHyPLO}
        \label{fig:sub-f}
    \end{subfigure}

    \caption{Distributions of hyperbolic embedding norms, i.e., distances from the origin, for parent, class, and image representations on OxfordPets. The proposed methods (b) HyPLO, (d) CoHyPLO, and (f) MaHyPLO learn a consistent radial ordering, $\text{parent} \rightarrow \text{class} \rightarrow \text{image}$, with parent concepts closest to the origin and image embeddings farthest away. In contrast, their respective non-hierarchical baselines (a) CoOp, (c) CoCoOp, and (e) MaPLe place parent and class text embeddings at similar radii (parent labels were fed to the class prompt learner to generate parent embeddings), while primarily separating image and text embeddings due to inherent capabilities of HyCoCLIP.}
    \label{fig:prompt_norms}
\end{figure}

\subsection{Domain Generalization}

Finally, we evaluate domain generalization by training prompts on ImageNet and testing them on ImageNetV2, ImageNet-Sketch, ImageNet-A, and ImageNet-R (Table~\ref{tab:domaingen}). In this setting, the label space is fixed while the image distribution changes.

The hierarchy-aware variants perform comparably to their underlying prompt learners under distribution shift. Their average differences are small, indicating that the hierarchy module neither materially improves nor degrades robustness across these four target distributions. This outcome is consistent with the role of the method: parent supervision structures the label space, whereas the main challenge in this setting is a shift in visual appearance.

\begin{table}[t]
    \centering
    \footnotesize
    \setlength{\tabcolsep}{6pt}
    \renewcommand{\arraystretch}{1.15}
    
    \begin{tabular}{l c @{\hspace{12pt}} cccc c}
    \toprule
    & \textbf{Source} & \multicolumn{4}{c}{\textbf{Target}} \\
    \cmidrule(lr){2-2} \cmidrule(lr){3-6}
    \textbf{Method}
    & \textbf{ImageNet}
    & \textbf{-V2}
    & \textbf{-S}
    & \textbf{-A}
    & \textbf{-R}
    & \textbf{Avg.} \\
    \midrule
    HyCoCLIP            & \textit{45.10} & 39.20 & 26.60 & 14.00 & 47.90 & 31.93 \\
    HyCoCLIP$^\dagger$  & \textit{44.90} & 38.80 & 26.10 & 14.60 & 47.40 & 31.74 \\
    \midrule
    + CoOp                & \textit{47.23} & 40.83 & 26.80 & 15.63 & 47.63 & 32.73 \\
    \rowcolor{gray!10}\textbf{+ HyPLO}   & \textbf{\textit{47.53}} & \textbf{41.47} & \textbf{27.07} & \textbf{15.67} & \textbf{48.10} & \textbf{33.08} \\
    \midrule
    + CoCoOp              & \textit{46.90} & 40.93 & 26.40 & \underline{\textbf{16.03}} & 47.80 & 32.79 \\
    \rowcolor{gray!10}\textbf{+ CoHyPLO} & \textbf{\textit{47.60}} & \textbf{41.67} & \textbf{26.77} & 15.83 & \textbf{47.93} & \textbf{33.05} \\
    \midrule
    + MaPLe              & \textit{46.57} & 40.50 & \textbf{26.20} & \textbf{15.40} & \textbf{47.97} & \textbf{32.52} \\
    \rowcolor{gray!10}\textbf{+ MaHyPLO} & \textbf{\textit{47.37}} & \textbf{41.13} & 26.17 & 15.30 & 47.40 & 32.50 \\
    \midrule
    + PromptSRC              & \underline{\textit{48.63}} & 41.97 & \underline{28.00} & 14.77 & \underline{49.70} & \underline{33.61} \\
    \midrule
    + CoPrompt              & \underline{\textit{48.63}} & \underline{42.40} & 27.87 & 14.87 & 49.00 & 33.54 \\
    \bottomrule
    \end{tabular}
    
    \caption{Domain generalization accuracy in \%. All methods are trained on ImageNet (source) and evaluated on four out-of-distribution variants: ImageNetV2 (-V2), ImageNet-Sketch (-S), ImageNet-A (-A), and ImageNet-R (-R). \textbf{Avg.} is the mean over the four target variants. HyPLO, CoHyPLO and MaHyPLO are our hierarchical versions of CoOp, CoCoOp and MaPLe respectively. \textbf{Bold} indicates the better result between each baseline and its hierarchical extension. \underline{Underlining} indicates the best result in each column across all methods.}
    \label{tab:domaingen}
\end{table}

\begin{table}[t]
    \centering
    \small
    \setlength{\tabcolsep}{6pt}
    \setlength{\aboverulesep}{0pt}
    \setlength{\belowrulesep}{0pt}
    \renewcommand{\arraystretch}{1.15}
    \begin{tabular}{l c c c c c c c}
    \toprule
    \multirow{2}{*}{\textbf{Method}}
    & \multirow{2}{*}{\textbf{Hier.}}
    & \multicolumn{2}{c}{\cellcolor{greenfirst}\textbf{Hierarchical Distance}}
    & \multicolumn{2}{c}{\cellcolor{greenalt}\textbf{Ancestor Overlap}}
    & \multicolumn{2}{c}{\cellcolor{greencol}\textbf{Local Accuracy}} \\
    \cmidrule(lr){3-4}
    \cmidrule(lr){5-6}
    \cmidrule(lr){7-8}
    &
    & \cellcolor{greenfirst}\textbf{TIE} $\downarrow$
    & \cellcolor{greenfirst}\textbf{LCA} $\downarrow$
    & \cellcolor{greenalt}$\mathbf{J}$ $\uparrow$
    & \cellcolor{greenalt}$\mathbf{F1_H}$ $\uparrow$
    & \cellcolor{greencol}\textbf{S-Acc} $\uparrow$
    & \cellcolor{greencol}\textbf{C-Acc} $\uparrow$ \\
    \midrule
    \addlinespace[4pt]
    HyCoCLIP & \textbf{--}
    & \bn{2.155}{2.110} & \bn{1.549}{1.578} & \bn{0.779}{0.775} & \bn{0.835}{0.828} & \bn{0.623}{0.646} & \bn{0.834}{0.803} \\
    \addlinespace[4pt]
    HyCoCLIP$^\dagger$ & \checkmark
    & \bn{2.234}{2.017} & \bn{1.573}{1.536} & \bn{0.769}{0.789} & \bn{0.826}{0.840} & \bn{0.614}{0.662} & \bn{0.814}{0.823} \\
    \midrule
    \addlinespace[2pt]
    + CoOp & \textbf{--}
    & \bn{1.348}{2.479}
    & \bn{1.326}{1.676}
    & \bn{0.872}{0.731}
    & \bn{0.907}{0.795}
    & \bn{0.767}{0.574}
    & \bn{0.908}{0.775} \\
    \addlinespace[4pt]
    \rowcolor{gray!10}
    \textbf{+ HyPLO} & \checkmark
    & \bn{\textbf{\underline{1.292}}}{\textbf{2.391}}
    & \bn{\textbf{\underline{1.303}}}{\textbf{1.646}}
    & \bn{\textbf{\underline{0.882}}}{\textbf{0.745}}
    & \bn{\textbf{\underline{0.914}}}{\textbf{0.806}}
    & \bn{\textbf{\underline{0.783}}}{\textbf{0.598}}
    & \bn{\textbf{\underline{0.914}}}{\textbf{0.784}} \\
    \midrule
    \addlinespace[2pt]
    + CoCoOp & \textbf{--}
    & \bn{1.548}{2.282}
    & \bn{1.377}{1.620}
    & \bn{0.849}{0.756}
    & \bn{0.890}{0.813}
    & \bn{0.732}{0.618}
    & \bn{0.892}{0.788} \\
    \addlinespace[4pt]
    \rowcolor{gray!10}
    \textbf{+ CoHyPLO} & \checkmark
    & \bn{\textbf{1.489}}{\textbf{2.156}}
    & \bn{\textbf{1.352}}{\textbf{1.579}}
    & \bn{\textbf{0.861}}{\textbf{0.773}}
    & \bn{\textbf{0.899}}{\textbf{0.827}}
    & \bn{\textbf{0.749}}{\textbf{0.638}}
    & \bn{\textbf{0.900}}{\textbf{0.810}} \\
    \midrule
    \addlinespace[2pt]
    + MaPLe & \textbf{--}
    & \bn{1.481}{2.173}
    & \bn{1.356}{1.587}
    & \bn{0.858}{0.769}
    & \bn{0.896}{0.824}
    & \bn{0.744}{0.636}
    & \bn{0.900}{0.801} \\
    \addlinespace[4pt]
    \rowcolor{gray!10}
    \textbf{+ MaHyPLO} & \checkmark
    & \bn{\textbf{1.382}}{\underline{\textbf{2.020}}}
    & \bn{\textbf{1.324}}{\underline{\textbf{1.535}}}
    & \bn{\textbf{0.872}}{\underline{\textbf{0.791}}}
    & \bn{\textbf{0.907}}{\underline{\textbf{0.842}}}
    & \bn{\textbf{0.769}}{\underline{\textbf{0.663}}}
    & \bn{\textbf{0.907}}{\underline{\textbf{0.827}}} \\
    \bottomrule
    \end{tabular}
    \caption{
    Hierarchical evaluation metrics macro-averaged over 11 datasets for different prompt learning methods. For each dataset and metric, results are first averaged over three random seeds; the resulting dataset-level values (shown in Appendix~\ref{app:hier_metrics}) are then macro-averaged. Each cell reports \textbf{Base} (top) and \textbf{New} (bottom) class results. Lower values are better for TIE and LCA; higher values are better for Jaccard similarity (J), hierarchical F1 ($\text{F1}_\text{H}$), sibling accuracy (S-Acc), and cousin accuracy (C-Acc). \textbf{Bold} marks the best result among each baseline and its hierarchical extension. \underline{Underline} marks the best value in each column.
    }
    \label{tab:hier_metrics_avg_b2n}
\end{table}

\subsection{Hierarchy-awareness Metrics}
\label{sec:hier_metrics}

Beyond classification accuracy, we evaluate whether predictions respect the supplied taxonomy. Figure~\ref{fig:prompt_norms} visualizes the radial organization of the learned embeddings, and Table~\ref{tab:hier_metrics_avg_b2n} reports six hierarchical metrics.

\paragraph{Embedding geometry.}
The entailment objective encourages a radial ordering by specificity: parent embeddings should lie closer to the origin than class embeddings, which should in turn lie closer than image embeddings. Figure~\ref{fig:prompt_norms} shows this ordering on OxfordPets. HyPLO, CoHyPLO, and MaHyPLO separate the parent, class, and image distributions, whereas the non-hierarchical prompt learning baselines place parent and class representations at similar radii. This result indicates that the proposed objective induces parent-class organization beyond that provided by the frozen hyperbolic backbone alone.

\paragraph{Hierarchical metrics.}
Table~\ref{tab:hier_metrics_avg_b2n} reports hierarchical distance (TIE and LCA), ancestor set overlap (Jaccard and hierarchical F1), and local error placement (sibling and cousin accuracy), macro-averaged over the 11 datasets with constructed hierarchies. HyPLO attains the strongest base class scores among the learned prompt methods across the reported metrics. On novel classes, MaHyPLO is strongest among the learned prompt methods, consistent with its best aggregate base-to-new performance. Thus, the hierarchy-aware variants improve not only classification performance in the relevant settings but are also more consistent with the supplied hierarchy.

\subsection{Ablation Study}
\label{sec:ablation}

We ablate parent supervision, entailment regularization, inference-time parent fusion, and backbone choice. Unless stated otherwise, ablation results are calculated like the \textit{Average} harmonic mean from Table~\ref{tab:b2n_all}. Ablation study was done on the best performing model, MaHyPLO, and, unless stated otherwise, setup from Section \ref{sec:experiments} is used.

\subsubsection{Loss Terms}

\begin{table}[t]
\centering
\footnotesize
\setlength{\aboverulesep}{0pt}
\setlength{\belowrulesep}{0pt}
\renewcommand{\arraystretch}{1.15}

\begin{subtable}[t]{0.36\textwidth}
\centering
\setlength{\tabcolsep}{5pt}
\begin{tabular}{ccc|c}
\toprule
\ensuremath{\mathcal{L}_{\mathrm{cls}}} &
\ensuremath{\mathcal{L}_{\mathrm{par}}} &
\ensuremath{\mathcal{L}_{\mathrm{ent}}} &
\textbf{H} \\
\midrule
\checkmark & --         & --         & 56.45 \\
\checkmark & \checkmark & --         & \textbf{57.21} \\
\checkmark & --         & \checkmark & 56.45 \\
\rowcolor{gray!10}
\checkmark & \checkmark & \checkmark & 57.18 \\
\bottomrule
\end{tabular}
\caption{Loss-component ablation on base-to-new generalization.}
\label{tab:abl_loss_components}
\end{subtable}
\hfill
\begin{subtable}[t]{0.61\textwidth}
\centering
\scriptsize
\setlength{\tabcolsep}{3.5pt}
\begin{tabular}{ccc|ccc}
\toprule
\ensuremath{\mathcal{L}_{\mathrm{cls}}} &
\ensuremath{\mathcal{L}_{\mathrm{par}}} &
\ensuremath{\mathcal{L}_{\mathrm{ent}}} &
\textbf{B2N} &
\textbf{XDT} &
\textbf{DG} \\
\midrule
\checkmark & \checkmark & -- &
\textbf{57.21} & \textbf{41.49} & 32.44 \\
\rowcolor{gray!10}
\checkmark & \checkmark & \checkmark &
57.18 & 41.47 & \textbf{32.50} \\
\bottomrule
\end{tabular}
\caption{Effect of entailment regularization across main experiments.}
\label{tab:abl_entailment_protocols}
\end{subtable}

\par\medskip

\begin{subtable}[t]{0.52\textwidth}
\centering
\footnotesize
\setlength{\tabcolsep}{7pt}
\begin{tabular}{lccccc}
\toprule
\ensuremath{\lambda_{\mathrm{par}}} &
0.0 &
\cellcolor{gray!10}0.25 &
0.5 &
0.75 &
1.0 \\
\midrule
\textbf{Base} &
63.00 &
\cellcolor{gray!10}\textbf{63.26} &
63.06 &
62.63 &
62.25 \\
\bottomrule
\end{tabular}
\caption{Inference-time parent-fusion ablation.}
\label{tab:abl_fusion}
\end{subtable}

\caption{
Ablations of hierarchical loss components and inference-time parent fusion.
\textbf{(a)} Comparison of loss combinations on base-to-new harmonic mean
(H). \textbf{(b)} Comparison between parent supervision alone and
the full method across base-to-new generalization (B2N), cross-dataset
transfer (XDT), and domain generalization (DG). \textbf{(c)} Validation
ablation of the inference-time parent-fusion weight. Scores are percentages
and macro-averaged over datasets after averaging three seeds within each
dataset, following the protocol of each experiment. Gray rows denote the full
model; gray cells in \textbf{(c)} denote the fusion weight used in the main
experiments.
}
\label{tab:abl_loss}
\end{table}

Table~\ref{tab:abl_loss_components} and \ref{tab:abl_entailment_protocols} isolate the contributions of the class loss, parent supervision, and entailment regularization. Parent supervision provides the main improvement over class-only prompt learning on base-to-new generalization. Entailment regularization yields only small accuracy differences across the main experiments, while inducing the desired radial organization of the hierarchy.

\begin{figure}[t]
\centering
\begin{subfigure}{0.48\textwidth}
  \centering
  \includegraphics[trim={0 0 0 25pt}, clip, width=\linewidth]{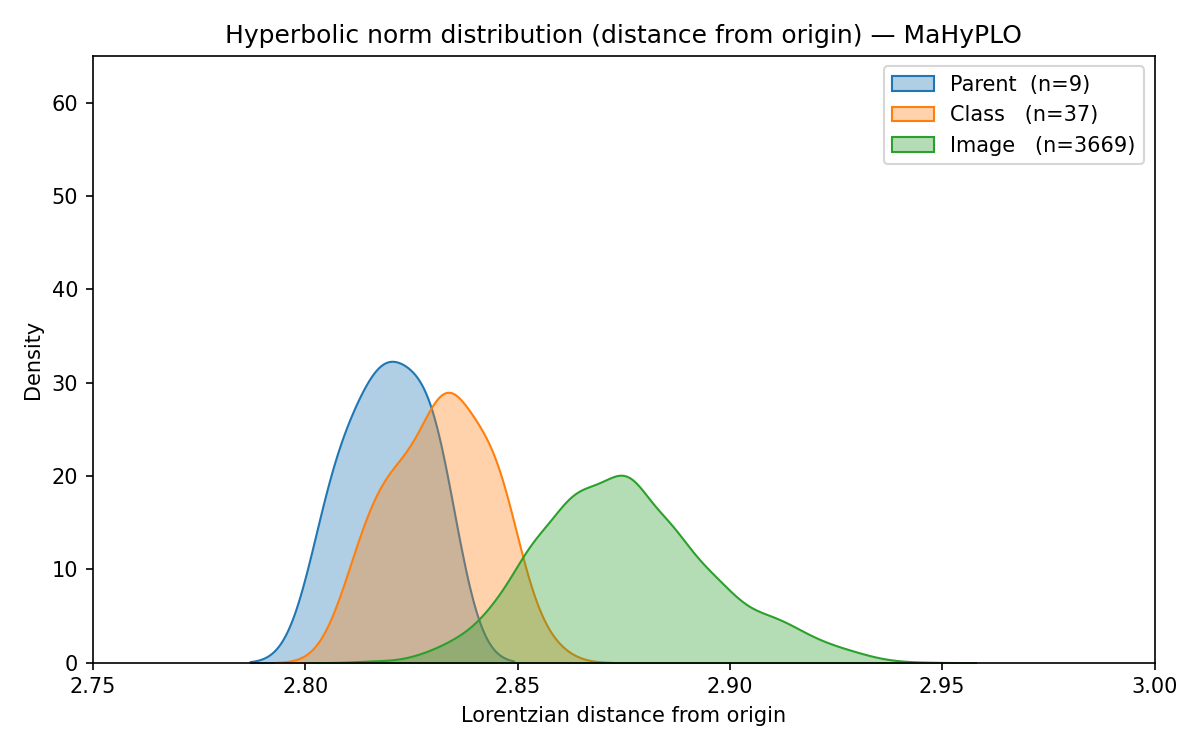}
  \caption{$\mathcal{L}_{\mathrm{cls}} + \mathcal{L}_{\mathrm{par}}$}
  \label{fig:abl_norms_noent}
\end{subfigure}
\hfill
\begin{subfigure}{0.48\textwidth}
  \centering
  \includegraphics[trim={0 0 0 25pt}, clip, width=\linewidth]{media/figures/mahyplo_eval_distance_test_norms.png}
  \caption{$\mathcal{L}_{\mathrm{cls}} + \mathcal{L}_{\mathrm{par}} + \mathcal{L}_{\mathrm{ent}}$ (our model)}
  \label{fig:abl_norms_full}
\end{subfigure}
\caption{Effect of the entailment term on embedding geometry (MaHyPLO on OxfordPets; 9 parents, 37 classes, 3669 images). \textbf{(a)} Without $\mathcal{L}_{\mathrm{ent}}$, the parent and class norm distributions sit at nearly the same radius. \textbf{(b)} With $\mathcal{L}_{\mathrm{ent}}$, the parent is drawn toward the origin and the class outward, recovering the parent $\rightarrow$ class $\rightarrow$ image ordering. The entailment term induces the radial separation between parent and class at negligible cost to accuracy (Table~\ref{tab:abl_loss}).}
\label{fig:ablation_norms}
\end{figure}

Figure~\ref{fig:ablation_norms} illustrates its structural effect. Without entailment regularization, parent and class embeddings occupy similar radii. With it, they separate into the intended parent$\rightarrow$class$\rightarrow$image ordering. We therefore retain the entailment term because it induces the desired hierarchical geometry while preserving the overall performance.

\subsubsection{Inference Fusion}
\label{sec:infer_fusion}

Table~\ref{tab:abl_fusion} evaluates the inference-time parent-fusion weight on base class validation accuracy across 11 datasets. Parent feedback improves over class-only inference, and the selected weight attains the highest macro-average validation accuracy. Larger weights gradually reduce performance, indicating that excessive parent feedback can override class-level evidence. We use \ensuremath{\lambda_{\mathrm{par}} = 0.25} in all main experiments.

\subsubsection{Vision-Language Model Backbone Choice}

Our experiments use HyCoCLIP. To assess backbone portability, we repeat the selected base-to-new experiment with MERU \citep{desai2023hyperbolic} while keeping the prompt learning protocol fixed (Table~\ref{tab:backbone}).

HyCoCLIP provides stronger absolute performance in this comparison. However, the hierarchy-aware variants improve over their corresponding prompt learners on both MERU and HyCoCLIP. This result supports the claim that the proposed plug-in is not specific to HyCoCLIP.

% ---------- Row 1 ----------
\begin{table}[t]
\centering
\setlength{\aboverulesep}{0pt}
\setlength{\belowrulesep}{0pt}
\begin{tabular}{lcc}
\toprule
\textbf{Method} & \textbf{MERU} & \cellcolor{gray!10}\textbf{HyCoCLIP} \\
\midrule
Backbone           & 45.27 & \cellcolor{gray!10}\textbf{49.40} \\
\midrule
% HyCoCLIP$^\dagger$ & 46.2 & \textbf{52.1} & 48.98 \\
+ CoOp               & 47.92 & \cellcolor{gray!10}\textbf{52.29} \\
\textbf{+ HyPLO}   & 49.00 & \cellcolor{gray!10}\textbf{53.19} \\
\midrule
+ CoCoOp             & 48.73 & \cellcolor{gray!10}\textbf{53.35} \\
\textbf{+ CoHyPLO} & 49.69 & \cellcolor{gray!10}\textbf{54.66} \\
\midrule
+ MaPLe             & 50.35 & \cellcolor{gray!10}\textbf{55.21} \\
\textbf{+ MaHyPLO} & 51.88 & \cellcolor{gray!10}\textbf{57.18} \\
\bottomrule
\end{tabular}
\caption{VLM backbone ablation. Tested on base-to-new generalization, reporting the harmonic mean H accuracy in \%.}
\label{tab:backbone}
\end{table}%

% Define colors
\definecolor{correct}{RGB}{0, 150, 0}
\definecolor{incorrect}{RGB}{200, 0, 0}
\definecolor{boxgreen}{RGB}{180, 230, 180}
\definecolor{boxred}{RGB}{230, 180, 180}

\begin{figure}[t]
    \centering

    \begin{tabular}{@{}p{0.28\linewidth}p{0.28\linewidth}p{0.28\linewidth}@{}}
        \centering\textbf{Ground Truth} &
        \centering\textbf{Class} &
        \centering\arraybackslash\textbf{Class + Parent (Ours)}
    \end{tabular}

    \vspace{4pt}

    % GOOD EXAMPLE
    \resizebox{\linewidth}{!}{%
        \begin{tikzpicture}
            \node[draw=correct, line width=2pt, rounded corners=8pt,
                  fill=correct!10, inner sep=10pt] {%
                \begin{tabular}{@{}ccc@{}}
                    \multicolumn{3}{l}{\textcolor{correct}{\textit{\textbf{GOOD EXAMPLE}}}} \\[4pt]
                    \includegraphics[width=0.28\linewidth]{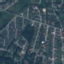} &
                    \includegraphics[width=0.28\linewidth]{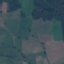} &
                    \includegraphics[width=0.28\linewidth]{media/figures/Residential_839.jpg} \\[4pt]
                    \begin{tabular}[t]{@{}l@{}}
                        \textbf{True class:} residential buildings \\
                        \textbf{True parent:} built-up area
                    \end{tabular}
                    &
                    \begin{tabular}[t]{@{}l@{}}
                        \textbf{Pred.\ class:} \textcolor{incorrect}{pasture land} \\
                        \textbf{Pred.\ parent:} - \\
                        \textbf{Test Fusion:} \textcolor{incorrect}{pasture land}
                    \end{tabular}
                    &
                    \begin{tabular}[t]{@{}l@{}}
                        \textbf{Pred.\ class:} \textcolor{incorrect}{pasture land} \\
                        \textbf{Pred.\ parent:} \textcolor{correct}{built-up area} \\
                        \textbf{Test Fusion:} \textcolor{correct}{residential buildings}
                    \end{tabular} \\
                \end{tabular}%
            };
        \end{tikzpicture}%
    }

    \vspace{10pt}

    % BAD EXAMPLE
    \resizebox{\linewidth}{!}{%
        \begin{tikzpicture}
            \node[draw=incorrect, line width=2pt, rounded corners=8pt,
                  fill=incorrect!10, inner sep=10pt] {%
                \begin{tabular}{@{}ccc@{}}
                    \multicolumn{3}{l}{\textcolor{incorrect}{\textit{\textbf{BAD EXAMPLE}}}} \\[4pt]
                    \includegraphics[width=0.28\linewidth]{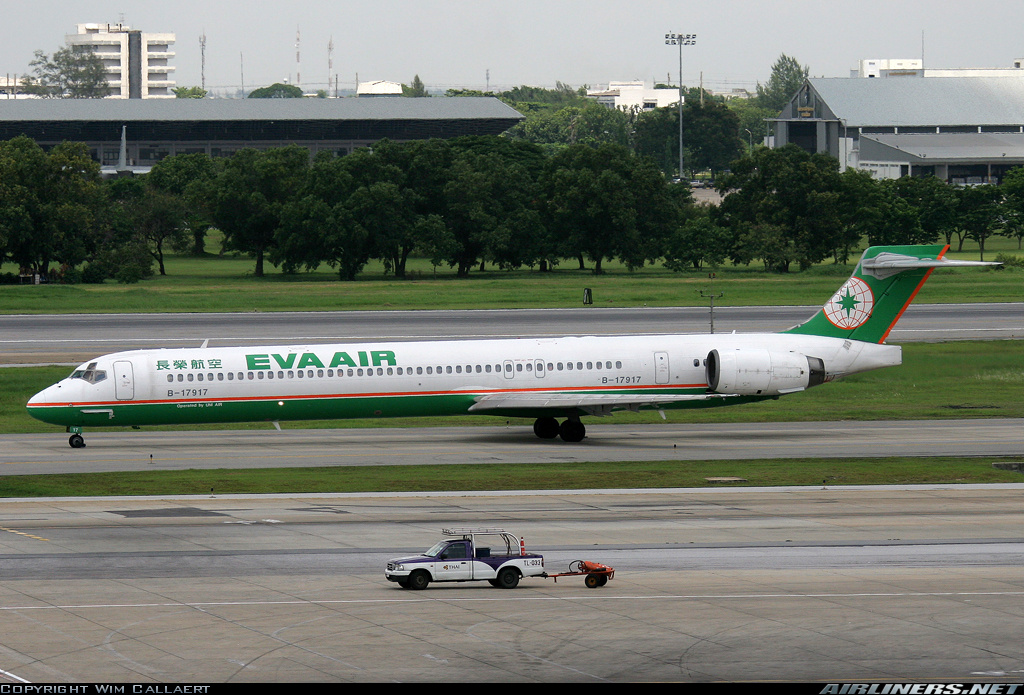} &
                    \includegraphics[width=0.28\linewidth]{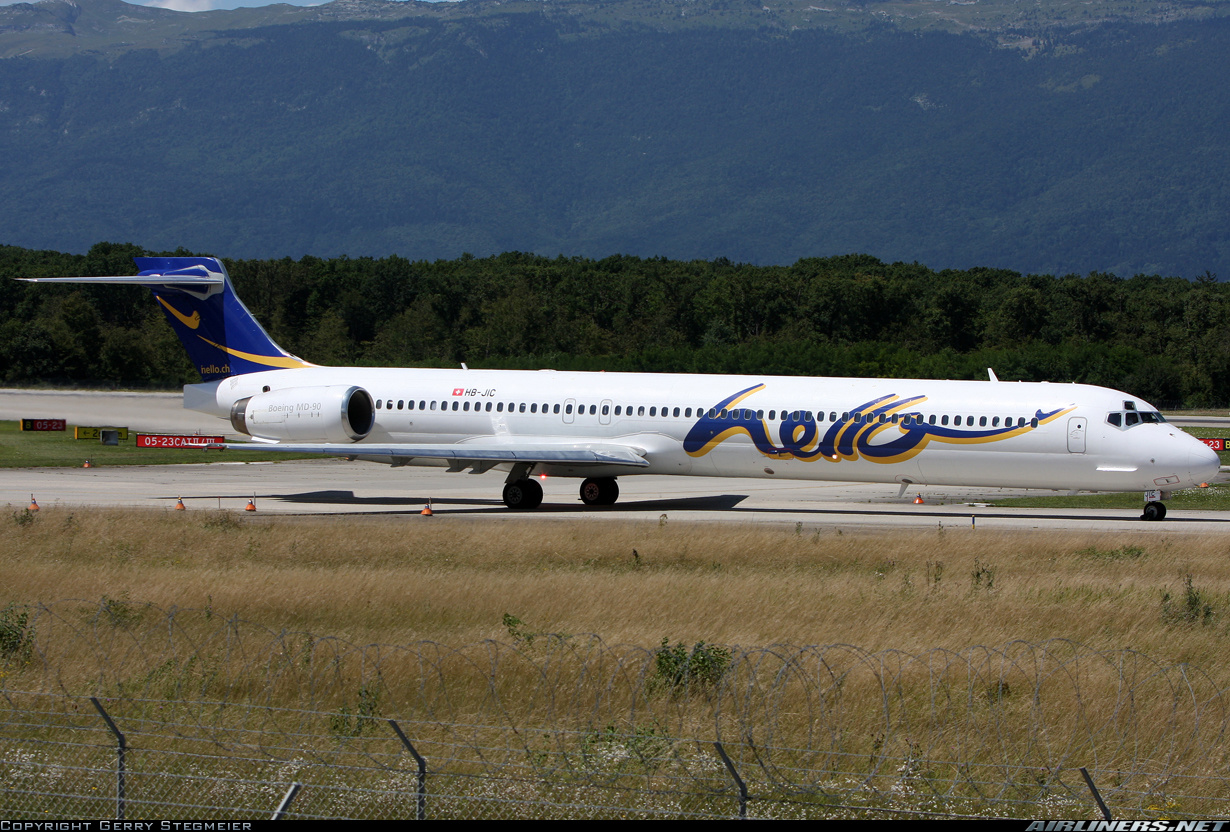} &
                    \includegraphics[width=0.28\linewidth]{media/figures/false_airplane.jpg} \\[4pt]
                    \begin{tabular}[t]{@{}l@{}}
                        \textbf{True class:} MD-87 \\
                        \textbf{True parent:} narrow-body airliner
                    \end{tabular}
                    &
                    \begin{tabular}[t]{@{}l@{}}
                        \textbf{Pred.\ class:} \textcolor{incorrect}{MD-90} \\
                        \textbf{Pred.\ parent:} - \\
                        \textbf{Test Fusion:} \textcolor{incorrect}{MD-90}
                    \end{tabular}
                    &
                    \begin{tabular}[t]{@{}l@{}}
                        \textbf{Pred.\ class:} \textcolor{incorrect}{MD-90} \\
                        \textbf{Pred.\ parent:} \textcolor{correct}{narrow-body airliner} \\
                        \textbf{Test Fusion:} \textcolor{incorrect}{MD-90}
                    \end{tabular} \\
                \end{tabular}%
            };
        \end{tikzpicture}%
    }

    \caption{Qualitative results showing where our method succeeds (above) and where our method fails (below). Left column are ground-truth images; middle column are images predicted by class-only prompt learner; right column are images predicted by our method.}
    \label{fig:qualitative}
\end{figure}

\subsection{Qualitative Analysis}

Figure~\ref{fig:qualitative} illustrates when parent-feedback fusion helps. In the EuroSAT example, the class head predicts a class from the wrong parent, while the parent head correctly identifies the broader category. Fusion then shifts the final prediction to the correct class. This is the intended role of parent feedback: resolving errors that cross parent boundaries.

The FGVCAircraft example shows the complementary limitation. The parent head correctly identifies the broad aircraft category, but the competing classes are siblings under that parent. Because parent feedback affects both candidates similarly, it cannot resolve their fine-grained distinction. Parent fusion is therefore most useful for cross-parent ambiguity, not for within-parent confusions.

\section{Conclusion}

We introduced a prompt-learner-agnostic hyperbolic hierarchical plug-in for frozen hyperbolic VLMs. The method preserves the original class prompt learner and adds a separate parent prompt learner. Given a fixed offline parent-class hierarchy, it combines class and parent supervision with hyperbolic entailment regularization and parent-feedback logit fusion. This incorporates taxonomic information into prompt adaptation without modifying the frozen backbone or requiring an auxiliary graph encoder.

Across base-to-new generalization, the hierarchy-aware variants improve the prompt learning methods they extend. HyPLO, CoHyPLO, and MaHyPLO preserve the characteristic behavior of their respective CoOp, CoCoOp, and MaPLe baselines while improving their overall balance between base and novel classes. The hierarchical metrics and embedding analysis further show that the learned representations exhibit the intended parent $\rightarrow$ class $\rightarrow$ image organization in hyperbolic space. These results indicate that hierarchical structure available in hyperbolic vision-language representations can also be exploited effectively at the prompt learning interface.

Our analysis also identifies the scope of these benefits. The method is most helpful when prediction requires positioning unseen classes within a meaningful and stable taxonomy. Its gains are slightly smaller under cross-dataset transfer, where source and target label hierarchies may align only partially, and limited under domain generalization, where the label space is fixed and the main challenge is visual distribution shift. Likewise, parent-level information is less informative for fine-grained confusions among sibling classes. The entailment regularizer consistently improves geometric organization, whereas parent-level supervision accounts for most of the accuracy gains.

Future work should consider variable-depth, incomplete, or noisy taxonomies; additional prompt learning baselines; and Euclidean vision-language backbones mapped into hyperbolic space at adaptation time. Learning the loss weights, parent-fusion weight, and hyperbolic softmax temperature could also reduce manual tuning and adapt the strength of hierarchical supervision to each dataset.

\bibliography{main}
\bibliographystyle{tmlr}

\newpage

\appendix
\section{Hierarchies}
\label{app:hierarchies}

This appendix lists the class and parent labels used to construct the semantic hierarchies for each dataset in our experiments. Following the process described in Algorithm~\ref{alg:hierarchy}, the hierarchies were curated to group semantically related classes under higher-level parent categories, providing the class-parent relationships required for hierarchical evaluation.

Due to space limitations, the complete hierarchies for SUN397 and ImageNet are not included in this document. Both datasets contain substantially larger label sets (397 and 1000 classes, respectively), making their full mappings impractical to reproduce.

\begin{table}[H]
\centering
\footnotesize
\renewcommand{\arraystretch}{1.2}
\begin{tabular}{@{}>{\bfseries}l p{0.72\columnwidth}@{}}
\toprule
\textnormal{\textbf{Parent}} & \textbf{Classes} \\
\midrule
anatomy & face, brain \\
feline & leopard, cougar body, cougar face, wild cat \\
vehicle & motorbike, airplane, car side, ferry, helicopter, ketch, schooner \\
musical instrument & accordion, electric guitar, euphonium, gramophone, grand piano, mandolin, metronome, saxophone \\
tool & anchor, scissors, stapler, umbrella, wrench \\
arthropod & ant, butterfly, dragonfly, mayfly, scorpion, tick \\
container & barrel, cup, ewer \\
aquatic vertebrate & bass, dolphin, sea horse \\
small mammal & beaver, dalmatian, hedgehog, kangaroo, panda, platypus \\
electronic device & binocular, camera, cellphone, headphone, laptop, watch \\
tree & bonsai, joshua tree \\
prehistoric animal & brontosaurus, stegosaurus, trilobite \\
symbolic object & buddha, dollar bill, menorah, stop sign, yin yang \\
weapon & cannon, revolver \\
home furnishing & ceiling fan, chair, chandelier, lamp, wheelchair, windsor chair \\
marine invertebrate & crab, crayfish, lobster, nautilus, octopus, starfish \\
reptile & crocodile, crocodile head, hawksbill \\
large herbivore & elephant, gerenuk, llama, okapi, rhino \\
bird & emu, flamingo, flamingo head, ibis, pigeon, rooster \\
cartoon character & garfield, snoopy \\
sports equipment & inline skate, soccer ball \\
flower & lotus, sunflower, water lilly \\
building & minaret, pagoda, pyramid \\
food & pizza, strawberry \\
\bottomrule
\end{tabular}
\caption{Caltech101 hierarchy: 100 classes grouped under 24 parents (mean 4.2 children per parent).}
\label{tab:hier_caltech101}
\end{table}

\begin{table}[H]
\centering
\footnotesize
\renewcommand{\arraystretch}{1.2}
\begin{tabular}{@{}>{\bfseries}l p{0.72\columnwidth}@{}}
\toprule
\textnormal{\textbf{Parent}} & \textbf{Classes} \\
\midrule
cropland & Annual Crop Land, Permanent Crop Land \\
natural vegetation & Forest, Herbaceous Vegetation Land, Pasture Land \\
built-up area & Highway or Road, Industrial Buildings, Residential Buildings \\
water body & River, Sea or Lake \\
\bottomrule
\end{tabular}
\caption{EuroSAT hierarchy: 10 classes grouped under 4 parents (mean 2.5 children per parent).}
\label{tab:hier_eurosat}
\end{table}

\begin{table}[H]
\centering
\footnotesize
\renewcommand{\arraystretch}{1.2}
\begin{tabular}{@{}>{\bfseries}l p{0.72\columnwidth}@{}}
\toprule
\textnormal{\textbf{Parent}} & \textbf{Classes} \\
\midrule
layered texture & banded, crystalline, scaly, stratified \\
spotted pattern & blotchy, dotted, flecked, freckled, polka-dotted, sprinkled \\
woven texture & braided, fibrous, gauzy, interlaced, knitted, lacelike, matted, woven \\
bumpy surface & bubbly, bumpy, grooved, studded \\
lattice pattern & chequered, crosshatched, grid, waffled \\
porous texture & cobwebbed, honeycombed, meshed, perforated, pitted, porous \\
damaged surface & cracked, potholed \\
wrinkled surface & frilly, pleated, wrinkled \\
linear pattern & lined, striped, veined, zigzagged \\
wavy pattern & marbled, paisley, spiralled, swirly \\
stained surface & smeared, stained \\
\bottomrule
\end{tabular}
\caption{DTD hierarchy: 47 classes grouped under 11 parents (mean 4.3 children per parent).}
\label{tab:hier_dtd}
\end{table}

\begin{table}[H]
\centering
\footnotesize
\renewcommand{\arraystretch}{1.2}
\begin{tabular}{@{}>{\bfseries}l p{0.72\columnwidth}@{}}
\toprule
\textnormal{\textbf{Parent}} & \textbf{Classes} \\
\midrule
spotted cat & abyssinian, bengal, egyptian mau \\
bulldog & american bulldog, american pit bull terrier, boxer \\
hunting dog & basset hound, beagle, english cocker spaniel, english setter, german shorthaired \\
longhair cat & birman, maine coon, persian, ragdoll \\
shorthair cat & bombay, british shorthair, russian blue, siamese, sphynx \\
toy dog & chihuahua, havanese, japanese chin, miniature pinscher, pug, yorkshire terrier \\
large dog & great pyrenees, leonberger, newfoundland, saint bernard \\
spitz dog & keeshond, pomeranian, samoyed, shiba inu \\
terrier & scottish terrier, staffordshire bull terrier, wheaten terrier \\
\bottomrule
\end{tabular}
\caption{OxfordPets hierarchy: 37 classes grouped under 9 parents (mean 4.1 children per parent).}
\label{tab:hier_oxfordpets}
\end{table}

\begin{table}[H]
\centering
\footnotesize
\renewcommand{\arraystretch}{1.2}
\begin{tabular}{@{}>{\bfseries}l p{0.72\columnwidth}@{}}
\toprule
\textnormal{\textbf{Parent}} & \textbf{Classes} \\
\midrule
classic long-range airliner & 707-320, 727-200, A340-200, A340-300, A340-500, A340-600, DC-10, DC-8, L-1011, MD-11 \\
narrow-body airliner & 737-200, 737-300, 737-400, 737-500, 737-600, 737-700, 737-800, 737-900, 757-200, 757-300, A318, A319, A320, A321, Boeing 717, Fokker 100, MD-80, MD-87, MD-90 \\
wide-body airliner & 747-100, 747-200, 747-300, 747-400, 767-200, 767-300, 767-400, 777-200, 777-300, A300B4, A310, A330-200, A330-300, A380 \\
regional turboprop & ATR-42, ATR-72, Beechcraft 1900, DHC-8-100, DHC-8-300, Dornier 328, EMB-120, Fokker 50, Metroliner, Saab 2000, Saab 340 \\
military transport & An-12, C-130, C-47, Il-76 \\
regional jet & BAE 146-200, BAE 146-300, CRJ-200, CRJ-700, CRJ-900, DC-9-30, E-170, E-190, E-195, ERJ 135, ERJ 145, Fokker 70, Tu-134, Tu-154, Yak-42 \\
business jet & BAE-125, Cessna 525, Cessna 560, Challenger 600, Embraer Legacy 600, Falcon 2000, Falcon 900, Global Express, Gulfstream IV, Gulfstream V \\
light propeller plane & Cessna 172, Cessna 208, DHC-6, DR-400, Model B200, PA-28, SR-20 \\
vintage propeller plane & DC-3, DC-6, DH-82, DHC-1, Spitfire \\
military jet & Eurofighter Typhoon, F-16A/B, F/A-18, Hawk T1, Tornado \\
\bottomrule
\end{tabular}
\caption{FGVCAircraft hierarchy: 100 classes grouped under 10 parents (mean 10.0 children per parent).}
\label{tab:hier_fgvcaircraft}
\end{table}

\begin{table}[H]
\centering
\footnotesize
\renewcommand{\arraystretch}{1.2}
\begin{tabular}{@{}>{\bfseries}l p{0.72\columnwidth}@{}}
\toprule
\textnormal{\textbf{Parent}} & \textbf{Classes} \\
\midrule
baked dessert & apple pie, baklava, bread pudding, carrot cake, cheesecake, chocolate cake, cup cakes, red velvet cake, strawberry shortcake, tiramisu \\
meat dish & baby back ribs, beef carpaccio, beef tartare, chicken curry, chicken quesadilla, chicken wings, filet mignon, foie gras, peking duck, pork chop, prime rib, steak \\
salad & beet salad, caesar salad, caprese salad, ceviche, greek salad, seaweed salad \\
fried pastry & beignets, cannoli, churros, donuts, macarons, waffles \\
noodle and rice & bibimbap, fried rice, pad thai, paella, pho, ramen, risotto \\
sandwich & breakfast burrito, club sandwich, croque madame, grilled cheese sandwich, hamburger, hot dog, lobster roll sandwich, pulled pork sandwich, tacos \\
dip or spread & bruschetta, cheese plate, escargots, guacamole, hummus \\
frozen dessert & chocolate mousse, creme brulee, frozen yogurt, ice cream, panna cotta \\
soup or stew & clam chowder, french onion soup, hot and sour soup, lobster bisque, miso soup \\
seafood dish & crab cakes, fish and chips, grilled salmon, mussels, oysters, sashimi, scallops, shrimp and grits, sushi, tuna tartare \\
small bite & deviled eggs, edamame \\
dumpling or roll & dumplings, gyoza, spring rolls, takoyaki \\
breakfast dish & eggs benedict, french toast, huevos rancheros, omelette, pancakes \\
fried bite & falafel, fried calamari, nachos, samosa \\
fried side & french fries, garlic bread, onion rings, poutine \\
pasta and pizza & gnocchi, lasagna, macaroni and cheese, pizza, ravioli, spaghetti bolognese, spaghetti carbonara \\
\bottomrule
\end{tabular}
\caption{Food101 hierarchy: 101 classes grouped under 16 parents (mean 6.3 children per parent).}
\label{tab:hier_food101}
\end{table}

\begin{table}[H]
\centering
\footnotesize
\renewcommand{\arraystretch}{1.2}
\begin{tabular}{@{}>{\bfseries}l p{0.72\columnwidth}@{}}
\toprule
\textnormal{\textbf{Parent}} & \textbf{Classes} \\
\midrule
perennial flower & pink primrose, monkshood, balloon flower, pincushion flower, lenten rose, gaura, japanese anemone, windflower, columbine, cyclamen, bee balm, foxglove, mallow \\
orchid & hard-leaved pocket orchid, moon orchid, king protea, ruby-lipped cattleya, cape flower \\
annual flower & canterbury bells, snapdragon, sweet william, carnation, garden phlox, love in the mist, bolero deep blue, wallflower, petunia, wild pansy, primula, pelargonium, geranium \\
climbing flower & sweet pea, morning glory, passion flower, clematis, bougainvillea, mexican petunia, trumpet creeper \\
dahlia or sunflower & english marigold, globe-flower, purple coneflower, marigold, sunflower, bishop of llandaff, orange dahlia, pink-yellow dahlia, blanket flower \\
lily & tiger lily, giant white arum lily, fire lily, toad lily, blackberry lily \\
tropical flower & bird of paradise, red ginger, siam tulip, anthurium, frangipani, hibiscus, canna lily, ball moss, bromelia \\
thistle or spiky flower & globe thistle, spear thistle, prince of wales feathers, artichoke, alpine sea holly, great masterwort \\
meadow flower & colt's foot, corn poppy, stemless gentian, buttercup, silverbush, californian poppy, tree poppy, thorn apple \\
bulb flower & yellow iris, peruvian lily, fritillary, grape hyacinth, daffodil, sword lily, cautleya spicata, spring crocus, bearded iris, hippeastrum \\
daisy-like flower & mexican aster, barbeton daisy, oxeye daisy, common dandelion, black-eyed susan, osteospermum, gazania \\
flowering shrub & poinsettia, azalea, rose, desert-rose, tree mallow, magnolia, camellia \\
aquatic flower & water lily, lotus, watercress \\
\bottomrule
\end{tabular}
\caption{OxfordFlowers hierarchy: 102 classes grouped under 13 parents (mean 7.8 children per parent).}
\label{tab:hier_oxfordflowers}
\end{table}

\begin{table}[H]
\centering
\scriptsize
\renewcommand{\arraystretch}{1.2}
\begin{tabular}{@{}>{\bfseries}l p{0.78\columnwidth}@{}}
\toprule
\textnormal{\textbf{Parent}} & \textbf{Classes} \\
\midrule
midsize suv & 2000 AM General Hummer SUV, 2007 Buick Rainier SUV, 2012 Chevrolet Traverse SUV, 2012 Chevrolet Tahoe Hybrid SUV, 2009 Chevrolet TrailBlazer SS, 2009 Chrysler Aspen SUV, 2012 Dodge Durango SUV, 2007 Dodge Durango SUV, 2009 Ford Expedition EL SUV, 2012 GMC Yukon Hybrid SUV, 2012 GMC Acadia SUV, 2012 Hyundai Veracruz SUV, 2008 Isuzu Ascender SUV, 2012 Jeep Grand Cherokee SUV, 2012 Toyota Sequoia SUV, 2012 Toyota 4Runner SUV \\
luxury sedan & 2012 Acura RL Sedan, 1994 Audi V8 Sedan, 1994 Audi 100 Sedan, 2011 Audi S6 Sedan, 2010 BMW M5 Sedan, 2009 Bentley Arnage Sedan, 2011 Bentley Mulsanne Sedan, 2007 Bentley Continental Flying Spur Sedan, 2012 Cadillac CTS-V Sedan, 2010 Chrysler 300 SRT-8, 2011 Lincoln Town Car Sedan, 2012 Mercedes-Benz E-Class Sedan, 2012 Mercedes-Benz S-Class Sedan, 2012 Porsche Panamera Sedan, 2012 Rolls-Royce Ghost Sedan, 2012 Rolls-Royce Phantom Sedan \\
midsize sedan & 2012 Acura TL Sedan, 2008 Acura TL Type-S, 2012 Audi S4 Sedan, 2007 Audi S4 Sedan, 2012 BMW ActiveHybrid 5 Sedan, 2012 BMW 3 Series Sedan, 2012 Buick Regal GS, 2007 Chevrolet Impala Sedan, 2010 Chevrolet Malibu Hybrid Sedan, 2007 Chevrolet Malibu Sedan, 2012 Dodge Charger Sedan, 2009 Dodge Charger SRT-8, 2012 Fisker Karma Sedan, 2012 Honda Accord Sedan, 2012 Hyundai Sonata Hybrid Sedan, 2012 Hyundai Genesis Sedan, 2012 Hyundai Sonata Sedan, 2012 Hyundai Azera Sedan, 2012 Tesla Model S Sedan, 2012 Toyota Camry Sedan \\
compact sedan & 2012 Acura TSX Sedan, 2012 Buick Verano Sedan, 2012 Chevrolet Sonic Sedan, 2007 Ford Focus Sedan, 2012 Ford Fiesta Sedan, 2007 Hyundai Elantra Sedan, 2012 Hyundai Accent Sedan, 2012 Mercedes-Benz C-Class Sedan, 2012 Mitsubishi Lancer Sedan, 2007 Suzuki Aerio Sedan, 2012 Suzuki Kizashi Sedan, 2012 Suzuki SX4 Sedan, 2012 Toyota Corolla Sedan, 1993 Volvo 240 Sedan \\
sports coupe & 2001 Acura Integra Type R, 2012 Audi A5 Coupe, 2012 Audi TTS Coupe, 2012 Audi R8 Coupe, 2012 Audi S5 Coupe, 2012 Audi TT RS Coupe, 2012 BMW 1 Series Coupe, 2012 BMW M3 Coupe, 2012 Chevrolet Corvette ZR1, 2007 Chevrolet Corvette Ron Fellows Edition Z06, 2010 Chevrolet Cobalt SS, 2011 Dodge Challenger SRT8, 2012 Infiniti G Coupe IPL, 2012 Jaguar XK XKR, 2009 Mercedes-Benz SL-Class Coupe \\
sport hatchback & 2012 Acura ZDX Hatchback, 2011 Audi TT Hatchback, 2012 Nissan Juke Hatchback \\
luxury convertible & 2012 Aston Martin V8 Vantage Convertible, 2012 Aston Martin Virage Convertible, 2012 Bentley Continental Supersports Conv. Convertible, 2012 Maybach Landaulet Convertible, 1993 Mercedes-Benz 300-Class Convertible, 2012 Rolls-Royce Phantom Drophead Coupe Convertible \\
exotic coupe & 2012 Aston Martin V8 Vantage Coupe, 2012 Aston Martin Virage Coupe, 2012 Bentley Continental GT Coupe, 2007 Bentley Continental GT Coupe, 2009 Bugatti Veyron 16.4 Coupe, 2012 Ferrari FF Coupe, 2012 Ferrari 458 Italia Coupe, 2006 Ford GT Coupe, 2008 Lamborghini Reventon Coupe, 2012 Lamborghini Aventador Coupe, 2012 Lamborghini Gallardo LP 570-4 Superleggera, 2001 Lamborghini Diablo Coupe, 2012 McLaren MP4-12C Coupe, 2009 Spyker C8 Coupe \\
sports convertible & 2008 Audi RS 4 Convertible, 2012 Audi S5 Convertible, 2007 BMW 6 Series Convertible, 2010 BMW M6 Convertible, 2012 BMW Z4 Convertible, 2009 Bugatti Veyron 16.4 Convertible, 2012 Chevrolet Corvette Convertible, 2012 Chevrolet Camaro Convertible, 2012 Ferrari California Convertible, 2012 Ferrari 458 Italia Convertible, 2007 Ford Mustang Convertible, 2009 Spyker C8 Convertible \\
wagon & 1994 Audi 100 Wagon, 2012 BMW 3 Series Wagon, 2002 Daewoo Nubira Wagon, 2012 Dodge Caliber Wagon, 2007 Dodge Caliber Wagon, 2008 Dodge Magnum Wagon, 2012 Ford E-Series Wagon Van \\
compact convertible & 2012 BMW 1 Series Convertible, 2010 Chrysler Sebring Convertible, 2008 Chrysler Crossfire Convertible, 2008 Chrysler PT Cruiser Convertible, 2012 FIAT 500 Convertible, 1993 Geo Metro Convertible, 2012 MINI Cooper Roadster Convertible, 2012 smart fortwo Convertible \\
luxury suv & 2007 BMW X5 SUV, 2012 BMW X6 SUV, 2012 BMW X3 SUV, 2012 Buick Enclave SUV, 2012 Cadillac SRX SUV, 2011 Infiniti QX56 SUV, 2012 Land Rover Range Rover SUV, 2012 Land Rover LR2 SUV, 2007 Volvo XC90 SUV \\
full-size pickup & 2007 Cadillac Escalade EXT Crew Cab, 2012 Chevrolet Silverado 1500 Hybrid Crew Cab, 2012 Chevrolet Avalanche Crew Cab, 2012 Chevrolet Silverado 2500HD Regular Cab, 2007 Chevrolet Silverado 1500 Classic Extended Cab, 2012 Chevrolet Silverado 1500 Extended Cab, 2012 Chevrolet Silverado 1500 Regular Cab, 2010 Dodge Ram Pickup 3500 Crew Cab, 2009 Dodge Ram Pickup 3500 Quad Cab, 2012 Ford F-450 Super Duty Crew Cab, 2012 Ford F-150 Regular Cab, 2007 Ford F-150 Regular Cab \\
minivan & 2010 Chevrolet HHR SS, 2012 Chrysler Town and Country Minivan, 1997 Dodge Caravan Minivan, 2007 Ford Freestar Minivan, 2012 Honda Odyssey Minivan, 2007 Honda Odyssey Minivan, 2012 Ram C/V Cargo Van Minivan \\
van & 2007 Chevrolet Express Cargo Van, 2007 Chevrolet Express Van, 2009 Dodge Sprinter Cargo Van, 2012 GMC Savana Van, 2012 Mercedes-Benz Sprinter Van, 2012 Nissan NV Passenger Van \\
standard coupe & 2007 Chevrolet Monte Carlo Coupe, 2012 FIAT 500 Abarth, 2012 Honda Accord Coupe, 1998 Nissan 240SX Coupe, 1999 Plymouth Neon Coupe \\
compact suv & 2012 Dodge Journey SUV, 2012 Ford Edge SUV, 2012 GMC Terrain SUV, 2012 Hyundai Santa Fe SUV, 2012 Hyundai Tucson SUV, 2012 Jeep Patriot SUV, 2012 Jeep Wrangler SUV, 2012 Jeep Liberty SUV, 2012 Jeep Compass SUV, 2011 Mazda Tribute SUV \\
midsize pickup & 2010 Dodge Dakota Crew Cab, 2007 Dodge Dakota Club Cab, 2011 Ford Ranger SuperCab, 2012 GMC Canyon Extended Cab, 2010 HUMMER H3T Crew Cab, 2009 HUMMER H2 SUT Crew Cab \\
compact hatchback & 1998 Eagle Talon Hatchback, 2012 Hyundai Veloster Hatchback, 2012 Hyundai Elantra Touring Hatchback, 2012 Nissan Leaf Hatchback, 2012 Scion xD Hatchback, 2012 Suzuki SX4 Hatchback, 2012 Volkswagen Golf Hatchback, 1991 Volkswagen Golf Hatchback, 2012 Volkswagen Beetle Hatchback, 2012 Volvo C30 Hatchback \\
\bottomrule
\end{tabular}
\caption{StanfordCars hierarchy: 196 classes grouped under 19 parents (mean 10.3 children per parent).}
\label{tab:hier_stanfordcars}
\end{table}

\begin{table}[H]
\centering
\footnotesize
\renewcommand{\arraystretch}{1.2}
\begin{tabular}{@{}>{\bfseries}l p{0.72\columnwidth}@{}}
\toprule
\textnormal{\textbf{Parent}} & \textbf{Classes} \\
\midrule
grooming & Apply Eye Makeup, Apply Lipstick, Blow Dry Hair, Brushing Teeth, Haircut, Head Massage, Shaving Beard \\
target and aim sport & Archery, Billiards, Bowling, Frisbee Catch, Golf Swing \\
hobby & Baby Crawling, Knitting \\
gymnastics apparatus & Balance Beam, Floor Gymnastics, Handstand Pushups, Handstand Walking, Parallel Bars, Pommel Horse, Still Rings, Trampoline Jumping, Uneven Bars \\
dance or parade & Band Marching, Hula Hoop, Ice Dancing, Military Parade, Salsa Spin \\
racket and bat sport & Baseball Pitch, Cricket Bowling, Cricket Shot, Table Tennis Shot, Tennis Swing \\
team ball sport & Basketball, Basketball Dunk, Field Hockey Penalty, Soccer Juggling, Soccer Penalty, Volleyball Spiking \\
bodyweight exercise & Bench Press, Body Weight Squats, Clean And Jerk, Jumping Jack, Jump Rope, Lunges, Pull Ups, Push Ups, Tai Chi, Wall Pushups \\
outdoor adventure & Biking, Rock Climbing Indoor, Rope Climbing, Skate Boarding, Swing \\
cooking & Blowing Candles, Cutting In Kitchen, Mixing, Pizza Tossing \\
combat sport & Boxing Punching Bag, Boxing Speed Bag, Fencing, Nunchucks, Punch, Sumo Wrestling \\
water or snow sport & Breast Stroke, Cliff Diving, Diving, Front Crawl, Kayaking, Rafting, Rowing, Skiing, Skijet, Sky Diving, Surfing \\
musical performance & Drumming, Playing Cello, Playing Daf, Playing Dhol, Playing Flute, Playing Guitar, Playing Piano, Playing Sitar, Playing Tabla, Playing Violin \\
cleaning task & Hammering, Mopping Floor \\
track and field & Hammer Throw, High Jump, Javelin Throw, Long Jump, Pole Vault, Shotput, Throw Discus \\
equestrian sport & Horse Race, Horse Riding, Walking With Dog \\
skill toy & Juggling Balls, Yo Yo \\
writing & Typing, Writing On Board \\
\bottomrule
\end{tabular}
\caption{UCF101 hierarchy: 101 classes grouped under 18 parents (mean 5.6 children per parent).}
\label{tab:hier_ucf101}
\end{table}

\section{HyCoCLIP$^\dagger$}
\label{app:hier_hyco}

\begin{figure}[H]
    \centering
    \begin{tikzpicture}[
        box/.style={
            draw,
            rounded corners,
            align=center,
            inner sep=4pt,
            font=\scriptsize,
            text width=5.8cm
        }
    ]

    \node[box] (p1) {
        \textbf{1.} \texttt{a photo of a \{class\}.}
    };

    \node[box, below=9mm of p1] (p2) {
        \textbf{2.} \texttt{a photo of a \{class\}, a \colorbox{yellow!40}{\{parent\}}.}
    };

    \node[box, below=3mm of p2] (p3) {
        \textbf{3.} \texttt{\colorbox{yellow!40}{\{parent\}}: a photo of a \{class\}.}
    };

    \node[box, below=3mm of p3] (p4) {
        \textbf{4.} \texttt{a photo of a \{class\}, a type of \colorbox{yellow!40}{\{parent\}}.}
    };

    \node[draw, dashed, rounded corners, fit=(p1), inner sep=4pt,
          label={[font=\scriptsize\bfseries]above:Instance}] {};

    \node[draw, dashed, rounded corners, fit=(p2)(p4), inner sep=4pt,
          label={[font=\scriptsize\bfseries]above:Instance + Parent}] {};

    \end{tikzpicture}
    \caption{
    The four hand-written templates instantiated for HyCoCLIP$^\dagger$. Template~1 is the
    standard instance-only prompt used by the zero-shot HyCoCLIP backbone. Templates~2-4
    additionally insert the ground-truth parent name (\colorbox{yellow!40}{highlighted})
    in three syntactic positions, giving HyCoCLIP$^\dagger$ access to the same
    hierarchy as our learned methods, but through fixed text rather than optimization.
    }
    \label{fig:hycoclip_dagger_templates}
\end{figure}
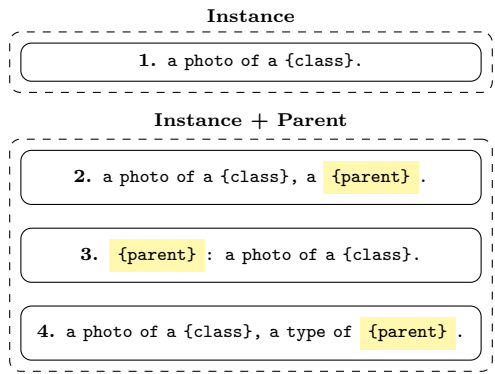

HyCoCLIP$^\dagger$ is the zero-shot HyCoCLIP backbone augmented with fixed, hand-written templates that inject the parent name directly into the text prompt, rather than learning it. For each class we encode all four templates in Figure~\ref{fig:hycoclip_dagger_templates} and average the resulting text embeddings before computing similarity with the image. This gives the backbone access to the same parent-class hierarchy used by HyPLO, CoHyPLO and MaHyPLO, isolating whether the gains observed in Section~\ref{sec:hier_metrics} require learned, optimized prompts or can be obtained from hierarchy alone, expressed through static text.

\section{Hierarchical Metrics}
\label{app:hier_extended}

This section details the hierarchical evaluation metrics used in our classification experiments. Given a predicted class $\hat{y}$ and its ground-truth class $y$, the \emph{Tree-Induced Error} (TIE) \citep{dekel2004large} measures their graph distance:
\begin{equation}
    \mathrm{TIE}(\hat{y}, y)
    =
    \sum_{e \in E(\hat{y}, y)} w_e,
\end{equation}
where $E(i,j)$ denotes the set of edges along the path between nodes $i$ and $j$, and $w_e$ is the weight of edge $e$. We assign unit weight to all edges in every hierarchy, i.e., $w_e = 1$.

The \emph{Lowest Common Ancestor} (LCA) error measures the distance from the predicted and ground-truth classes to their deepest shared ancestor in the hierarchy. It thus captures the extent to which a prediction departs from the correct branch of the taxonomy.

For set-based metrics, let $\hat{Y}_{\mathrm{anc}}$ and $Y_{\mathrm{anc}}$ denote the sets of ancestor nodes of $\hat{y}$ and $y$ respectively. We compute the Jaccard similarity $J$, hierarchical precision $P_H$, hierarchical recall $R_H$, and hierarchical F1 score $F1_H$ \citep{kosmopoulos2015evaluation} as
\begin{equation}
\begin{aligned}
    J &=
    \frac{\left|\hat{Y}_{\mathrm{anc}} \cap Y_{\mathrm{anc}}\right|}
         {\left|\hat{Y}_{\mathrm{anc}} \cup Y_{\mathrm{anc}}\right|},
    &
    P_H &=
    \frac{\left|\hat{Y}_{\mathrm{anc}} \cap Y_{\mathrm{anc}}\right|}
         {\left|\hat{Y}_{\mathrm{anc}}\right|}, \\
    R_H &=
    \frac{\left|\hat{Y}_{\mathrm{anc}} \cap Y_{\mathrm{anc}}\right|}
         {\left|Y_{\mathrm{anc}}\right|},
    &
    F1_H &=
    \frac{2P_H R_H}{P_H + R_H}.
\end{aligned}
\label{eq:hierarchical_metrics}
\end{equation}

We additionally report sibling and cousin accuracy, which assess whether the predicted and ground-truth classes share an ancestor at a fixed hierarchical level. Sibling accuracy measures the proportion of predictions that share at least one direct parent with the ground-truth class. Cousin accuracy measures the proportion of predictions that share at least one grandparent with the ground-truth class. Exact predictions are included in both metrics.

\section{Per-Dataset Hierarchical Metrics}
\label{app:hier_metrics}

Section~\ref{sec:hier_metrics} averages the six hierarchical metrics across eleven datasets. Tables~\ref{tab:hier_metrics_cal_b2n}-\ref{tab:hier_metrics_ucf101_b2n} give the per-dataset breakdown. Marking conventions match the main text: underline is the best value per column, bold is the better method within each baseline/extension pair.

% ImageNet
\begin{table}[H]
    \centering
    \small
    \setlength{\tabcolsep}{6pt}
    \setlength{\aboverulesep}{0pt}
    \setlength{\belowrulesep}{0pt}
    \renewcommand{\arraystretch}{1.15}
    \begin{tabular}{l c c c c c c c}
    \toprule
    \multirow{2}{*}{\textbf{Method}}
    & \multirow{2}{*}{\textbf{Hier.}}
    & \multicolumn{2}{c}{\cellcolor{greenfirst}\textbf{Hierarchical Distance}}
    & \multicolumn{2}{c}{\cellcolor{greenalt}\textbf{Ancestor Overlap}}
    & \multicolumn{2}{c}{\cellcolor{greencol}\textbf{Local Accuracy}} \\
    \cmidrule(lr){3-4}
    \cmidrule(lr){5-6}
    \cmidrule(lr){7-8}
    &
    & \cellcolor{greenfirst}\textbf{TIE} \ensuremath{\downarrow}
    & \cellcolor{greenfirst}\textbf{LCA} \ensuremath{\downarrow}
    & \cellcolor{greenalt}\ensuremath{\mathbf{J}} \ensuremath{\uparrow}
    & \cellcolor{greenalt}\ensuremath{\mathbf{F1_H}} \ensuremath{\uparrow}
    & \cellcolor{greencol}\textbf{S-Acc} \ensuremath{\uparrow}
    & \cellcolor{greencol}\textbf{C-Acc} \ensuremath{\uparrow} \\
    \midrule
    \addlinespace[4pt]
    HyCoCLIP & \textbf{--}
    & \bn{2.682}{3.191} & \bn{1.840}{2.119} & \bn{0.871}{0.783}
    & \bn{0.920}{0.849} & \bn{0.584}{0.577} & \bn{0.644}{0.596} \\
    \addlinespace[4pt]
    HyCoCLIP$^\dagger$ & \checkmark
    & \bn{2.645}{3.065} & \bn{1.784}{\underline{2.041}} & \bn{0.874}{0.792}
    & \bn{0.922}{0.857} & \bn{0.585}{0.583} & \bn{0.643}{0.604} \\
    \midrule
    \addlinespace[2pt]
    + CoOp & \textbf{--}
    & \bn{2.485}{3.302} & \bn{1.731}{2.144}
    & \bn{0.882}{0.776} & \bn{0.927}{0.844}
    & \bn{0.608}{0.557} & \bn{0.666}{0.579} \\
    \addlinespace[4pt]
    \rowcolor{gray!10}
    \textbf{+ HyPLO} & \checkmark
    & \bn{\textbf{2.427}}{\textbf{3.199}} & \bn{\textbf{1.719}}{\textbf{2.139}}
    & \bn{\textbf{0.886}}{\textbf{0.785}} & \bn{\textbf{0.930}}{\textbf{0.850}}
    & \bn{\textbf{0.616}}{\textbf{0.571}} & \bn{\textbf{0.674}}{\textbf{0.591}} \\
    \midrule
    \addlinespace[2pt]
    + CoCoOp & \textbf{--}
    & \bn{2.491}{3.225} & \bn{1.737}{2.124}
    & \bn{0.881}{0.782} & \bn{0.927}{0.847}
    & \bn{0.608}{0.570} & \bn{0.665}{0.592} \\
    \addlinespace[4pt]
    \rowcolor{gray!10}
    \textbf{+ CoHyPLO} & \checkmark
    & \bn{\textbf{2.459}}{\textbf{3.091}} & \bn{\textbf{1.730}}{\textbf{2.100}}
    & \bn{\textbf{0.885}}{\textbf{0.792}} & \bn{\textbf{0.929}}{\textbf{0.856}}
    & \bn{\textbf{0.613}}{\textbf{0.586}} & \bn{\textbf{0.671}}{\textbf{0.608}} \\
    \midrule
    \addlinespace[2pt]
    + MaPLe & \textbf{--}
    & \bn{2.412}{3.187} & \bn{1.709}{2.102}
    & \bn{0.885}{0.783} & \bn{0.929}{0.849}
    & \bn{0.617}{0.571} & \bn{0.674}{0.592} \\
    \addlinespace[4pt]
    \rowcolor{gray!10}
    \textbf{+ MaHyPLO} & \checkmark
    & \bn{\underline{\textbf{2.349}}}{\underline{\textbf{3.051}}} & \bn{\underline{\textbf{1.686}}}{\textbf{2.079}}
    & \bn{\underline{\textbf{0.890}}}{\underline{\textbf{0.795}}} & \bn{\underline{\textbf{0.932}}}{\underline{\textbf{0.858}}}
    & \bn{\underline{\textbf{0.629}}}{\underline{\textbf{0.589}}} & \bn{\underline{\textbf{0.684}}}{\underline{\textbf{0.610}}} \\
    \bottomrule
    \end{tabular}
    \caption{
    Hierarchical evaluation metrics on ImageNet. Each cell reports \textbf{Base} (top) and \textbf{New} (bottom), averaged over three seeds. Hierarchical F1 is computed as the harmonic mean of hierarchical precision and hierarchical recall for each dataset. \underline{Underline}: best in column. \textbf{Bold}: best within each baseline/extension pair (CoOp vs.\ HyPLO; CoCoOp vs.\ CoHyPLO; MaPLe vs.\ MaHyPLO). \ensuremath{\downarrow} lower is better; \ensuremath{\uparrow} higher is better. \textit{Hier.}: explicitly hierarchy-aware.
    }
    \label{tab:hier_metrics_imagenet_b2n}
\end{table}

% Caltech101
\begin{table}[H]
    \centering
    \small
    \setlength{\tabcolsep}{6pt}
    \setlength{\aboverulesep}{0pt}
    \setlength{\belowrulesep}{0pt}
    \renewcommand{\arraystretch}{1.15}
    \begin{tabular}{l c c c c c c c}
    \toprule
    \multirow{2}{*}{\textbf{Method}}
    & \multirow{2}{*}{\textbf{Hier.}}
    & \multicolumn{2}{c}{\cellcolor{greenfirst}\textbf{Hierarchical Distance}}
    & \multicolumn{2}{c}{\cellcolor{greenalt}\textbf{Ancestor Overlap}}
    & \multicolumn{2}{c}{\cellcolor{greencol}\textbf{Local Accuracy}} \\
    \cmidrule(lr){3-4}
    \cmidrule(lr){5-6}
    \cmidrule(lr){7-8}
    &
    & \cellcolor{greenfirst}\textbf{TIE} $\downarrow$
    & \cellcolor{greenfirst}\textbf{LCA} $\downarrow$
    & \cellcolor{greenalt}$\mathbf{J}$ $\uparrow$
    & \cellcolor{greenalt}$\mathbf{F1_H}$ $\uparrow$
    & \cellcolor{greencol}\textbf{S-Acc} $\uparrow$
    & \cellcolor{greencol}\textbf{C-Acc} $\uparrow$ \\
    \midrule
    \addlinespace[4pt]
    HyCoCLIP & \textbf{--}
    & \bn{0.240}{\underline{0.303}} & \bn{1.049}{\underline{1.062}} & \bn{0.978}{\underline{0.971}} & \bn{0.990}{\underline{0.979}} & \bn{0.964}{\underline{0.949}} & \bn{0.994}{0.989} \\
    \addlinespace[4pt]
    HyCoCLIP$^\dagger$ & \checkmark
    & \bn{0.234}{0.341} & \bn{1.049}{1.068} & \bn{0.978}{0.968} & \bn{0.984}{0.977} & \bn{0.963}{0.943} & \bn{0.988}{0.989} \\
    \midrule
    \addlinespace[2pt]
    + CoOp & \textbf{--}
    & \bn{\textbf{0.148}}{0.545} & \bn{\textbf{1.030}}{1.103}
    & \bn{\textbf{0.986}}{0.952} & \bn{\textbf{0.990}}{0.966}
    & \bn{0.976}{0.915} & \bn{\textbf{0.994}}{0.982} \\
    \addlinespace[4pt]
    \rowcolor{gray!10}
    \textbf{+ HyPLO} & \checkmark
    & \bn{0.153}{\textbf{0.491}} & \bn{\textbf{1.030}}{\textbf{1.088}}
    & \bn{\textbf{0.986}}{\textbf{0.958}} & \bn{\textbf{0.990}}{\textbf{0.971}}
    & \bn{\textbf{0.977}}{\textbf{0.923}} & \bn{0.993}{\textbf{0.990}} \\
    \midrule
    \addlinespace[2pt]
    + CoCoOp & \textbf{--}
    & \bn{0.163}{0.387} & \bn{1.032}{\textbf{1.075}}
    & \bn{0.985}{\textbf{0.965}} & \bn{0.989}{\textbf{0.975}}
    & \bn{0.975}{\textbf{0.937}} & \bn{\textbf{0.993}}{\textbf{0.989}} \\
    \addlinespace[4pt]
    \rowcolor{gray!10}
    \textbf{+ CoHyPLO} & \checkmark
    & \bn{\textbf{0.157}}{\textbf{0.367}} & \bn{\textbf{1.031}}{1.076}
    & \bn{\textbf{0.986}}{0.964} & \bn{\textbf{0.990}}{\textbf{0.975}}
    & \bn{\textbf{0.976}}{0.935} & \bn{\textbf{0.993}}{\textbf{0.989}} \\
    \midrule
    \addlinespace[2pt]
    + MaPLe & \textbf{--}
    & \bn{0.145}{\textbf{0.307}} & \bn{1.028}{\textbf{1.058}}
    & \bn{0.987}{\textbf{0.973}} & \bn{\underline{\textbf{0.991}}}{\textbf{0.981}}
    & \bn{0.979}{\textbf{0.950}} & \bn{0.993}{\underline{\textbf{0.992}}} \\
    \addlinespace[4pt]
    \rowcolor{gray!10}
    \textbf{+ MaHyPLO} & \checkmark
    & \bn{\underline{\textbf{0.135}}}{0.331} & \bn{\underline{\textbf{1.026}}}{1.068}
    & \bn{\underline{\textbf{0.988}}}{0.967} & \bn{\underline{\textbf{0.991}}}{0.977}
    & \bn{\underline{\textbf{0.980}}}{0.939} & \bn{\underline{\textbf{0.995}}}{\underline{\textbf{0.992}}} \\
    \bottomrule
    \end{tabular}
    \caption{
    Hierarchical evaluation metrics on Caltech101. Each cell reports \textbf{Base} (top) and \textbf{New} (bottom), averaged over three seeds. Hierarchical F1 is computed as the harmonic mean of hierarchical precision and hierarchical recall for each dataset. \underline{Underline}: best in column. \textbf{Bold}: best within each baseline/extension pair (CoOp vs.\ HyPLO; CoCoOp vs.\ CoHyPLO; MaPLe vs.\ MaHyPLO). \ensuremath{\downarrow} lower is better; \ensuremath{\uparrow} higher is better. \textit{Hier.}: explicitly hierarchy-aware.
    }
    \label{tab:hier_metrics_cal_b2n}
\end{table}

% OxfordPets
\begin{table}[H]
    \centering
    \small
    \setlength{\tabcolsep}{6pt}
    \setlength{\aboverulesep}{0pt}
    \setlength{\belowrulesep}{0pt}
    \renewcommand{\arraystretch}{1.15}
    \begin{tabular}{l c c c c c c c}
    \toprule
    \multirow{2}{*}{\textbf{Method}}
    & \multirow{2}{*}{\textbf{Hier.}}
    & \multicolumn{2}{c}{\cellcolor{greenfirst}\textbf{Hierarchical Distance}}
    & \multicolumn{2}{c}{\cellcolor{greenalt}\textbf{Ancestor Overlap}}
    & \multicolumn{2}{c}{\cellcolor{greencol}\textbf{Local Accuracy}} \\
    \cmidrule(lr){3-4}
    \cmidrule(lr){5-6}
    \cmidrule(lr){7-8}
    &
    & \cellcolor{greenfirst}\textbf{TIE} $\downarrow$
    & \cellcolor{greenfirst}\textbf{LCA} $\downarrow$
    & \cellcolor{greenalt}$\mathbf{J}$ $\uparrow$
    & \cellcolor{greenalt}$\mathbf{F1_H}$ $\uparrow$
    & \cellcolor{greencol}\textbf{S-Acc} $\uparrow$
    & \cellcolor{greencol}\textbf{C-Acc} $\uparrow$ \\
    \midrule
    \addlinespace[4pt]
    HyCoCLIP & \textbf{--}
    & \bn{1.427}{\underline{0.713}} & \bn{1.289}{1.100} & \bn{0.867}{0.952} & \bn{0.904}{0.967} & \bn{0.770}{0.908} & \bn{0.941}{0.993} \\
    \addlinespace[4pt]
    HyCoCLIP$^\dagger$ & \checkmark
    & \bn{1.147}{0.694} & \bn{1.178}{1.096} & \bn{0.912}{0.953} & \bn{0.941}{0.968} & \bn{0.825}{0.907} & \bn{\underline{0.997}}{\underline{0.998}} \\
    \midrule
    \addlinespace[2pt]
    + CoOp & \textbf{--}
    & \bn{0.671}{1.116} & \bn{1.101}{1.176}
    & \bn{0.950}{\textbf{0.914}} & \bn{0.966}{0.941}
    & \bn{0.903}{\textbf{0.833}} & \bn{0.996}{0.991} \\
    \addlinespace[4pt]
    \rowcolor{gray!10}
    \textbf{+ HyPLO} & \checkmark
    & \bn{\underline{\textbf{0.606}}}{\textbf{1.110}} & \bn{\underline{\textbf{1.089}}}{\textbf{1.173}}
    & \bn{\underline{\textbf{0.956}}}{\textbf{0.914}} & \bn{\underline{\textbf{0.970}}}{\textbf{0.942}}
    & \bn{\underline{\textbf{0.914}}}{0.831} & \bn{\underline{\textbf{0.997}}}{\textbf{0.996}} \\
    \midrule
    \addlinespace[2pt]
    + CoCoOp & \textbf{--}
    & \bn{0.693}{1.043} & \bn{1.104}{1.189}
    & \bn{0.949}{0.911} & \bn{0.965}{0.937}
    & \bn{0.901}{0.836} & \bn{0.995}{0.975} \\
    \addlinespace[4pt]
    \rowcolor{gray!10}
    \textbf{+ CoHyPLO} & \checkmark
    & \bn{\textbf{0.660}}{\textbf{0.902}} & \bn{\textbf{1.094}}{\textbf{1.136}}
    & \bn{\textbf{0.954}}{\textbf{0.933}} & \bn{\textbf{0.969}}{\textbf{0.955}}
    & \bn{\textbf{0.910}}{\textbf{0.870}} & \bn{\textbf{0.996}}{\textbf{0.994}} \\
    \midrule
    \addlinespace[2pt]
    + MaPLe & \textbf{--}
    & \bn{0.734}{0.830} & \bn{1.114}{1.125}
    & \bn{0.944}{0.939} & \bn{0.962}{0.958}
    & \bn{0.891}{0.883} & \bn{\textbf{0.996}}{0.992} \\
    \addlinespace[4pt]
    \rowcolor{gray!10}
    \textbf{+ MaHyPLO} & \checkmark
    & \bn{\textbf{0.650}}{\textbf{0.720}} & \bn{\textbf{1.102}}{\underline{\textbf{1.091}}}
    & \bn{\textbf{0.951}}{\underline{\textbf{0.955}}} & \bn{\textbf{0.966}}{\underline{\textbf{0.970}}}
    & \bn{\textbf{0.906}}{\underline{\textbf{0.910}}} & \bn{0.992}{\underline{\textbf{0.998}}} \\
    \bottomrule
    \end{tabular}
    \caption{
    Hierarchical evaluation metrics on OxfordPets. Each cell reports \textbf{Base} (top) and \textbf{New} (bottom), averaged over three seeds. Hierarchical F1 is computed as the harmonic mean of hierarchical precision and hierarchical recall for each dataset. \underline{Underline}: best in column. \textbf{Bold}: best within each baseline/extension pair (CoOp vs.\ HyPLO; CoCoOp vs.\ CoHyPLO; MaPLe vs.\ MaHyPLO). \ensuremath{\downarrow} lower is better; \ensuremath{\uparrow} higher is better. \textit{Hier.}: explicitly hierarchy-aware.
    }
    \label{tab:hier_metrics_oxford_pets_b2n}
\end{table}

% StanfordCars
\begin{table}[H]
    \centering
    \small
    \setlength{\tabcolsep}{6pt}
    \setlength{\aboverulesep}{0pt}
    \setlength{\belowrulesep}{0pt}
    \renewcommand{\arraystretch}{1.15}
    \begin{tabular}{l c c c c c c c}
    \toprule
    \multirow{2}{*}{\textbf{Method}}
    & \multirow{2}{*}{\textbf{Hier.}}
    & \multicolumn{2}{c}{\cellcolor{greenfirst}\textbf{Hierarchical Distance}}
    & \multicolumn{2}{c}{\cellcolor{greenalt}\textbf{Ancestor Overlap}}
    & \multicolumn{2}{c}{\cellcolor{greencol}\textbf{Local Accuracy}} \\
    \cmidrule(lr){3-4}
    \cmidrule(lr){5-6}
    \cmidrule(lr){7-8}
    &
    & \cellcolor{greenfirst}\textbf{TIE} $\downarrow$
    & \cellcolor{greenfirst}\textbf{LCA} $\downarrow$
    & \cellcolor{greenalt}$\mathbf{J}$ $\uparrow$
    & \cellcolor{greenalt}$\mathbf{F1_H}$ $\uparrow$
    & \cellcolor{greencol}\textbf{S-Acc} $\uparrow$
    & \cellcolor{greencol}\textbf{C-Acc} $\uparrow$ \\
    \midrule
    \addlinespace[4pt]
    HyCoCLIP & \textbf{--}
    & \bn{3.223}{3.132} & \bn{1.756}{1.761} & \bn{0.649}{0.651} & \bn{0.748}{0.746} & \bn{0.380}{0.398} & \bn{0.865}{0.841} \\
    \addlinespace[4pt]
    HyCoCLIP$^\dagger$ & \checkmark
    & \bn{3.157}{3.075} & \bn{1.732}{1.740} & \bn{0.660}{0.662} & \bn{0.756}{0.753} & \bn{0.398}{0.421} & \bn{0.870}{0.838} \\
    \midrule
    \addlinespace[2pt]
    + CoOp & \textbf{--}
    & \bn{2.627}{3.148} & \bn{1.588}{1.730}
    & \bn{0.725}{0.661} & \bn{0.804}{0.757}
    & \bn{0.505}{0.400} & \bn{0.906}{\textbf{0.870}} \\
    \addlinespace[4pt]
    \rowcolor{gray!10}
    \textbf{+ HyPLO} & \checkmark
    & \bn{\underline{\textbf{2.625}}}{\textbf{3.069}} & \bn{\textbf{1.575}}{\textbf{1.704}}
    & \bn{\underline{\textbf{0.731}}}{\textbf{0.675}} & \bn{\textbf{0.808}}{\textbf{0.766}}
    & \bn{\underline{\textbf{0.517}}}{\textbf{0.430}} & \bn{\textbf{0.909}}{0.867} \\
    \midrule
    \addlinespace[2pt]
    + CoCoOp & \textbf{--}
    & \bn{2.877}{2.988} & \bn{1.644}{1.695}
    & \bn{0.698}{0.678} & \bn{0.785}{0.768}
    & \bn{0.458}{0.434} & \bn{0.898}{\textbf{0.871}} \\
    \addlinespace[4pt]
    \rowcolor{gray!10}
    \textbf{+ CoHyPLO} & \checkmark
    & \bn{\textbf{2.854}}{\textbf{2.945}} & \bn{\textbf{1.625}}{\textbf{1.682}}
    & \bn{\textbf{0.707}}{\textbf{0.687}} & \bn{\textbf{0.791}}{\textbf{0.773}}
    & \bn{\textbf{0.472}}{\underline{\textbf{0.457}}} & \bn{\textbf{0.902}}{0.861} \\
    \midrule
    \addlinespace[2pt]
    + MaPLe & \textbf{--}
    & \bn{2.754}{2.961} & \bn{1.605}{1.678}
    & \bn{0.716}{0.686} & \bn{0.798}{0.774}
    & \bn{0.487}{0.447} & \bn{0.908}{\underline{\textbf{0.875}}} \\
    \addlinespace[4pt]
    \rowcolor{gray!10}
    \textbf{+ MaHyPLO} & \checkmark
    & \bn{\textbf{2.669}}{\underline{\textbf{2.940}}} & \bn{\textbf{1.571}}{\underline{\textbf{1.675}}}
    & \bn{\underline{\textbf{0.731}}}{\underline{\textbf{0.688}}} & \bn{\underline{\textbf{0.810}}}{\underline{\textbf{0.775}}}
    & \bn{\textbf{0.514}}{\textbf{0.455}} & \bn{\underline{\textbf{0.915}}}{0.870} \\
    \bottomrule
    \end{tabular}
    \caption{
    Hierarchical evaluation metrics on StanfordCars. Each cell reports \textbf{Base} (top) and \textbf{New} (bottom), averaged over three seeds. Hierarchical F1 is computed as the harmonic mean of hierarchical precision and hierarchical recall for each dataset. \underline{Underline}: best in column. \textbf{Bold}: best within each baseline/extension pair (CoOp vs.\ HyPLO; CoCoOp vs.\ CoHyPLO; MaPLe vs.\ MaHyPLO). \ensuremath{\downarrow} lower is better; \ensuremath{\uparrow} higher is better. \textit{Hier.}: explicitly hierarchy-aware.
    }
    \label{tab:hier_metrics_stanford_cars_b2n}
\end{table}

%oxford flowers
\begin{table}[H]
    \centering
    \small
    \setlength{\tabcolsep}{6pt}
    \setlength{\aboverulesep}{0pt}
    \setlength{\belowrulesep}{0pt}
    \renewcommand{\arraystretch}{1.15}
    \begin{tabular}{l c c c c c c c}
    \toprule
    \multirow{2}{*}{\textbf{Method}}
    & \multirow{2}{*}{\textbf{Hier.}}
    & \multicolumn{2}{c}{\cellcolor{greenfirst}\textbf{Hierarchical Distance}}
    & \multicolumn{2}{c}{\cellcolor{greenalt}\textbf{Ancestor Overlap}}
    & \multicolumn{2}{c}{\cellcolor{greencol}\textbf{Local Accuracy}} \\
    \cmidrule(lr){3-4}
    \cmidrule(lr){5-6}
    \cmidrule(lr){7-8}
    &
    & \cellcolor{greenfirst}\textbf{TIE} $\downarrow$
    & \cellcolor{greenfirst}\textbf{LCA} $\downarrow$
    & \cellcolor{greenalt}$\mathbf{J}$ $\uparrow$
    & \cellcolor{greenalt}$\mathbf{F1_H}$ $\uparrow$
    & \cellcolor{greencol}\textbf{S-Acc} $\uparrow$
    & \cellcolor{greencol}\textbf{C-Acc} $\uparrow$ \\
    \midrule
    \addlinespace[4pt]
    HyCoCLIP & \textbf{--}
    & \bn{2.498}{2.762} & \bn{1.616}{1.762} & \bn{0.727}{0.677} & \bn{0.795}{0.746} & \bn{0.559}{0.526} & \bn{0.824}{0.713} \\
    \addlinespace[4pt]
    HyCoCLIP$^\dagger$ & \checkmark
    & \bn{2.557}{\underline{2.738}} & \bn{1.639}{\underline{1.741}} & \bn{0.718}{\underline{0.685}} & \bn{0.787}{\underline{0.753}} & \bn{0.548}{\underline{0.537}} & \bn{0.813}{\underline{0.722}} \\
    \midrule
    \addlinespace[2pt]
    + CoOp & \textbf{--}
    & \bn{\underline{\textbf{0.874}}}{\textbf{3.565}} & \bn{1.225}{\textbf{2.003}}
    & \bn{\underline{\textbf{0.902}}}{\textbf{0.574}} & \bn{0.925}{\textbf{0.666}}
    & \bn{\underline{\textbf{0.848}}}{\textbf{0.374}} & \bn{0.927}{\textbf{0.623}} \\
    \addlinespace[4pt]
    \rowcolor{gray!10}
    \textbf{+ HyPLO} & \checkmark
    & \bn{0.889}{3.704} & \bn{\underline{\textbf{1.222}}}{2.049}
    & \bn{\underline{\textbf{0.902}}}{0.557} & \bn{\underline{\textbf{0.926}}}{0.650}
    & \bn{0.845}{0.358} & \bn{\underline{\textbf{0.933}}}{0.593} \\
    \midrule
    \addlinespace[2pt]
    + CoCoOp & \textbf{--}
    & \bn{1.630}{3.204} & \bn{1.411}{1.870}
    & \bn{0.818}{0.631} & \bn{0.863}{0.710}
    & \bn{0.706}{0.458} & \bn{\textbf{0.883}}{0.672} \\
    \addlinespace[4pt]
    \rowcolor{gray!10}
    \textbf{+ CoHyPLO} & \checkmark
    & \bn{\textbf{1.606}}{\textbf{3.087}} & \bn{\textbf{1.394}}{\textbf{1.841}}
    & \bn{\textbf{0.826}}{\textbf{0.642}} & \bn{\textbf{0.869}}{\textbf{0.720}}
    & \bn{\textbf{0.724}}{\textbf{0.471}} & \bn{\textbf{0.883}}{\textbf{0.688}} \\
    \midrule
    \addlinespace[2pt]
    + MaPLe & \textbf{--}
    & \bn{1.461}{2.962} & \bn{1.370}{1.802}
    & \bn{0.837}{0.658} & \bn{0.877}{0.732}
    & \bn{0.743}{0.495} & \bn{0.886}{0.702} \\
    \addlinespace[4pt]
    \rowcolor{gray!10}
    \textbf{+ MaHyPLO} & \checkmark
    & \bn{\textbf{1.337}}{\textbf{2.824}} & \bn{\textbf{1.326}}{\textbf{1.760}}
    & \bn{\textbf{0.857}}{\textbf{0.676}} & \bn{\textbf{0.891}}{\textbf{0.747}}
    & \bn{\textbf{0.772}}{\textbf{0.522}} & \bn{\textbf{0.901}}{\textbf{0.718}} \\
    \bottomrule
    \end{tabular}
    \caption{
    Hierarchical evaluation metrics on OxfordFlowers. Each cell reports \textbf{Base} (top) and \textbf{New} (bottom), averaged over three seeds. Hierarchical F1 is computed as the harmonic mean of hierarchical precision and hierarchical recall for each dataset. \underline{Underline}: best in column. \textbf{Bold}: best within each baseline/extension pair (CoOp vs.\ HyPLO; CoCoOp vs.\ CoHyPLO; MaPLe vs.\ MaHyPLO). \ensuremath{\downarrow} lower is better; \ensuremath{\uparrow} higher is better. \textit{Hier.}: explicitly hierarchy-aware.
    }
    \label{tab:hier_metrics_oxford_flowers_b2n}
\end{table}

% Food101
\begin{table}[H]
    \centering
    \small
    \setlength{\tabcolsep}{6pt}
    \setlength{\aboverulesep}{0pt}
    \setlength{\belowrulesep}{0pt}
    \renewcommand{\arraystretch}{1.15}
    \begin{tabular}{l c c c c c c c}
    \toprule
    \multirow{2}{*}{\textbf{Method}}
    & \multirow{2}{*}{\textbf{Hier.}}
    & \multicolumn{2}{c}{\cellcolor{greenfirst}\textbf{Hierarchical Distance}}
    & \multicolumn{2}{c}{\cellcolor{greenalt}\textbf{Ancestor Overlap}}
    & \multicolumn{2}{c}{\cellcolor{greencol}\textbf{Local Accuracy}} \\
    \cmidrule(lr){3-4}
    \cmidrule(lr){5-6}
    \cmidrule(lr){7-8}
    &
    & \cellcolor{greenfirst}\textbf{TIE} $\downarrow$
    & \cellcolor{greenfirst}\textbf{LCA} $\downarrow$
    & \cellcolor{greenalt}$\mathbf{J}$ $\uparrow$
    & \cellcolor{greenalt}$\mathbf{F1_H}$ $\uparrow$
    & \cellcolor{greencol}\textbf{S-Acc} $\uparrow$
    & \cellcolor{greencol}\textbf{C-Acc} $\uparrow$ \\
    \midrule
    \addlinespace[4pt]
    HyCoCLIP & \textbf{--}
    & \bn{1.425}{1.134} & \bn{1.381}{1.279} & \bn{0.835}{0.877} & \bn{0.873}{0.907} & \bn{0.745}{0.803} & \bn{0.874}{0.918} \\
    \addlinespace[4pt]
    HyCoCLIP$^\dagger$ & \checkmark
    & \bn{1.437}{1.185} & \bn{1.384}{1.291} & \bn{0.832}{0.872} & \bn{0.872}{0.903} & \bn{0.737}{0.795} & \bn{0.879}{0.914} \\
    \midrule
    \addlinespace[2pt]
    + CoOp & \textbf{--}
    & \bn{1.177}{1.513} & \bn{1.311}{1.361}
    & \bn{0.865}{0.839} & \bn{0.896}{0.880}
    & \bn{0.794}{0.736} & \bn{0.895}{0.903} \\
    \addlinespace[4pt]
    \rowcolor{gray!10}
    \textbf{+ HyPLO} & \checkmark
    & \bn{\textbf{1.109}}{\textbf{1.419}} & \bn{\underline{\textbf{1.288}}}{\textbf{1.328}}
    & \bn{\underline{\textbf{0.875}}}{\textbf{0.854}} & \bn{\underline{\textbf{0.904}}}{\textbf{0.891}}
    & \bn{\textbf{0.807}}{\textbf{0.763}} & \bn{\underline{\textbf{0.905}}}{\textbf{0.909}} \\
    \midrule
    \addlinespace[2pt]
    + CoCoOp & \textbf{--}
    & \bn{1.177}{1.286} & \bn{1.311}{1.307}
    & \bn{0.865}{0.863} & \bn{0.896}{0.898}
    & \bn{0.793}{0.778} & \bn{0.896}{0.914} \\
    \addlinespace[4pt]
    \rowcolor{gray!10}
    \textbf{+ CoHyPLO} & \checkmark
    & \bn{\textbf{1.143}}{\textbf{1.182}} & \bn{\textbf{1.297}}{\textbf{1.286}}
    & \bn{\textbf{0.871}}{\textbf{0.874}} & \bn{\textbf{0.901}}{\textbf{0.905}}
    & \bn{\textbf{0.800}}{\textbf{0.796}} & \bn{\textbf{0.902}}{\textbf{0.918}} \\
    \midrule
    \addlinespace[2pt]
    + MaPLe & \textbf{--}
    & \bn{1.133}{1.146} & \bn{1.299}{1.276}
    & \bn{0.871}{0.878} & \bn{0.900}{0.908}
    & \bn{0.801}{0.805} & \bn{0.900}{0.919} \\
    \addlinespace[4pt]
    \rowcolor{gray!10}
    \textbf{+ MaHyPLO} & \checkmark
    & \bn{\underline{\textbf{1.106}}}{\underline{\textbf{1.125}}} & \bn{\underline{\textbf{1.288}}}{\underline{\textbf{1.273}}}
    & \bn{\underline{\textbf{0.875}}}{\underline{\textbf{0.880}}} & \bn{\underline{\textbf{0.904}}}{\underline{\textbf{0.909}}}
    & \bn{\underline{\textbf{0.808}}}{\underline{\textbf{0.808}}} & \bn{\textbf{0.904}}{\underline{\textbf{0.920}}} \\
    \bottomrule
    \end{tabular}
    \caption{
    Hierarchical evaluation metrics on Food101. Each cell reports \textbf{Base} (top) and \textbf{New} (bottom), averaged over three seeds. Hierarchical F1 is computed as the harmonic mean of hierarchical precision and hierarchical recall for each dataset. \underline{Underline}: best in column. \textbf{Bold}: best within each baseline/extension pair (CoOp vs.\ HyPLO; CoCoOp vs.\ CoHyPLO; MaPLe vs.\ MaHyPLO). \ensuremath{\downarrow} lower is better; \ensuremath{\uparrow} higher is better. \textit{Hier.}: explicitly hierarchy-aware.
    }
    \label{tab:hier_metrics_food101_b2n}
\end{table}

% Aircraft
\begin{table}[H]
    \centering
    \small
    \setlength{\tabcolsep}{6pt}
    \setlength{\aboverulesep}{0pt}
    \setlength{\belowrulesep}{0pt}
    \renewcommand{\arraystretch}{1.15}
    \begin{tabular}{l c c c c c c c}
    \toprule
    \multirow{2}{*}{\textbf{Method}}
    & \multirow{2}{*}{\textbf{Hier.}}
    & \multicolumn{2}{c}{\cellcolor{greenfirst}\textbf{Hierarchical Distance}}
    & \multicolumn{2}{c}{\cellcolor{greenalt}\textbf{Ancestor Overlap}}
    & \multicolumn{2}{c}{\cellcolor{greencol}\textbf{Local Accuracy}} \\
    \cmidrule(lr){3-4}
    \cmidrule(lr){5-6}
    \cmidrule(lr){7-8}
    &
    & \cellcolor{greenfirst}\textbf{TIE} $\downarrow$
    & \cellcolor{greenfirst}\textbf{LCA} $\downarrow$
    & \cellcolor{greenalt}$\mathbf{J}$ $\uparrow$
    & \cellcolor{greenalt}$\mathbf{F1_H}$ $\uparrow$
    & \cellcolor{greencol}\textbf{S-Acc} $\uparrow$
    & \cellcolor{greencol}\textbf{C-Acc} $\uparrow$ \\
    \midrule
    \addlinespace[4pt]
    HyCoCLIP & \textbf{--}
    & \bn{3.611}{3.921} & \bn{1.834}{2.026} & \bn{0.602}{0.561} & \bn{0.722}{0.658} & \bn{0.262}{0.345} & \bn{0.904}{0.629} \\
    \addlinespace[4pt]
    HyCoCLIP$^\dagger$ & \checkmark
    & \bn{3.396}{3.935} & \bn{1.737}{2.041} & \bn{0.651}{0.552} & \bn{0.754}{0.653} & \bn{0.359}{0.325} & \bn{0.904}{0.634} \\
    \midrule
    \addlinespace[2pt]
    + CoOp & \textbf{--}
    & \bn{2.976}{4.190} & \bn{1.630}{2.151}
    & \bn{0.700}{0.503} & \bn{0.790}{0.616}
    & \bn{0.443}{0.244} & \bn{0.927}{0.606} \\
    \addlinespace[4pt]
    \rowcolor{gray!10}
    \textbf{+ HyPLO} & \checkmark
    & \bn{\underline{\textbf{2.773}}}{\textbf{4.109}} & \bn{\underline{\textbf{1.539}}}{\textbf{2.118}}
    & \bn{\underline{\textbf{0.745}}}{\textbf{0.526}} & \bn{\underline{\textbf{0.820}}}{\textbf{0.627}}
    & \bn{\underline{\textbf{0.532}}}{\textbf{0.305}} & \bn{\underline{\textbf{0.929}}}{\textbf{0.577}} \\
    \midrule
    \addlinespace[2pt]
    + CoCoOp & \textbf{--}
    & \bn{3.186}{4.291} & \bn{1.696}{2.205}
    & \bn{0.670}{0.491} & \bn{0.768}{0.598}
    & \bn{0.397}{0.264} & \bn{0.906}{0.531} \\
    \addlinespace[4pt]
    \rowcolor{gray!10}
    \textbf{+ CoHyPLO} & \checkmark
    & \bn{\textbf{3.016}}{\textbf{3.829}} & \bn{\textbf{1.611}}{\textbf{1.986}}
    & \bn{\textbf{0.710}}{\textbf{0.570}} & \bn{\textbf{0.796}}{\textbf{0.671}}
    & \bn{\textbf{0.465}}{\textbf{0.331}} & \bn{\textbf{0.924}}{\underline{\textbf{0.683}}} \\
    \midrule
    \addlinespace[2pt]
    + MaPLe & \textbf{--}
    & \bn{3.254}{4.334} & \bn{1.708}{2.228}
    & \bn{0.664}{0.485} & \bn{0.764}{0.591}
    & \bn{0.381}{0.269} & \bn{0.911}{0.503} \\
    \addlinespace[4pt]
    \rowcolor{gray!10}
    \textbf{+ MaHyPLO} & \checkmark
    & \bn{\textbf{2.945}}{\underline{\textbf{3.707}}} & \bn{\textbf{1.589}}{\underline{\textbf{1.933}}}
    & \bn{\textbf{0.722}}{\underline{\textbf{0.600}}} & \bn{\textbf{0.803}}{\underline{\textbf{0.689}}}
    & \bn{\textbf{0.494}}{\underline{\textbf{0.396}}} & \bn{\textbf{0.917}}{\textbf{0.671}} \\
    \bottomrule
    \end{tabular}
    \caption{
    Hierarchical evaluation metrics on FGVCAircraft. Each cell reports \textbf{Base} (top) and \textbf{New} (bottom), averaged over three seeds. Hierarchical F1 is computed as the harmonic mean of hierarchical precision and hierarchical recall for each dataset. \underline{Underline}: best in column. \textbf{Bold}: best within each baseline/extension pair (CoOp vs.\ HyPLO; CoCoOp vs.\ CoHyPLO; MaPLe vs.\ MaHyPLO). \ensuremath{\downarrow} lower is better; \ensuremath{\uparrow} higher is better. \textit{Hier.}: explicitly hierarchy-aware.
    }
    \label{tab:hier_metrics_aircraft_b2n}
\end{table}

% SUN397
\begin{table}[H]
    \centering
    \small
    \setlength{\tabcolsep}{6pt}
    \setlength{\aboverulesep}{0pt}
    \setlength{\belowrulesep}{0pt}
    \renewcommand{\arraystretch}{1.15}
    \begin{tabular}{l c c c c c c c}
    \toprule
    \multirow{2}{*}{\textbf{Method}}
    & \multirow{2}{*}{\textbf{Hier.}}
    & \multicolumn{2}{c}{\cellcolor{greenfirst}\textbf{Hierarchical Distance}}
    & \multicolumn{2}{c}{\cellcolor{greenalt}\textbf{Ancestor Overlap}}
    & \multicolumn{2}{c}{\cellcolor{greencol}\textbf{Local Accuracy}} \\
    \cmidrule(lr){3-4}
    \cmidrule(lr){5-6}
    \cmidrule(lr){7-8}
    &
    & \cellcolor{greenfirst}\textbf{TIE} $\downarrow$
    & \cellcolor{greenfirst}\textbf{LCA} $\downarrow$
    & \cellcolor{greenalt}$\mathbf{J}$ $\uparrow$
    & \cellcolor{greenalt}$\mathbf{F1_H}$ $\uparrow$
    & \cellcolor{greencol}\textbf{S-Acc} $\uparrow$
    & \cellcolor{greencol}\textbf{C-Acc} $\uparrow$ \\
    \midrule
    \addlinespace[4pt]
    HyCoCLIP & \textbf{--}
    & \bn{1.328}{\underline{1.108}} & \bn{1.305}{1.251} & \bn{0.862}{0.883} & \bn{0.898}{0.916} & \bn{0.767}{0.789} & \bn{0.928}{0.961} \\
    \addlinespace[4pt]
    HyCoCLIP$^\dagger$ & \checkmark
    & \bn{1.329}{1.111} & \bn{1.295}{1.249} & \bn{0.865}{\underline{0.884}} & \bn{0.902}{0.917} & \bn{0.765}{\underline{0.791}} & \bn{0.940}{0.960} \\
    \midrule
    \addlinespace[2pt]
    + CoOp & \textbf{--}
    & \bn{1.007}{1.385} & \bn{1.218}{1.307}
    & \bn{0.899}{0.855} & \bn{0.928}{0.898}
    & \bn{0.820}{0.738} & \bn{0.961}{0.955} \\
    \addlinespace[4pt]
    \rowcolor{gray!10}
    \textbf{+ HyPLO} & \checkmark
    & \bn{\underline{\textbf{0.982}}}{\textbf{1.302}} & \bn{\underline{\textbf{1.207}}}{\textbf{1.286}}
    & \bn{\underline{\textbf{0.904}}}{\textbf{0.865}} & \bn{\underline{\textbf{0.931}}}{\textbf{0.905}}
    & \bn{\underline{\textbf{0.828}}}{\textbf{0.754}} & \bn{\underline{\textbf{0.965}}}{\textbf{0.961}} \\
    \midrule
    \addlinespace[2pt]
    + CoCoOp & \textbf{--}
    & \bn{1.101}{1.207} & \bn{1.239}{1.266}
    & \bn{0.889}{0.874} & \bn{0.920}{0.911}
    & \bn{0.805}{0.770} & \bn{0.956}{0.963} \\
    \addlinespace[4pt]
    \rowcolor{gray!10}
    \textbf{+ CoHyPLO} & \checkmark
    & \bn{\textbf{1.063}}{\textbf{1.137}} & \bn{\textbf{1.227}}{\textbf{1.249}}
    & \bn{\textbf{0.895}}{\textbf{0.882}} & \bn{\textbf{0.924}}{\textbf{0.917}}
    & \bn{\textbf{0.813}}{\textbf{0.785}} & \bn{\textbf{0.960}}{\textbf{0.966}} \\
    \midrule
    \addlinespace[2pt]
    + MaPLe & \textbf{--}
    & \bn{1.052}{1.191} & \bn{1.229}{1.266}
    & \bn{0.894}{0.874} & \bn{0.924}{0.912}
    & \bn{0.813}{0.772} & \bn{0.958}{0.962} \\
    \addlinespace[4pt]
    \rowcolor{gray!10}
    \textbf{+ MaHyPLO} & \checkmark
    & \bn{\textbf{1.005}}{\textbf{1.123}} & \bn{\textbf{1.212}}{\underline{\textbf{1.247}}}
    & \bn{\textbf{0.901}}{\textbf{0.883}} & \bn{\textbf{0.929}}{\underline{\textbf{0.918}}}
    & \bn{\textbf{0.824}}{\textbf{0.786}} & \bn{\textbf{0.964}}{\underline{\textbf{0.968}}} \\
    \bottomrule
    \end{tabular}
    \caption{
    Hierarchical evaluation metrics on SUN397. Each cell reports \textbf{Base} (top) and \textbf{New} (bottom), averaged over three seeds. Hierarchical F1 is computed as the harmonic mean of hierarchical precision and hierarchical recall for each dataset. \underline{Underline}: best in column. \textbf{Bold}: best within each baseline/extension pair (CoOp vs.\ HyPLO; CoCoOp vs.\ CoHyPLO; MaPLe vs.\ MaHyPLO). \ensuremath{\downarrow} lower is better; \ensuremath{\uparrow} higher is better. \textit{Hier.}: explicitly hierarchy-aware.
    }
    \label{tab:hier_metrics_sun397_b2n}
\end{table}

% DTD
\begin{table}[H]
    \centering
    \small
    \setlength{\tabcolsep}{6pt}
    \setlength{\aboverulesep}{0pt}
    \setlength{\belowrulesep}{0pt}
    \renewcommand{\arraystretch}{1.15}
    \begin{tabular}{l c c c c c c c}
    \toprule
    \multirow{2}{*}{\textbf{Method}}
    & \multirow{2}{*}{\textbf{Hier.}}
    & \multicolumn{2}{c}{\cellcolor{greenfirst}\textbf{Hierarchical Distance}}
    & \multicolumn{2}{c}{\cellcolor{greenalt}\textbf{Ancestor Overlap}}
    & \multicolumn{2}{c}{\cellcolor{greencol}\textbf{Local Accuracy}} \\
    \cmidrule(lr){3-4}
    \cmidrule(lr){5-6}
    \cmidrule(lr){7-8}
    &
    & \cellcolor{greenfirst}\textbf{TIE} $\downarrow$
    & \cellcolor{greenfirst}\textbf{LCA} $\downarrow$
    & \cellcolor{greenalt}$\mathbf{J}$ $\uparrow$
    & \cellcolor{greenalt}$\mathbf{F1_H}$ $\uparrow$
    & \cellcolor{greencol}\textbf{S-Acc} $\uparrow$
    & \cellcolor{greencol}\textbf{C-Acc} $\uparrow$ \\
    \midrule
    \addlinespace[4pt]
    HyCoCLIP & \textbf{--}
    & \bn{3.512}{2.957} & \bn{2.084}{1.853} & \bn{0.556}{0.643} & \bn{0.639}{0.716} & \bn{0.405}{0.494} & \bn{0.510}{0.653} \\
    \addlinespace[4pt]
    HyCoCLIP$^\dagger$ & \checkmark
    & \bn{3.461}{\underline{2.522}} & \bn{2.042}{\underline{1.676}} & \bn{0.575}{\underline{0.716}} & \bn{0.653}{\underline{0.775}} & \bn{0.435}{\underline{0.595}} & \bn{0.523}{\underline{0.728}} \\
    \midrule
    \addlinespace[2pt]
    + CoOp & \textbf{--}
    & \bn{1.606}{3.691} & \bn{1.484}{2.125}
    & \bn{0.801}{0.537} & \bn{0.839}{0.625}
    & \bn{0.731}{0.372} & \bn{0.786}{0.504} \\
    \addlinespace[4pt]
    \rowcolor{gray!10}
    \textbf{+ HyPLO} & \checkmark
    & \bn{\underline{\textbf{1.449}}}{\textbf{3.507}} & \bn{\underline{\textbf{1.427}}}{\textbf{2.029}}
    & \bn{\underline{\textbf{0.824}}}{\textbf{0.571}} & \bn{\underline{\textbf{0.858}}}{\textbf{0.657}}
    & \bn{\underline{\textbf{0.758}}}{\textbf{0.397}} & \bn{\underline{\textbf{0.814}}}{\textbf{0.574}} \\
    \midrule
    \addlinespace[2pt]
    + CoCoOp & \textbf{--}
    & \bn{1.974}{3.225} & \bn{1.587}{1.951}
    & \bn{0.759}{0.605} & \bn{0.805}{0.683}
    & \bn{0.673}{0.451} & \bn{0.740}{0.598} \\
    \addlinespace[4pt]
    \rowcolor{gray!10}
    \textbf{+ CoHyPLO} & \checkmark
    & \bn{\textbf{1.809}}{\textbf{3.124}} & \bn{\textbf{1.532}}{\textbf{1.916}}
    & \bn{\textbf{0.781}}{\textbf{0.619}} & \bn{\textbf{0.823}}{\textbf{0.695}}
    & \bn{\textbf{0.703}}{\textbf{0.469}} & \bn{\textbf{0.765}}{\textbf{0.615}} \\
    \midrule
    \addlinespace[2pt]
    + MaPLe & \textbf{--}
    & \bn{1.769}{\textbf{2.834}} & \bn{1.528}{\textbf{1.799}}
    & \bn{0.783}{\textbf{0.664}} & \bn{0.824}{\textbf{0.734}}
    & \bn{0.707}{\textbf{0.522}} & \bn{0.765}{\textbf{0.678}} \\
    \addlinespace[4pt]
    \rowcolor{gray!10}
    \textbf{+ MaHyPLO} & \checkmark
    & \bn{\textbf{1.600}}{2.882} & \bn{\textbf{1.477}}{1.819}
    & \bn{\textbf{0.803}}{0.656} & \bn{\textbf{0.841}}{0.727}
    & \bn{\textbf{0.734}}{0.507} & \bn{\textbf{0.789}}{0.673} \\
    \bottomrule
    \end{tabular}
    \caption{
    Hierarchical evaluation metrics on DTD. Each cell reports \textbf{Base} (top) and \textbf{New} (bottom), averaged over three seeds. Hierarchical F1 is computed as the harmonic mean of hierarchical precision and hierarchical recall for each dataset. \underline{Underline}: best in column. \textbf{Bold}: best within each baseline/extension pair (CoOp vs.\ HyPLO; CoCoOp vs.\ CoHyPLO; MaPLe vs.\ MaHyPLO). \ensuremath{\downarrow} lower is better; \ensuremath{\uparrow} higher is better. \textit{Hier.}: explicitly hierarchy-aware.
    }
    \label{tab:hier_metrics_dtd_b2n}
\end{table}

% Eurosat
\begin{table}[H]
    \centering
    \small
    \setlength{\tabcolsep}{6pt}
    \setlength{\aboverulesep}{0pt}
    \setlength{\belowrulesep}{0pt}
    \renewcommand{\arraystretch}{1.15}
    \begin{tabular}{l c c c c c c c}
    \toprule
    \multirow{2}{*}{\textbf{Method}}
    & \multirow{2}{*}{\textbf{Hier.}}
    & \multicolumn{2}{c}{\cellcolor{greenfirst}\textbf{Hierarchical Distance}}
    & \multicolumn{2}{c}{\cellcolor{greenalt}\textbf{Ancestor Overlap}}
    & \multicolumn{2}{c}{\cellcolor{greencol}\textbf{Local Accuracy}} \\
    \cmidrule(lr){3-4}
    \cmidrule(lr){5-6}
    \cmidrule(lr){7-8}
    &
    & \cellcolor{greenfirst}\textbf{TIE} $\downarrow$
    & \cellcolor{greenfirst}\textbf{LCA} $\downarrow$
    & \cellcolor{greenalt}$\mathbf{J}$ $\uparrow$
    & \cellcolor{greenalt}$\mathbf{F1_H}$ $\uparrow$
    & \cellcolor{greencol}\textbf{S-Acc} $\uparrow$
    & \cellcolor{greencol}\textbf{C-Acc} $\uparrow$ \\
    \midrule
    \addlinespace[4pt]
    HyCoCLIP & \textbf{--}
    & \bn{2.237}{2.503} & \bn{1.578}{1.807} & \bn{0.761}{0.671} & \bn{0.807}{0.731} & \bn{0.671}{0.568} & \bn{0.751}{0.624} \\
    \addlinespace[4pt]
    HyCoCLIP$^\dagger$ & \checkmark
    & \bn{3.445}{\underline{1.985}} & \bn{2.049}{\underline{1.616}} & \bn{0.558}{\underline{0.743}} & \bn{0.635}{\underline{0.795}} & \bn{0.432}{\underline{0.640}} & \bn{0.474}{\underline{0.744}} \\
    \midrule
    \addlinespace[2pt]
    + CoOp & \textbf{--}
    & \bn{0.358}{2.878} & \bn{1.090}{1.905}
    & \bn{0.963}{0.626} & \bn{0.970}{0.698}
    & \bn{0.951}{0.489} & \bn{0.959}{0.606} \\
    \addlinespace[4pt]
    \rowcolor{gray!10}
    \textbf{+ HyPLO} & \checkmark
    & \bn{\underline{\textbf{0.318}}}{\textbf{2.593}} & \bn{\underline{\textbf{1.076}}}{\textbf{1.815}}
    & \bn{\underline{\textbf{0.969}}}{\textbf{0.663}} & \bn{\underline{\textbf{0.975}}}{\textbf{0.728}}
    & \bn{\underline{\textbf{0.959}}}{\textbf{0.537}} & \bn{\underline{\textbf{0.965}}}{\textbf{0.648}} \\
    \midrule
    \addlinespace[2pt]
    + CoCoOp & \textbf{--}
    & \bn{0.592}{2.579} & \bn{1.161}{\textbf{1.781}}
    & \bn{0.934}{\textbf{0.680}} & \bn{0.946}{\textbf{0.739}}
    & \bn{0.914}{\textbf{0.571}} & \bn{0.924}{\textbf{0.648}} \\
    \addlinespace[4pt]
    \rowcolor{gray!10}
    \textbf{+ CoHyPLO} & \checkmark
    & \bn{\textbf{0.517}}{\textbf{2.532}} & \bn{\textbf{1.132}}{1.792}
    & \bn{\textbf{0.946}}{0.676} & \bn{\textbf{0.956}}{0.736}
    & \bn{\textbf{0.930}}{0.566} & \bn{\textbf{0.938}}{0.642} \\
    \midrule
    \addlinespace[2pt]
    + MaPLe & \textbf{--}
    & \bn{0.555}{2.501} & \bn{1.143}{1.768}
    & \bn{0.941}{0.681} & \bn{0.952}{0.744}
    & \bn{0.921}{0.559} & \bn{0.936}{0.673} \\
    \addlinespace[4pt]
    \rowcolor{gray!10}
    \textbf{+ MaHyPLO} & \checkmark
    & \bn{\textbf{0.465}}{\textbf{2.079}} & \bn{\textbf{1.120}}{\textbf{1.639}}
    & \bn{\textbf{0.951}}{\textbf{0.732}} & \bn{\textbf{0.960}}{\textbf{0.787}}
    & \bn{\textbf{0.936}}{\textbf{0.618}} & \bn{\textbf{0.944}}{\textbf{0.743}} \\
    \bottomrule
    \end{tabular}
    \caption{
    Hierarchical evaluation metrics on EuroSAT. Each cell reports \textbf{Base} (top) and \textbf{New} (bottom), averaged over three seeds. Hierarchical F1 is computed as the harmonic mean of hierarchical precision and hierarchical recall for each dataset. \underline{Underline}: best in column. \textbf{Bold}: best within each baseline/extension pair (CoOp vs.\ HyPLO; CoCoOp vs.\ CoHyPLO; MaPLe vs.\ MaHyPLO). \ensuremath{\downarrow} lower is better; \ensuremath{\uparrow} higher is better. \textit{Hier.}: explicitly hierarchy-aware.
    }
    \label{tab:hier_metrics_eurosat_b2n}
\end{table}

% UCF101
\begin{table}[H]
    \centering
    \small
    \setlength{\tabcolsep}{6pt}
    \setlength{\aboverulesep}{0pt}
    \setlength{\belowrulesep}{0pt}
    \renewcommand{\arraystretch}{1.15}
    \begin{tabular}{l c c c c c c c}
    \toprule
    \multirow{2}{*}{\textbf{Method}}
    & \multirow{2}{*}{\textbf{Hier.}}
    & \multicolumn{2}{c}{\cellcolor{greenfirst}\textbf{Hierarchical Distance}}
    & \multicolumn{2}{c}{\cellcolor{greenalt}\textbf{Ancestor Overlap}}
    & \multicolumn{2}{c}{\cellcolor{greencol}\textbf{Local Accuracy}} \\
    \cmidrule(lr){3-4}
    \cmidrule(lr){5-6}
    \cmidrule(lr){7-8}
    &
    & \cellcolor{greenfirst}\textbf{TIE} $\downarrow$
    & \cellcolor{greenfirst}\textbf{LCA} $\downarrow$
    & \cellcolor{greenalt}$\mathbf{J}$ $\uparrow$
    & \cellcolor{greenalt}$\mathbf{F1_H}$ $\uparrow$
    & \cellcolor{greencol}\textbf{S-Acc} $\uparrow$
    & \cellcolor{greencol}\textbf{C-Acc} $\uparrow$ \\
    \midrule
    \addlinespace[4pt]
    HyCoCLIP & \textbf{--}
    & \bn{1.516}{1.485} & \bn{1.308}{1.332} & \bn{0.857}{0.850} & \bn{0.897}{0.889} & \bn{0.746}{0.749} & \bn{0.946}{0.919} \\
    \addlinespace[4pt]
    HyCoCLIP$^\dagger$ & \checkmark
    & \bn{1.764}{1.534} & \bn{1.368}{1.340} & \bn{0.832}{0.846} & \bn{0.877}{0.887} & \bn{0.711}{0.740} & \bn{0.920}{0.920} \\
    \midrule
    \addlinespace[2pt]
    + CoOp & \textbf{--}
    & \bn{0.902}{1.932} & \bn{1.173}{1.432}
    & \bn{0.919}{0.802} & \bn{0.942}{0.856}
    & \bn{0.853}{0.661} & \bn{\textbf{0.974}}{0.906} \\
    \addlinespace[4pt]
    \rowcolor{gray!10}
    \textbf{+ HyPLO} & \checkmark
    & \bn{\underline{\textbf{0.877}}}{\textbf{1.797}} & \bn{\textbf{1.162}}{\textbf{1.374}}
    & \bn{\textbf{0.924}}{\textbf{0.831}} & \bn{\textbf{0.946}}{\textbf{0.875}}
    & \bn{\underline{\textbf{0.865}}}{\textbf{0.714}} & \bn{0.973}{\textbf{0.913}} \\
    \midrule
    \addlinespace[2pt]
    + CoCoOp & \textbf{--}
    & \bn{1.150}{1.669} & \bn{1.227}{1.358}
    & \bn{0.895}{0.838} & \bn{0.924}{0.881}
    & \bn{0.818}{0.729} & \bn{0.955}{0.913} \\
    \addlinespace[4pt]
    \rowcolor{gray!10}
    \textbf{+ CoHyPLO} & \checkmark
    & \bn{\textbf{1.094}}{\textbf{1.517}} & \bn{\textbf{1.202}}{\textbf{1.301}}
    & \bn{\textbf{0.906}}{\textbf{0.861}} & \bn{\textbf{0.933}}{\textbf{0.900}}
    & \bn{\textbf{0.836}}{\textbf{0.758}} & \bn{\textbf{0.961}}{\underline{\textbf{0.941}}} \\
    \midrule
    \addlinespace[2pt]
    + MaPLe & \textbf{--}
    & \bn{1.028}{1.648} & \bn{1.184}{1.356}
    & \bn{0.914}{0.838} & \bn{0.939}{0.881}
    & \bn{0.846}{0.723} & \bn{0.971}{0.922} \\
    \addlinespace[4pt]
    \rowcolor{gray!10}
    \textbf{+ MaHyPLO} & \checkmark
    & \bn{\textbf{0.942}}{\underline{\textbf{1.439}}} & \bn{\underline{\textbf{1.161}}}{\underline{\textbf{1.298}}}
    & \bn{\underline{\textbf{0.925}}}{\underline{\textbf{0.864}}} & \bn{\underline{\textbf{0.947}}}{\underline{\textbf{0.901}}}
    & \bn{\textbf{0.864}}{\underline{\textbf{0.768}}} & \bn{\underline{\textbf{0.976}}}{\textbf{0.934}} \\
    \bottomrule
    \end{tabular}
    \caption{
    Hierarchical evaluation metrics on UCF101. Each cell reports \textbf{Base} (top) and \textbf{New} (bottom), averaged over three seeds. Hierarchical F1 is computed as the harmonic mean of hierarchical precision and hierarchical recall for each dataset. \underline{Underline}: best in column. \textbf{Bold}: best within each baseline/extension pair (CoOp vs.\ HyPLO; CoCoOp vs.\ CoHyPLO; MaPLe vs.\ MaHyPLO). \ensuremath{\downarrow} lower is better; \ensuremath{\uparrow} higher is better. \textit{Hier.}: explicitly hierarchy-aware.
    }
    \label{tab:hier_metrics_ucf101_b2n}
\end{table}

\section{Table 1,2 and 3 Extended}
\label{app:stds}

This section provides the extended tables of the main experiments that additionally provide the standard deviation to show the uncertainty across seeds. All average values are excluded.

\begin{table}[H]
    \centering
    \footnotesize
    \setlength{\tabcolsep}{6pt}
    \renewcommand{\arraystretch}{1.15}
    \newcommand{\std}[1]{\ensuremath{\pm}\,#1}
    
    \begin{tabular}{l c @{\hspace{12pt}} cccc}
    \toprule
    & \textbf{Source} & \multicolumn{4}{c}{\textbf{Target}} \\
    \cmidrule(lr){2-2} \cmidrule(lr){3-6}
    \textbf{Method}
    & \textbf{ImageNet}
    & \textbf{-V2}
    & \textbf{-S}
    & \textbf{-A}
    & \textbf{-R} \\
    \midrule
    + CoOp
    & \textit{47.23} \std{0.15}
    & 40.83 \std{0.25}
    & 26.80 \std{0.17}
    & 15.63 \std{0.32}
    & 47.63 \std{0.25} \\
    \rowcolor{gray!10}\textbf{+ HyPLO}
    & \textbf{\textit{47.53}} \std{0.15}
    & \textbf{41.47} \std{0.12}
    & \textbf{27.07} \std{0.06}
    & \textbf{15.67} \std{0.21}
    & \textbf{48.10} \std{0.20} \\
    \midrule
    + CoCoOp
    & \textit{46.90} \std{0.56}
    & 40.93 \std{0.45}
    & 26.40 \std{0.17}
    & \underline{\textbf{16.03}} \std{0.29}
    & 47.80 \std{0.20} \\
    \rowcolor{gray!10}\textbf{+ CoHyPLO}
    & \textbf{\textit{47.60}} \std{0.10}
    & \textbf{41.67} \std{0.06}
    & \textbf{26.77} \std{0.15}
    & 15.83 \std{0.21}
    & \textbf{47.93} \std{0.31} \\
    \midrule
    + MaPLe
    & \textit{46.57} \std{0.06}
    & 40.50 \std{0.10}
    & \textbf{26.20} \std{0.44}
    & \textbf{15.40} \std{0.20}
    & \textbf{47.97} \std{0.15} \\
    \rowcolor{gray!10}\textbf{+ MaHyPLO}
    & \textbf{\textit{47.37}} \std{0.15}
    & \textbf{41.13} \std{0.06}
    & 26.17 \std{0.12}
    & 15.30 \std{0.26}
    & 47.40 \std{0.26} \\
    \midrule
    + PromptSRC
    & \underline{\textit{48.63}} \std{0.06}
    & 41.97 \std{0.06}
    & \underline{28.00} \std{0.20}
    & 14.77 \std{0.15}
    & \underline{49.70} \std{0.10} \\
    \midrule
    + CoPrompt
    & \underline{\textit{48.63}} \std{0.06}
    & \underline{42.40} \std{0.35}
    & 27.87 \std{0.21}
    & 14.87 \std{0.15}
    & 49.00 \std{0.26} \\
    \bottomrule
    \end{tabular}
    
    \caption{Domain generalization Table~\ref{tab:domaingen} with standard deviation. The standard deviation is taken over three seeds per table value.}
    \label{tab:domaingen_std}
\end{table}

\begin{table}[H]
\centering
\scriptsize
\setlength{\tabcolsep}{3pt}
\renewcommand{\arraystretch}{1.05}
\newcommand{\std}[1]{\ensuremath{\pm}\,#1}

% ---------- Row 1 ----------
\begin{subtable}{0.31\textwidth}\centering
\caption{ImageNet}%%%%%%%%%%%%%%%%%%%%%%%%%%%%%%%%%%
\begin{tabular}{@{}lcc@{}}\toprule
\textbf{Method} & \textbf{Base} & \textbf{New} \\\midrule
+ CoOp & 48.90 \std{0.10} & 49.17 \std{0.15} \\
\rowcolor{gray!10}\textbf{+ HyPLO} & \textbf{49.07} \std{0.06} & \textbf{50.47} \std{0.50} \\\midrule
+ CoCoOp & \textbf{48.90} \std{0.10} & 50.47 \std{0.38} \\
\rowcolor{gray!10}\textbf{+ CoHyPLO} & 48.47 \std{0.15} & \textbf{52.03} \std{0.21} \\\midrule
+ MaPLe & 50.27 \std{0.06} & 50.97 \std{0.55} \\
\rowcolor{gray!10}\textbf{+ MaHyPLO} & \textbf{50.50} \std{0.10} & \underline{\textbf{52.33}} \std{0.42} \\\midrule
+ PromptSRC & \underline{52.77} \std{0.12} & 51.30 \std{0.36} \\\midrule
+ CoPrompt & 50.17 \std{0.46} & 50.90 \std{0.20} \\\bottomrule
\end{tabular}\end{subtable}\hfill%
\begin{subtable}{0.31\textwidth}\centering
\caption{Caltech101}%%%%%%%%%%%%%%%%%%%%%%%%%%%%%%%%%%
\begin{tabular}{@{}lcc@{}}\toprule
\textbf{Method} & \textbf{Base} & \textbf{New} \\\midrule
+ CoOp & \textbf{95.53} \std{0.06} & 83.07 \std{1.30} \\
\rowcolor{gray!10}\textbf{+ HyPLO} & 95.37 \std{0.47} & \textbf{84.20} \std{1.87} \\\midrule
+ CoCoOp & 95.10 \std{0.30} & 88.13 \std{0.86} \\
\rowcolor{gray!10}\textbf{+ CoHyPLO} & \textbf{95.23} \std{0.25} & \textbf{89.20} \std{0.10} \\\midrule
+ MaPLe & 95.57 \std{0.06} & \underline{\textbf{90.40}} \std{0.87} \\
\rowcolor{gray!10}\textbf{+ MaHyPLO} & \textbf{95.83} \std{0.23} & 90.30 \std{1.15} \\\midrule
+ PromptSRC & \underline{96.13} \std{0.15} & 85.30 \std{3.56} \\\midrule
+ CoPrompt & 96.03 \std{0.35} & 88.47 \std{0.47} \\\bottomrule
\end{tabular}\end{subtable}\hfill%
\begin{subtable}{0.31\textwidth}\centering
\caption{OxfordPets}%%%%%%%%%%%%%%%%%%%%%%%%%%%%%%%%%%
\begin{tabular}{@{}lcc@{}}\toprule
\textbf{Method} & \textbf{Base} & \textbf{New} \\\midrule
+ CoOp & 76.60 \std{1.14} & \textbf{61.83} \std{0.31} \\
\rowcolor{gray!10}\textbf{+ HyPLO} & \textbf{78.63} \std{1.24} & \textbf{61.83} \std{2.66} \\\midrule
+ CoCoOp & 75.8 \std{0.50} & 66.77 \std{1.19} \\
\rowcolor{gray!10}\textbf{+ CoHyPLO} & \textbf{76.40} \std{1.35} & \textbf{68.47} \std{0.64} \\\midrule
+ MaPLe & 74.60 \std{0.70} & 70.97 \std{2.59} \\
\rowcolor{gray!10}\textbf{+ MaHyPLO} & \textbf{77.73} \std{0.06} & \underline{\textbf{73.13}} \std{1.12} \\\midrule
+ PromptSRC & \underline{79.90} \std{0.78} & 71.33 \std{0.67} \\\midrule
+ CoPrompt & 76.93 \std{0.91} & 69.07 \std{2.40} \\\bottomrule
\end{tabular}\end{subtable}

\vspace{0.8em}

% ---------- Row 2 ----------
\begin{subtable}{0.31\textwidth}\centering
\caption{StanfordCars}%%%%%%%%%%%%%%%%%%%%%%%%%%%%%%%%%%
\begin{tabular}{@{}lcc@{}}\toprule
\textbf{Method} & \textbf{Base} & \textbf{New} \\\midrule
+ CoOp & \textbf{27.47} \std{0.57} & 15.60 \std{1.04} \\
\rowcolor{gray!10}\textbf{+ HyPLO} & 26.20 \std{0.56} & \textbf{16.90} \std{0.85} \\\midrule
+ CoCoOp & \textbf{20.57} \std{0.21} & 20.10 \std{0.30} \\
\rowcolor{gray!10}\textbf{+ CoHyPLO} & 19.83 \std{0.21} & \underline{\textbf{20.97}} \std{0.38} \\\midrule
+ MaPLe & 22.77 \std{0.21} & 19.73 \std{0.98} \\
\rowcolor{gray!10}\textbf{+ MaHyPLO} & \textbf{23.70} \std{0.56} & \textbf{20.53} \std{0.06} \\\midrule
+ PromptSRC & \underline{31.07} \std{0.57} & 20.93 \std{1.00} \\\midrule
+ CoPrompt & 25.37 \std{0.60} & \underline{20.97} \std{0.65} \\\bottomrule
\end{tabular}\end{subtable}\hfill%
\begin{subtable}{0.31\textwidth}\centering
\caption{Flowers102}%%%%%%%%%%%%%%%%%%%%%%%%%%%%%%%%%%
\begin{tabular}{@{}lcc@{}}\toprule
\textbf{Method} & \textbf{Base} & \textbf{New} \\\midrule
+ CoOp & \textbf{78.80} \std{0.40} & \textbf{22.00} \std{2.72} \\
\rowcolor{gray!10}\textbf{+ HyPLO} & 78.73 \std{0.86} & 19.67 \std{3.75} \\\midrule
+ CoCoOp & \textbf{59.63} \std{5.55} & 26.77 \std{2.03} \\
\rowcolor{gray!10}\textbf{+ CoHyPLO} & 59.03 \std{3.17} & \textbf{29.70} \std{5.35} \\\midrule
+ MaPLe & 63.97 \std{3.59} & 32.20 \std{1.75} \\
\rowcolor{gray!10}\textbf{+ MaHyPLO} & \textbf{65.80} \std{1.82} & \underline{\textbf{34.80}} \std{2.16} \\\midrule
+ PromptSRC & \underline{85.77} \std{1.88} & 29.27 \std{2.30} \\\midrule
+ CoPrompt & 70.27 \std{1.63} & 27.37 \std{2.94} \\\bottomrule
\end{tabular}\end{subtable}\hfill%
\begin{subtable}{0.31\textwidth}\centering
\caption{Food101}%%%%%%%%%%%%%%%%%%%%%%%%%%%%%%%%%%
\begin{tabular}{@{}lcc@{}}\toprule
\textbf{Method} & \textbf{Base} & \textbf{New} \\\midrule
+ CoOp & 72.20 \std{0.61} & 60.47 \std{2.55} \\
\rowcolor{gray!10}\textbf{+ HyPLO} & \textbf{73.37} \std{0.61} & \textbf{61.90} \std{1.41} \\\midrule
+ CoCoOp & 72.30 \std{0.17} & 66.43 \std{1.01} \\
\rowcolor{gray!10}\textbf{+ CoHyPLO} & \textbf{72.60} \std{0.00} & \textbf{69.50} \std{0.62} \\\midrule
+ MaPLe & 73.20 \std{0.26} & 70.27 \std{0.71} \\
\rowcolor{gray!10}\textbf{+ MaHyPLO} & \textbf{73.57} \std{0.21} & \underline{\textbf{71.00}} \std{0.44} \\\midrule
+ PromptSRC & \underline{74.57} \std{0.25} & 69.67 \std{0.50} \\\midrule
+ CoPrompt & 73.20 \std{0.10} & 69.77 \std{0.76} \\\bottomrule
\end{tabular}\end{subtable}

\vspace{0.8em}

% ---------- Row 3 ----------
\begin{subtable}{0.31\textwidth}\centering
\caption{FGVCAircraft}%%%%%%%%%%%%%%%%%%%%%%%%%%%%%%%%%%
\begin{tabular}{@{}lcc@{}}\toprule
\textbf{Method} & \textbf{Base} & \textbf{New} \\\midrule
+ CoOp & 14.20 \std{0.44} & 5.57 \std{0.06} \\
\rowcolor{gray!10}\textbf{+ HyPLO} & \underline{\textbf{15.30}} \std{0.62} & \textbf{6.30} \std{1.15} \\\midrule
+ CoCoOp & \textbf{10.33} \std{0.91} & 5.90 \std{0.35} \\
\rowcolor{gray!10}\textbf{+ CoHyPLO} & \textbf{10.33} \std{0.65} & \textbf{7.17} \std{1.40} \\\midrule
+ MaPLe & 8.07 \std{0.87} & 6.13 \std{1.15} \\
\rowcolor{gray!10}\textbf{+ MaHyPLO} & \textbf{11.63} \std{0.71} & \underline{\textbf{7.93}} \std{0.50} \\\midrule
+ PromptSRC & 12.73 \std{0.59} & 5.30 \std{1.06} \\\midrule
+ CoPrompt & 10.40 \std{0.79} & 4.10 \std{1.25} \\\bottomrule
\end{tabular}\end{subtable}\hfill%
\begin{subtable}{0.31\textwidth}\centering
\caption{SUN397}%%%%%%%%%%%%%%%%%%%%%%%%%%%%%%%%%%
\begin{tabular}{@{}lcc@{}}\toprule
\textbf{Method} & \textbf{Base} & \textbf{New} \\\midrule
+ CoOp & 71.43 \std{0.12} & 61.47 \std{0.15} \\
\rowcolor{gray!10}\textbf{+ HyPLO} & \textbf{71.63} \std{0.21} & \textbf{63.47} \std{0.84} \\\midrule
+ CoCoOp & 68.90 \std{0.53} & 66.33 \std{0.06} \\
\rowcolor{gray!10}\textbf{+ CoHyPLO} & \textbf{69.57} \std{0.32} & \textbf{68.00} \std{0.10} \\\midrule
+ MaPLe & 70.30 \std{0.44} & 67.07 \std{0.49} \\
\rowcolor{gray!10}\textbf{+ MaHyPLO} & \textbf{71.00} \std{0.35} & \textbf{68.50} \std{0.80} \\\midrule
+ PromptSRC & \underline{73.13} \std{0.31} & \underline{68.73} \std{0.40} \\\midrule
+ CoPrompt & 70.37 \std{0.45} & 66.90 \std{0.44} \\\bottomrule
\end{tabular}\end{subtable}\hfill%
\begin{subtable}{0.31\textwidth}\centering
\caption{DTD}%%%%%%%%%%%%%%%%%%%%%%%%%%%%%%%%%%
\begin{tabular}{@{}lcc@{}}\toprule
\textbf{Method} & \textbf{Base} & \textbf{New} \\\midrule
+ CoOp & 68.10 \std{0.95} & \textbf{27.93} \std{0.91} \\
\rowcolor{gray!10}\textbf{+ HyPLO} & \textbf{70.3} \std{2.76} & 27.57 \std{1.36} \\\midrule
+ CoCoOp & 60.00 \std{3.44} & 33.87 \std{3.32} \\
\rowcolor{gray!10}\textbf{+ CoHyPLO} & \textbf{62.70} \std{2.05} & \textbf{35.40} \std{2.08} \\\midrule
+ MaPLe & 64.33 \std{1.04} & \textbf{38.23} \std{1.36} \\
\rowcolor{gray!10}\textbf{+ MaHyPLO} & \textbf{67.73} \std{2.90} & 37.87 \std{2.40} \\\midrule
+ PromptSRC & \underline{71.63} \std{1.75} & 36.63 \std{0.81} \\\midrule
+ CoPrompt & 68.83 \std{1.82} & \underline{39.30} \std{2.82} \\\bottomrule
\end{tabular}\end{subtable}

\vspace{0.8em}

% ---------- Row 4 ----------
\begin{subtable}{0.31\textwidth}\centering
\caption{EuroSAT}%%%%%%%%%%%%%%%%%%%%%%%%%%%%%%%%%%
\begin{tabular}{@{}lcc@{}}\toprule
\textbf{Method} & \textbf{Base} & \textbf{New} \\\midrule
+ CoOp & 91.07 \std{1.67} & 46.60 \std{5.34} \\
\rowcolor{gray!10}\textbf{+ HyPLO} & \underline{\textbf{91.70}} \std{0.30} & \textbf{51.83} \std{2.26} \\\midrule
+ CoCoOp & 86.53 \std{3.03} & 49.13 \std{11.15} \\
\rowcolor{gray!10}\textbf{+ CoHyPLO} & \textbf{87.33} \std{1.10} & \textbf{52.60} \std{11.84} \\\midrule
+ MaPLe & 86.57 \std{1.29} & 51.70 \std{9.15} \\
\rowcolor{gray!10}\textbf{+ MaHyPLO} & \textbf{88.73} \std{2.50} & \underline{\textbf{59.90}} \std{2.70} \\\midrule
+ PromptSRC & 87.23 \std{1.39} & 43.87 \std{2.47} \\\midrule
+ CoPrompt & 86.10 \std{2.17} & 47.20 \std{3.48} \\\bottomrule
\end{tabular}\end{subtable}\hfill%
\begin{subtable}{0.31\textwidth}\centering
\caption{UCF101}%%%%%%%%%%%%%%%%%%%%%%%%%%%%%%%%%%
\begin{tabular}{@{}lcc@{}}\toprule
\textbf{Method} & \textbf{Base} & \textbf{New} \\\midrule
+ CoOp & 72.20 \std{0.72} & 46.63 \std{1.16} \\
\rowcolor{gray!10}\textbf{+ HyPLO} & \textbf{72.33} \std{1.01} & \textbf{47.50} \std{0.90} \\\midrule
+ CoCoOp & 65.17 \std{1.12} & 52.37 \std{1.99} \\
\rowcolor{gray!10}\textbf{+ CoHyPLO} & \textbf{65.57} \std{1.18} & \textbf{54.27} \std{0.67} \\\midrule
+ MaPLe & 67.03 \std{0.35} & 53.20 \std{0.30} \\
\rowcolor{gray!10}\textbf{+ MaHyPLO} & \textbf{69.00} \std{0.72} & \underline{\textbf{57.87}} \std{1.07} \\\midrule
+ PromptSRC & \underline{73.67} \std{1.27} & 55.90 \std{1.13} \\\midrule
+ CoPrompt & 69.77 \std{0.91} & 52.23 \std{1.86} \\\bottomrule
\end{tabular}\end{subtable}

\caption{Base-to-new generalization Table~\ref{tab:b2n_all} with standard deviation. The standard deviation is taken over three seeds per table value.}
\label{tab:b2n_all_appendix}
\end{table}

\begin{table}
\centering
\footnotesize
\setlength{\tabcolsep}{6pt}
\renewcommand{\arraystretch}{1.15}
\newcommand{\std}[1]{\ensuremath{\pm}\,#1}

\resizebox{\textwidth}{!}{%
\begin{tabular}{l c @{\hspace{12pt}} cccccccccc}
\toprule
& \textbf{Source} & \multicolumn{10}{c}{\textbf{Target}} \\
\cmidrule(lr){2-2} \cmidrule(lr){3-12}
\textbf{Method}
& \rotatebox{90}{ImageNet}
& \rotatebox{90}{Caltech101}
& \rotatebox{90}{OxfordPets}
& \rotatebox{90}{StanfordCars}
& \rotatebox{90}{Flowers102}
& \rotatebox{90}{Food101}
& \rotatebox{90}{Aircraft}
& \rotatebox{90}{SUN397}
& \rotatebox{90}{DTD}
& \rotatebox{90}{EuroSAT}
& \rotatebox{90}{UCF101} \\
\midrule
CoOp
& \textit{47.23} \std{0.15}
& 86.17 \std{0.15}
& 54.80 \std{0.90}
& 10.27 \std{0.31}
& 29.10 \std{0.92}
& 56.40 \std{1.51}
& \textbf{3.70} \std{0.36}
& 51.33 \std{0.40}
& 27.10 \std{0.20}
& 33.43 \std{0.97}
& 45.03 \std{0.42} \\

\rowcolor{gray!10}\textbf{HyPLO}
& \textbf{\textit{47.53}} \std{0.15}
& \textbf{86.80} \std{0.36}
& \textbf{55.27} \std{0.21}
& \textbf{11.83} \std{0.31}
& \textbf{30.10} \std{0.53}
& \textbf{58.20} \std{0.10}
& 3.33 \std{0.15}
& \textbf{52.83} \std{0.06}
& \textbf{27.17} \std{0.21}
& \textbf{34.90} \std{2.11}
& \textbf{46.63} \std{0.40} \\
\midrule

CoCoOp
& \textit{46.90} \std{0.56}
& 85.90 \std{0.26}
& 53.70 \std{1.07}
& 11.33 \std{0.29}
& 28.77 \std{0.40}
& 56.70 \std{1.21}
& \textbf{3.27} \std{0.50}
& 52.30 \std{0.10}
& \textbf{27.23} \std{0.23}
& \textbf{35.67} \std{2.21}
& 45.90 \std{1.78} \\

\rowcolor{gray!10}\textbf{CoHyPLO}
& \textbf{\textit{47.60}} \std{0.10}
& \textbf{86.90} \std{0.10}
& \textbf{55.97} \std{0.50}
& \underline{\textbf{11.93}} \std{0.40}
& \textbf{30.17} \std{0.47}
& \textbf{58.57} \std{0.75}
& 3.23 \std{0.25}
& \textbf{54.30} \std{0.17}
& 26.87 \std{0.68}
& 34.97 \std{2.53}
& \textbf{47.10} \std{0.61} \\
\midrule

MaPLe
& \textit{46.57} \std{0.06}
& 86.43 \std{0.40}
& 55.10 \std{0.87}
& 10.40 \std{0.00}
& 26.77 \std{0.50}
& 54.57 \std{0.84}
& 3.53 \std{0.55}
& 51.73 \std{0.32}
& 27.40 \std{1.06}
& \underline{\textbf{35.83}} \std{1.55}
& 45.50 \std{1.05} \\

\rowcolor{gray!10}\textbf{MaHyPLO}
& \textbf{\textit{47.37}} \std{0.15}
& \textbf{88.03} \std{0.15}
& \textbf{55.90} \std{0.62}
& \textbf{11.27} \std{0.15}
& \underline{\textbf{30.70}} \std{0.87}
& \underline{\textbf{59.00}} \std{1.23}
& \textbf{3.93} \std{0.25}
& \underline{\textbf{54.93}} \std{0.21}
& \underline{\textbf{27.97}} \std{0.55}
& 35.77 \std{1.40}
& \textbf{47.20} \std{0.61} \\
\midrule

PromptSRC
& \underline{\textit{48.63}} \std{0.06}
& 87.40 \std{0.10}
& 56.80 \std{0.53}
& 10.57 \std{0.32}
& 29.00 \std{0.69}
& 55.67 \std{0.38}
& 3.70 \std{0.62}
& 52.70 \std{0.20}
& 26.73 \std{0.25}
& 34.20 \std{1.28}
& 46.43 \std{0.67} \\
\midrule

CoPrompt
& \underline{\textit{48.63}} \std{0.06}
& \underline{88.80} \std{0.35}
& \underline{57.20} \std{0.46}
& 10.83 \std{0.23}
& 28.27 \std{0.40}
& 57.37 \std{0.95}
& \underline{4.53} \std{0.47}
& 53.43 \std{0.15}
& 25.93 \std{1.06}
& 22.43 \std{2.34}
& \underline{48.57} \std{0.57} \\
\bottomrule
\end{tabular}%
}

\caption{
Cross-dataset transfer Table~\ref{tab:xdataset} with standard deviation. The standard deviation is taken over three seeds per table value.
}
\label{tab:xdataset_std}
\end{table}

\end{document}